%% file: manuscript.tex
\documentclass[lettersize,journal]{IEEEtran}
\usepackage{amsmath,amsfonts}
\usepackage{algorithmic}
\usepackage{algorithm}
\usepackage{array}

\usepackage{subcaption}
\usepackage{textcomp}
\usepackage{stfloats}
\usepackage{url}
\usepackage{verbatim}
\usepackage{graphicx}
\usepackage{cite}
\usepackage{scalerel}
\usepackage{tikz}
\usetikzlibrary{svg.path}

\definecolor{orcidlogocol}{HTML}{A6CE39}
\tikzset{
  orcidlogo/.pic={
    \fill[orcidlogocol] svg{M256,128c0,70.7-57.3,128-128,128C57.3,256,0,198.7,0,128C0,57.3,57.3,0,128,0C198.7,0,256,57.3,256,128z};
    \fill[white] svg{M86.3,186.2H70.9V79.1h15.4v48.4V186.2z}
                 svg{M108.9,79.1h41.6c39.6,0,57,28.3,57,53.6c0,27.5-21.5,53.6-56.8,53.6h-41.8V79.1z M124.3,172.4h24.5c34.9,0,42.9-26.5,42.9-39.7c0-21.5-13.7-39.7-43.7-39.7h-23.7V172.4z}
                 svg{M88.7,56.8c0,5.5-4.5,10.1-10.1,10.1c-5.6,0-10.1-4.6-10.1-10.1c0-5.6,4.5-10.1,10.1-10.1C84.2,46.7,88.7,51.3,88.7,56.8z};
  }
}

\newcommand\orcidicon[1]{\href{https://orcid.org/#1}{\mbox{\scalerel*{
\begin{tikzpicture}[yscale=-1,transform shape]
\pic{orcidlogo};
\end{tikzpicture}
}{|}}}}

\usepackage{amssymb}
\usepackage{pifont}
\usepackage{xcolor}
\newcommand{\firstbest}[1]{\textcolor{red}{\textbf{#1}}}
\newcommand{\secondbest}[1]{\textcolor{blue}{#1}}
\newcommand{\colorlegend}{\bigskip \raggedright \footnotesize Legend: \firstbest{best}, \secondbest{second best} result.}%

\usepackage{bm}

\usepackage{soul} 

\definecolor{personyc}{RGB}{155, 58, 246} %
\definecolor{caryc}{RGB}{242, 163, 164} %
\definecolor{bicycleyc}{RGB}{235, 82, 73}   \definecolor{white}{RGB}{255, 255, 255}     

\newcommand{\hlpersonyc}[1]{\sethlcolor{personyc}\textcolor{white}{\hl{#1}}}
\newcommand{\hlcaryc}[1]{\sethlcolor{caryc}\textcolor{white}{\hl{#1}}}
\newcommand{\hlbicycleyc}[1]{\sethlcolor{bicycleyc}\textcolor{white}{\hl{#1}}}

\newcommand{\yololegend}{\bigskip \raggedright \footnotesize Legend: \hlpersonyc{person}, \hlbicycleyc{bicycle}, \hlcaryc{car}.}%

\newcommand{\rebuttal}[1]{{#1}}
\newcommand{\rebuttalcontainer}{}

\usepackage{graphicx}

\usepackage{hyperref}

\begin{document}

\title{Leveraging Content and Context Cues for Low-Light Image Enhancement}

\author{Igor Morawski \orcidicon{0009-0001-4851-4946}, Kai He, Shusil Dangi, and Winston H. Hsu \orcidicon{0000-0002-3330-0638}, ~\IEEEmembership{Senior Member,~IEEE} 
\thanks{ }
\thanks{ }

}

\markboth{ }{ }

\IEEEpubid{} 


\maketitle

\input{Sections/X_Abstract}
\input{Sections/1_Introduction}
\input{Sections/2_Related_Work}

\input{Sections/3_Proposed_Method}
\input{Sections/4_Experimental_Results}

\input{Sections/5_Conclusion}
\input{Sections/X_Acknowledgement}

\bibliographystyle{IEEEtran}
\bibliography{manuscript}

\newpage
\input{Biographies/Biographies_Compilation}
\vfill

\end{document}

%% file: Sections/X_Abstract.tex
\begin{abstract}
Low-light conditions have an adverse impact on machine cognition, limiting the performance of computer vision systems in real life. Since low-light data is limited and difficult to annotate, we focus on image processing to enhance low-light images and improve the performance of any downstream task model, instead of fine-tuning each of the models which can be prohibitively expensive. We propose to improve the existing zero-reference low-light enhancement by leveraging the CLIP model to capture image prior and for semantic guidance. Specifically, we propose a data augmentation strategy to learn an image prior via prompt learning, based on image sampling, to learn the image prior without any need for paired or unpaired normal-light data. Next, we propose a semantic guidance strategy that maximally takes advantage of existing low-light annotation by introducing both content and context cues about the image training patches. We experimentally show, in a qualitative study, that the proposed prior and semantic guidance help to improve the overall image contrast and hue, as well as improve background-foreground discrimination, resulting in reduced over-saturation and noise over-amplification, common in related zero-reference methods. As we target machine cognition, rather than rely on assuming the correlation between human perception and downstream task performance, we conduct and present an ablation study and comparison with related zero-reference methods in terms of task-based performance across many low-light datasets, including image classification, object and face detection, showing the effectiveness of our proposed method.
\end{abstract} 


Repository: 
\href{https://github.com/igor-morawski/tmm-sem/}{github.com/igor-morawski/tmm-sem}

\scriptsize{\textcopyright 2024 IEEE. Personal use of this material is permitted. Permission from IEEE must be obtained for all other uses, in any current or future media, including reprinting/republishing this material for advertising or promotional purposes, creating new collective works, for resale or redistribution to servers or lists, or reuse of any copyrighted component of this work in other works.}\normalsize

%% file: Sections/1_Introduction.tex
\input{Figures/Figures_First}

\section{Introduction}

\IEEEPARstart{C}{urrently}, low-light conditions remain a challenging problem in image processing and computer vision, not only adversely impacting human perception but also the performance of downstream task models in the computer vision pipeline. Noise caused by physical limitations in photon-limited imaging, out-of-focus and motion blur and unnatural appearance caused by lighting from the camera flash, all common in low-light images, contribute to impacting the image quality negatively. Moreover, the image signal processor pipelines often break down in extreme conditions, introducing errors such as incorrect white-balancing and tone-mapping propagated down the computer vision pipeline, presenting further challenges for computer cognition. 

Although it is possible to improve the performance of downstream task models by directly training on low-light data \cite{loh2019getting}, this approach remains impractical if there are many downstream task models, all requiring separate fine-tuning, and when data annotation is difficult or expensive, which is especially true for low-light conditions. On the other hand, introducing an image enhancement model into the cognition pipeline can positively affect the performance of many downstream models. While many image processing methods attempt to address challenges presented by low-light conditions by correlating image quality as perceived by the human visual system with machine cognition performance, we focus on optimizing the image quality directly for machine cognition.

 \IEEEpubidadjcol
Many related works \cite{xu2020learning,zheng2021adaptive,fan2022half,SNRAware,Wu_2023_CVPR,Xu_2023_CVPR,wei2018deep,zhang2019kindling,zhang2021beyond,yang2021sparse,zhang2022deep} require paired low- and normal-light examples for supervising the enhancement model. However, collecting such datasets is costly and laborious because usually the data needs to be captured in multiple shots while varying exposure time or equivalent camera settings. Moreover, because of concerns about misalignment, this often limits the variety of data to static environments, discarding dynamic scenes especially important in real-life scenarios. Motivated by this, some works propose zero-reference enhancement methods \cite{guo2020zero,li2021learning,zheng2022semantic}, which only require low-light data for training. However, these methods focus on human perception and do not leverage semantic knowledge about the data to enrich the enhanced images for downstream task models. 

On the other hand, we propose to focus on machine cognition and leverage zero-shot visual-linguistic models to enrich images in a way that improves the performance of the downstream task models. 

\rebuttal{In our previous work \cite{morawski2024unsupervised}, we proposed an unsupervised image prior leveraging pre-trained CLIP \cite{radford2021learning} as well as semantic segmentation based on detection instances. Compared with our prior work \cite{morawski2024unsupervised}, we introduce two new tasks -- content and context-based tasks -- to integrate the semantic knowledge into the training. In our prior work \cite{morawski2024unsupervised}, we used CLIP \cite{radford2021learning} and antonym pair-based prompts to classify image patches consisting of object instances extracted from the training set. In contrast with this, in this work we used CLIP \cite{radford2021learning} to match an entire batch of images to their corresponding contents (instances visible in a training patch) and context (instances extending outside a training patch). These two tasks work synergistically to improve task-based performance and help making the maximal use of the existing valuable low-light annotation.}

\rebuttal{Our primary motivation is to enhance and enrich low-light images in a way that benefits machine cognition rather than assuming a correlation between human and machine perception. To this end, we propose a semantic guidance method to make the maximal use of the existing valuable low-light annotation. Additionally, we re-use the CLIP \cite{radford2021learning} to constrain the image appearance by learning prompts to constrain the image prior, based on a simple augmentation strategy which only uses low-light images, avoiding any need for paired or unpaired normal-light data in training. Our proposed semantic guidance enriches the images to help downstream task performance. Our proposed learned prior helps by constraining the image contrast, avoiding over-saturation commonly seen in related zero-reference methods.}

Our contributions are as follows:
\begin{itemize}
    \item We propose to semantically guide low-light image enhancement using image abstracted content and context descriptions at a loss level. Abstracted content and context guidance leverage zero-shot capabilities of the CLIP model to scale favorably to include various datasets without any limitations on annotated object categories, instead of fixing the training category set. To maximally take advantage of annotation in low-light images for enriching low-light enhanced images, we propose to realize semantic guidance in two steps. In the first step, within a training batch, we match image descriptions about the image content to images. Conversely, in the second step, we match descriptions of objects extending outside of the image to each image. 
    \item We propose to learn an image prior using prompt learning, using a data augmentation strategy based on image resampling,  eliminating any need for paired or unpaired normal-light data. We experimentally show that the proposed prompts help to guide the image enhancement model by improving the overall image contrast, reducing under- and overexposure, leading to decreased information loss, and reducing over-amplification of noise common in unsupervised enhancement models.
    \item We conduct extensive experiments, including comparison with related unsupervised low-light image enhancement methods and ablation study, to show that our proposed method results in consistent improvements across multiple high-level low-light benchmarks.
\end{itemize}

%% file: Figures/Figures_First.tex
\begin{figure}
    \centering
    \small
    \begin{subfigure}{.20\textwidth}
      \centering
        \includegraphics[width=1\linewidth]{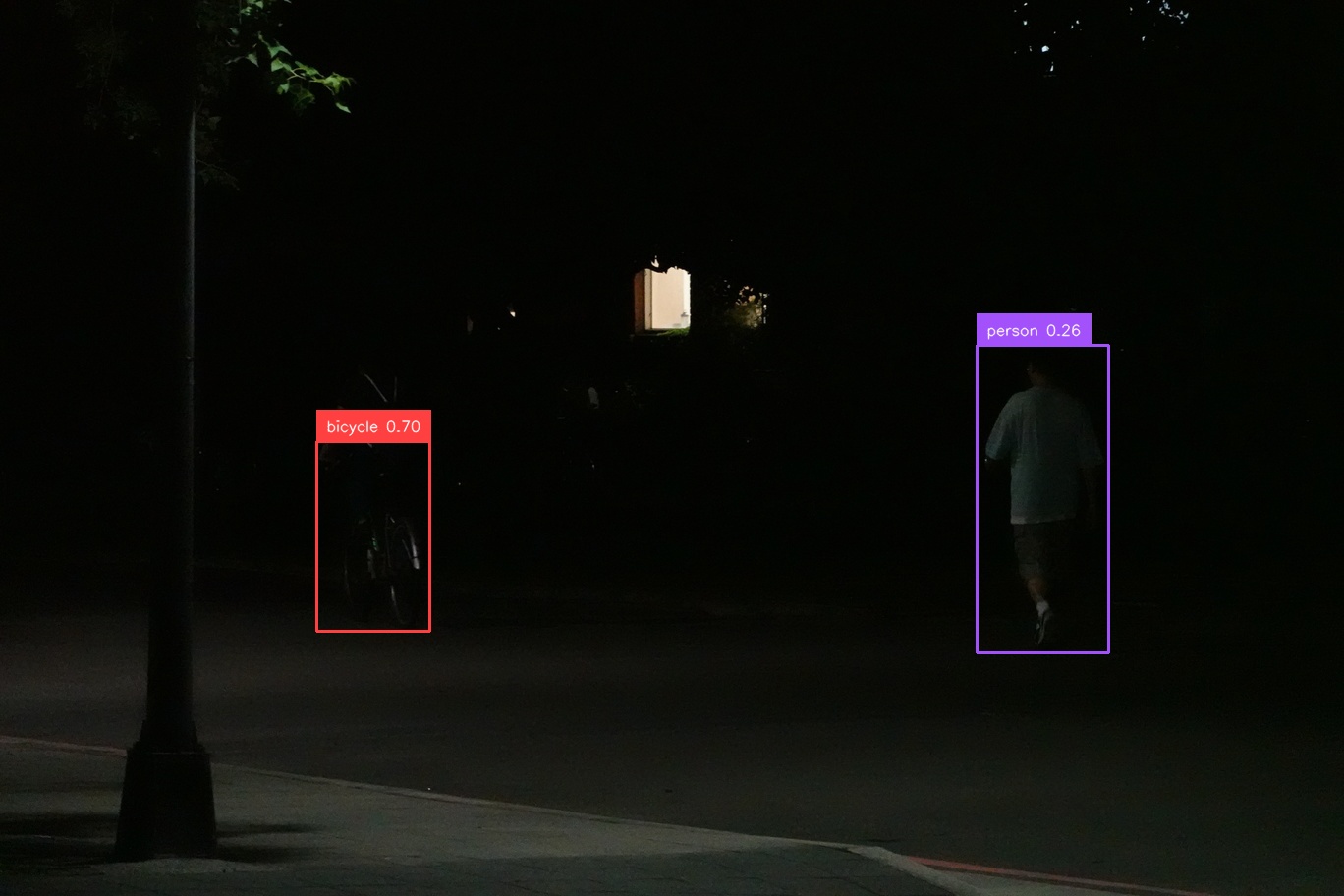}
        \caption*{}
    \end{subfigure} %
    \begin{subfigure}{.20\textwidth}
      \centering
        \includegraphics[width=1\linewidth]{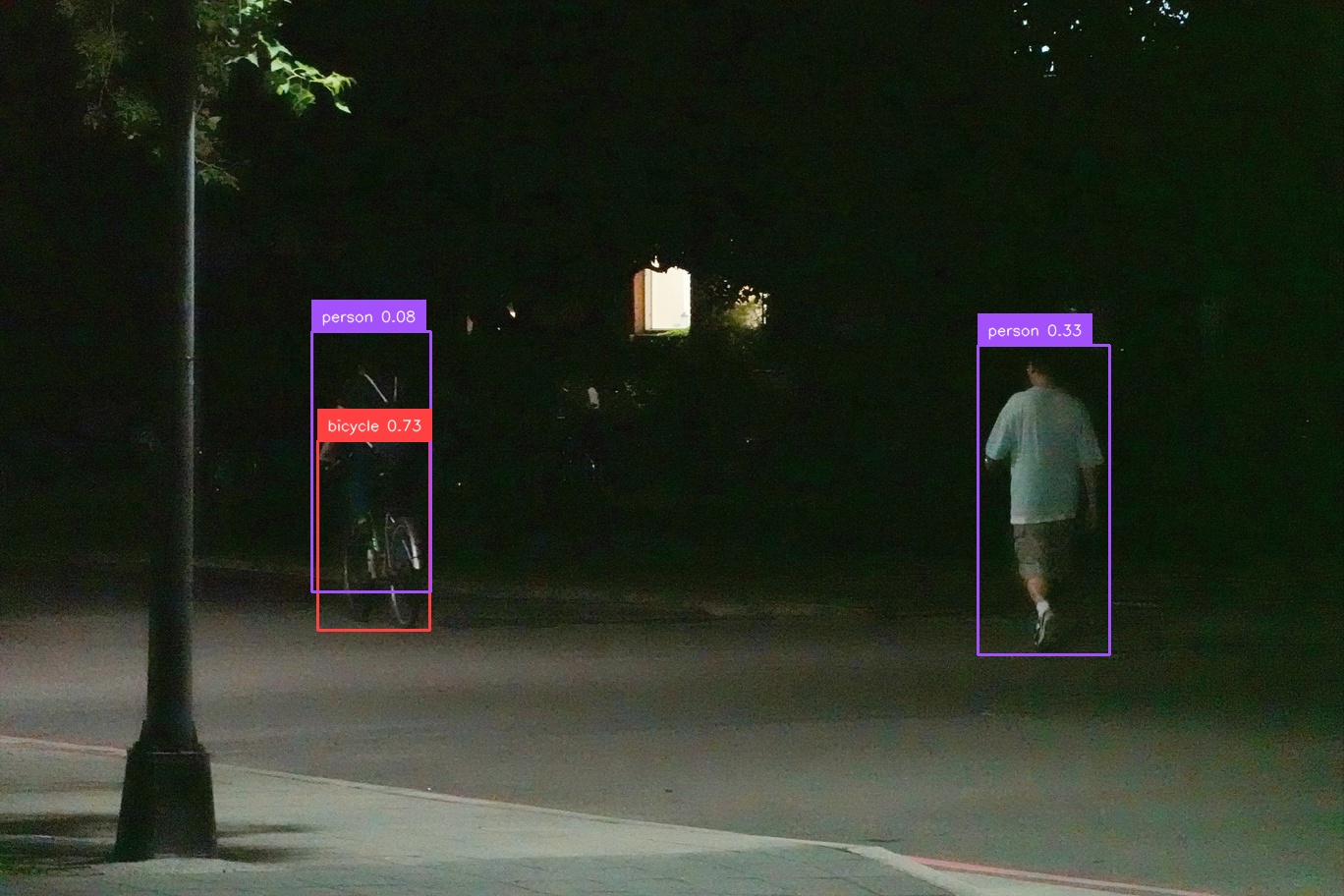}
        \caption*{}
    \end{subfigure} %

\vspace{-1.2em}

    \begin{subfigure}{.20\textwidth}
      \centering
        \includegraphics[width=1\linewidth]{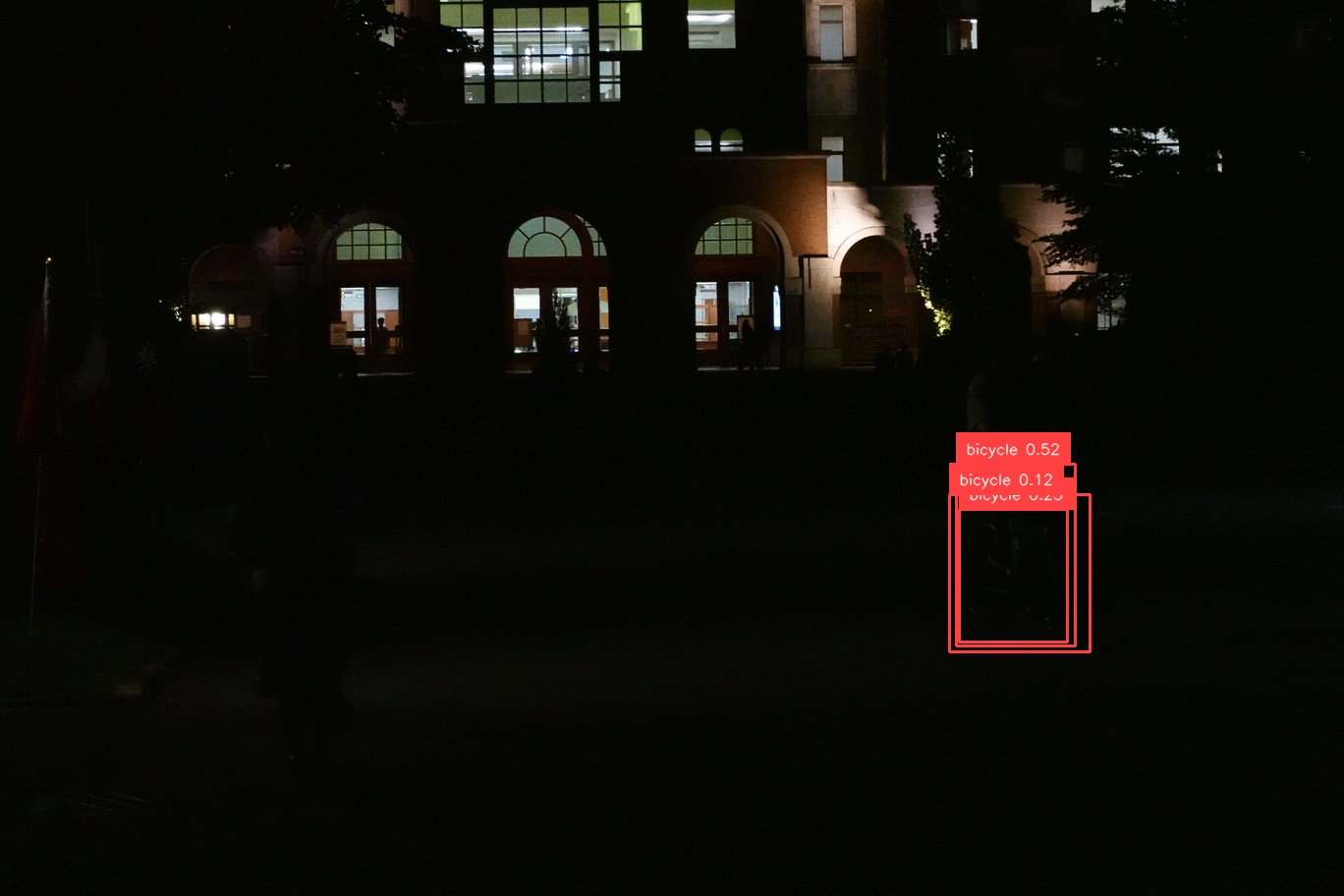}
        \caption*{ Low-Light Data}
    \end{subfigure} 
    \begin{subfigure}{.20\textwidth}
      \centering
        \includegraphics[width=1\linewidth]{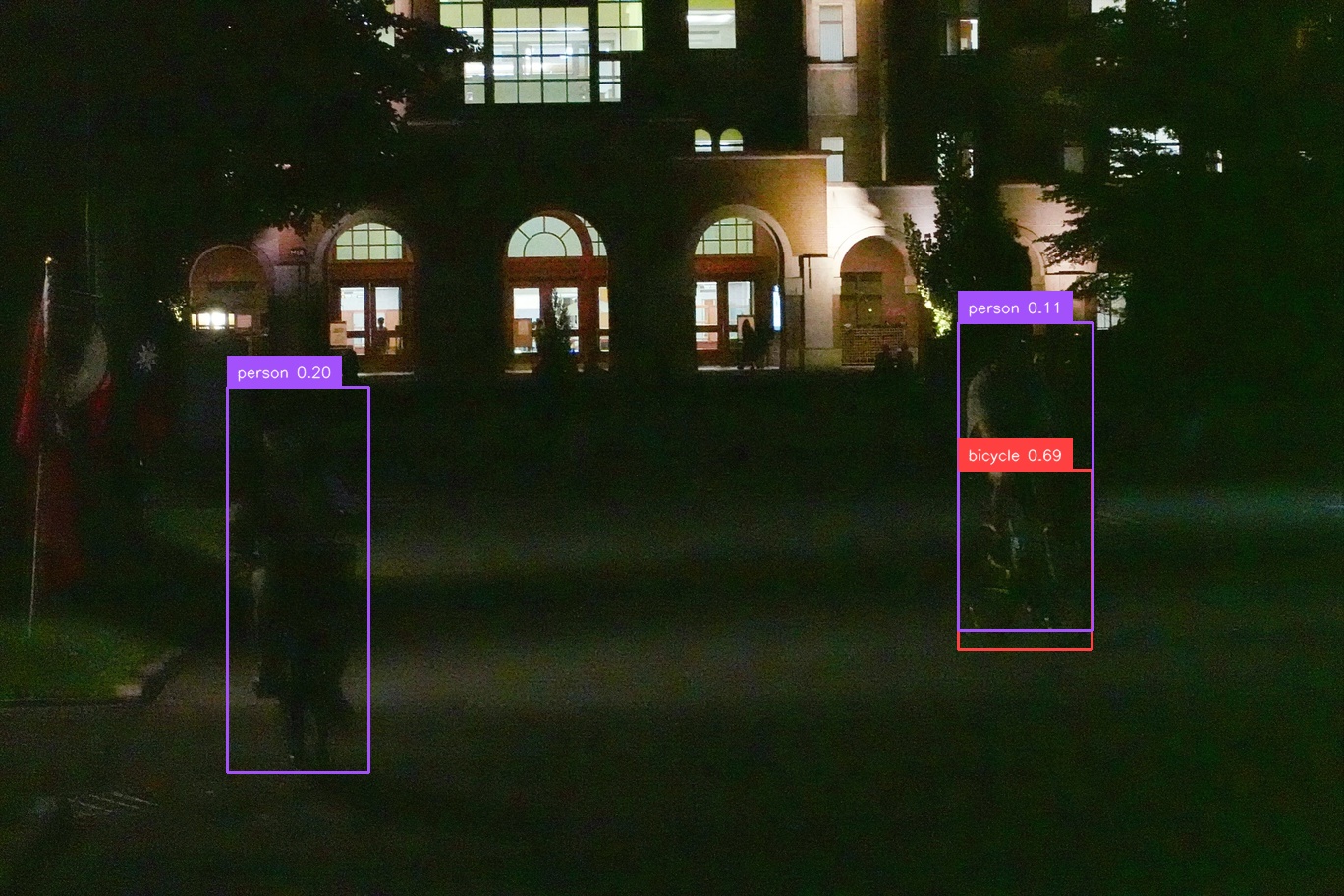}
        \caption*{ Ours}
    \end{subfigure} %


    \caption{As we target machine cognition, rather than rely on assuming the correlation between human perception and downstream task performance, we conduct and present an ablation study and comparison with related zero-reference methods in terms of task-based performance across many low-light datasets, including image classification, object and face detection, showing the effectiveness of our proposed method.}
    \label{fig:first-gifure}
\end{figure}

%% file: Sections/2_Related_Work.tex
\section{Related Work}
\label{sec:related_work}

\input{Figures/Timeline}

\subsection{Low-Light Enhancement}
\textbf{Traditional methods}. Traditional methods for low-light image enhancement include histogram-based and Retinex-based methods. The first group uses histograms to map pixel values of an image at a global or local level. Histogram-based methods are simple and efficient; however they discount structural and semantic information in the images, typically do not address the problem of noise, and produce unnatural visual artifacts such as hue shifts. The latter group, Retinex-based methods, use the basic assumption of Retinex color theory \cite{land1977retinex} that a natural image can be decomposed into reflectance and illumination components and use the estimated reflectance component or modify it as the enhanced output. While the seminal works, single-scale \cite{jobson1997properties} and multi-scale \cite{jobson1997multiscale} Retinex, used relatively simple   Gaussian filters to estimate the reflectance, more complex methods such as \cite{wang2013naturalness,fu2016fusion,guo2016lime} have been proposed over the years, including methods for joint low-light enhancement and denoising  \cite{li2017joint,li2018structure}. Since traditional methods based on Retinex use hand-crafted features and assumptions  about natural images, careful parameter tuning is required for good performance. Because of that, traditional Retinex-based method might fail to generalize to the variety of images in real life and are often not robust to image degradation. \rebuttal{Recently, Wang \textit{et al.} \cite{wang2023low} proposed a virtual exposure method inspired by multi-exposure fusion, which, based on statistical analysis, generates and fuses a set of virtual exposure images based on a single low-light input image. }

\textbf{Learning-based Low-Light Enhancement}. Recent works, motivated by the success of deep learning methods in image processing, propose learning-based low-light image enhancement methods to improve the robustness and limited performance of traditional methods.  Existing learning-based methods can be categorized into end-to-end methods \cite{xu2020learning,jiang2021enlightengan,zheng2021adaptive,fan2022half,dong2022abandoning,SNRAware,Xu_2023_CVPR,Wu_2023_CVPR} and Retinex-based methods \cite{wei2018deep,zhang2019kindling,zhang2021beyond,liu2021retinex,yang2021sparse,wang2022low,wu2022uretinex,ma2022toward,fu2023you,liang2023iterative,zhang2022deep}. End-to-end methods directly enhance the low-light input image; on the other hand, Retinex-based methods incorporate the physics prior by decomposing the image into into reflectance and illumination components. Deep learning methods typically require significant amounts of low- and normal-light image pairs for supervision during the training stage \cite{xu2020learning,zheng2021adaptive,fan2022half,SNRAware,Wu_2023_CVPR,Xu_2023_CVPR,wei2018deep,zhang2019kindling,zhang2021beyond,yang2021sparse,zhang2022deep}. While it is possible to generate synthetic darkened data, models trained on such data have limited capabilities to generalize to real-life images.   At the same time, collecting real paired low- and normal-light data presents significant practical difficulties and high collection costs. The image pairs can be either captured by varying camera settings and lighting parameters \cite{chen2018learning,wei2018deep,cai2018learning,chen2019seeing,hai2023r2rnet} or by post-processing done by experts \cite{bychkovsky2011learning} which requires significant human effort. As an alternative to capturing the scene with the same sensor while varying camera or environment parameters, the same scene can be captured simultaneously using solutions like beam splitter \cite{jiang2019learning,ExLPose_2023_CVPR}. However, because of practical considerations, the majority of existing datasets use one camera to capture static scenes while varying exposure settings. Motivated by significant costs and efforts required to capture and post-process real datasets of paired images, methods that use unpaired low- and normal-light data, such as \rebuttal{\cite{yang2020fidelity,jiang2021enlightengan,liang2023iterative,liang2024pie}}, have been proposed. However, the limitation of these approaches is that a careful selection of well-lit data is still required. \rebuttal{Alternatively, Fu \textit{et al.} \cite{Fu_2023_CVPR_Learning} proposed a Retinex-based method trained using paired low-light images, eliminating the need for clean normal-light data and alleviating some of the practical issues in data collection. However, this proposed method \cite{Fu_2023_CVPR_Learning} still limits the choice of training data to specialized datasets comprised of paired low-light data. }
Motivated by such limitations, Guo \textit{et al.} \cite{guo2020zero,li2021learning} proposed a curve-based method supervised by a set of zero-reference constraints, which does not require any paired or unpaired well-lit data. The zero-reference loss functions proposed in \cite{guo2020zero,li2021learning} are based on assumptions about natural images, namely the Gray-World hypothesis \cite{buchsbaum1980spatial}, average image intensity and spatial consistency of input and out images. Instead of end-to-end image enhancement, these methods formulate low-light enhancement as a problem of estimating curve parameter maps that are recursively applied to the input image, and an additional TV-variation loss term is applied to the estimated parameter map.

\rebuttal{A new promising direction in low-light enhancement is the incorporation of powerful diffusion-based models in the enhancement frameworks \cite{Feng_2024_CVPR,wang2024zero,jiang2023low, yi2023diff, zhou2023pyramid}. Diffusion-based models are generative models that can handle heavy detail distortion restoration and detail inpainting, typically needed in low-light imaging due to extreme noise and limited dynamic range \cite{Feng_2024_CVPR,jiang2023low}. However, the expressive capacity of diffusion-based models comes at the cost of heavy computational resource use during testing, long inference time and difficulties in fitting high-resolution images in the GPU memory. Motivated by this, \cite{jiang2023low} proposes wavelet-based diffusion to improve computational complexity and inference time, \cite{Feng_2024_CVPR} proposes progressive patch fusion at the test time for improved handling of UHD images. \cite{yi2023diff} combined diffusion-based model with Retinex-based enhancement and \cite{zhou2023pyramid} introduced pyramid diffusion to address inference speed and global degradation to which diffusion-based models are typically prone. In contrast with methods that use low-light image as the input to a diffusion model \cite{Feng_2024_CVPR,jiang2023low, yi2023diff, zhou2023pyramid}, Wang \textit{et al.} \cite{wang2024zero} proposes a novel physical quadruple illumination invariant prior and diffusion-based prior-to-image framework to enhance low-light images. Interestingly, this framework \cite{wang2024zero} enables learning low-light enhancement using normal-light images solely, without any need for real low-light data. }

\rebuttal{Tangential to improvement of low-light enhancement, some works \cite{guo2020zero,li2021learning,li2023low} focus on improving computational complexity based on characteristics of low light conditions. Li \textit{et al.} \cite{li2023low} proposed a knowledge distillation method based on gradient-guided low-light enhancement. Guo \textit{et al.} \cite{guo2020zero} proposed a curve-based enhancement, and Li \textit{et al.} \cite{li2021learning} proposed inference-time down- and up-scaling of the curve parameter map for lightweight restoration. In our work, we use Zero-DCE \cite{guo2020zero} as our baseline method because of its remarkable training speed and low inference complexity.}  

\subsection{Low-Light Image Understanding} In recent years, there has been continued interest in image understanding in the challenging low-light conditions, important for real-life deployment of computer vision systems. Low-light image understanding has been explored in high-level computer vision tasks such as image recognition \cite{loh2019getting}, face detection \cite{poor_visibility_benchmark,liang2021recurrent,wang2021hla}, object detection \cite{morawski2021nod,morawski2022genisp,hong2021crafting,loh2019getting}, pose estimation \cite{ExLPose_2023_CVPR} and action recognition \cite{xu2021arid}. Loh and Chan \cite{loh2019getting} showed that convolutional neural networks for image recognition, such as ResNet \cite{he2016deep}, trained using low- and normal-light, do not learn to normalize image features with respect to the lighting conditions, leading to a conclusion that special attention is required to improve the performance in low light conditions. 

Another line of works proposes to improve low-light image understanding by introducing a learning-based enhancement model to the computer vision framework and jointly optimizing the enhancement model and downstream task model or learning the enhancement model under its guidance \cite{liang2021recurrent,morawski2021nod,wang2023tienet,zheng2022semantic,hashmi2023featenhancer}. Some methods incorporate semantic guidance at a loss \cite{liu2017image,aakerberg2022semantic,Wu_2023_CVPR} or feature level \cite{wang2018recovering,li2020blind,Wu_2023_CVPR}. \rebuttal{In low-light image enhancement, \cite{Wu_2023_CVPR} employs both feature- and loss-level semantic guidance using a segmentation model. However, \cite{Wu_2023_CVPR} still requires a fully paired low- and normal-light training dataset, limiting the use of this method.
} \rebuttal{On the other hand,} some works, rather than correlating human perception and downstream task performance, propose to directly optimize image enhancement methods for downstream task performance \cite{liang2021recurrent,morawski2021nod,wang2023tienet,zheng2022semantic,hashmi2023featenhancer}. All in all, introducing an image enhancement model as a part of the computer vision pipeline for high-level cognition tasks can be beneficial in resource-limited scenarios because a single enhancement model can be reused to improve the performance of many different downstream task models, rather than fine-tuning each of the models separately.

Another related direction is to optimize image processing pipeline (ISP) for low-light machine cognition. This line of work is motivated by the observation that, in low light, detectors trained using raw sensor data are inherently more robust than detectors trained using processed sRGB data, as reported in some works \cite{hong2021crafting,ljungbergh2023raw}. Intuitively, this is because traditional ISP pipelines typically break down under extreme low light conditions and difficult to invert errors, introduced by any of many expert-tuned modules, propagate throughout the computer vision pipeline. \rebuttal{Affi and Brown \cite{afifi2019else} experimentally demonstrated that incorrect white balancing negatively impacts image classification and semantic segmentation, thus showing that this limitation can be important for all images, including normal-light images}. Motivated by this observation, \cite{robidoux2021end} proposed a method to tune ISP in a hardware-in-the-loop, optimizing for downstream task performance, and \cite{morawski2022genisp,ljungbergh2023raw,yoshimura2023dynamicisp} propose to use neural ISPs instead. Although using raw sensor data is beneficial in terms of task-based performance, this approach is typically device-specific, meaning that a neural ISP has to be trained for a given camera sensor, and thus has limited use cases. Moreover, the development of neural ISP optimized for machine cognition is further limited by the sparsity of dedicated datasets.

We present a timeline including recent developments in task-based optimization of low-light image enhancement methods, datasets
dedicated to low-light image understanding in Fig. \ref{fig:timeline}. \rebuttal{We also summarize advantages of sRGB and raw sensor data as two different modalities proposed for optimization for machine cognition rather than human perception in Tab. \ref{tab:table-advantages}. Considering pros and cons, in this paper, we use sRGB data. Although raw sensor data is inherently richer than the post-processed photo-finished sRGB data, our approach based on semantic guidance leverages the CLIP  \cite{radford2021learning} model and thus benefits more from the advantages of the sRGB data. First, sRGB datasets are more readily available and require less storage space. Next, because sRGB uses a standard color space, we can leverage pre-trained sRGB models such as CLIP \cite{radford2021learning}, typically not available for sensor-specific color spaces. Finally, standard color space translates to generalization to any sRGB input data. This is in contrast with sensor-specific models using raw sensor data. }

\input{Figures/Figure_Method}

\input{Figures/Figure_Guidance_Methods}

\subsection{CLIP for Image Enhancement}
Contrastive Language-Image Pre-Training (CLIP) \cite{radford2021learning} has achieved remarkable success because of its zero-shot capabilities. \cite{zang2022open,kuo2022f,zhou2022extract,cheng2024yolo} proposed to leverage CLIP \cite{radford2021learning} to generalize to open-vocabulary problems. Wang \textit{et al.} \cite{wang2023exploring} demonstrated that the rich linguistic-visual CLIP prior captures image quality as well as the abstract \textit{feel} of  images. Motivated by this, \cite{liang2023iterative} showed that CLIP can be used to discriminate between different lighting conditions in natural images at a global and local level. Next, Liang \textit{et al.} \cite{liang2023iterative} proposed to use CLIP and prompt learning for backlit image enhancement, and iteratively improve prompts to differentiate between normal-light and backlit images. 


%% file: Figures/Timeline.tex
\newcommand{\foo}{\hspace{-8.5pt}$\bullet$ \hspace{5pt}}
\newcommand{\splitlinefoo}{ \vspace{-4pt} \\ & \hspace{7pt} }
\newcommand{\newlinefoo}{\\ }

\begin{table*}[t]
\centering
\small

\renewcommand{\arraystretch}{1.3}
\begin{tabular}{r | l }

 2018 
 &  \foo Loh and Chan \cite{loh2019getting} seminal work in low-light image understanding; releases ExDark, \splitlinefoo an object detection and image recognition dataset (sRGB) \newlinefoo
 &  \foo Chen et al. \cite{chen2018learning} seminal work in low-light neural ISPs (RAW) \newlinefoo
 
 2019 
 &  \foo Afifi and Brown \cite{afifi2019else} show that white-balancing, as a part of image signal processing pipeline, \splitlinefoo can affect classification and segmentation drastically \newlinefoo
 &  \foo Yang et al. \cite{poor_visibility_benchmark} releases DARKFACE, low-light face recognition dataset (sRGB) \newlinefoo

 2020
 & \foo \rebuttal{ Ho et al. \cite{ho2020denoising} seminal work: Denoising Diffusion Probabilistic Models  (diffusion-based, sRGB) }\newlinefoo 
 &  \foo Guo et al. \cite{guo2020zero} formulates low-light enhancement as a curve parameter estimation problem \splitlinefoo zero-reference low-light image enhancement (sRGB) \newlinefoo
 &  \foo Mosleh et al. \cite{mosleh2020hardware} proposes hardware-in-the-loop ISP for task-based optimization (RAW) \newlinefoo

 2021
 &  \foo Radford et al. \cite{radford2021learning} releases the CLIP model \newlinefoo
 &  \foo Liang et al. \cite{liang2021recurrent} proposes a detection-with-enhancement method for low-light face detection (sRGB) \newlinefoo
 &  \foo Morawski et al. \cite{morawski2021nod} proposes a detection-with-enhancement method for low-light object detection and \splitlinefoo releases the NOD object detection dataset (sRGB) \newlinefoo
 &  \foo Hong et al. \cite{hong2021crafting} releases LOD, a low-light objected detection dataset and shows that \splitlinefoo  using raw sensor data can be more robust than sRGB in low-light object detection (sRGB, RAW) \newlinefoo

 2022
 &  \foo Zheng and Gupta \cite{zheng2022semantic} propose to guide zero-reference low-light \splitlinefoo enhancement unsupervised semantic segmentation (sRGB) \newlinefoo
 &  \foo Morawski et al. \cite{morawski2022genisp} propose a task-oriented neural ISP (RAW) \newlinefoo

 2023
 & \foo Wang et al. \cite{wang2023exploring} shows that CLIP can be used to capture the quality and feel of images \newlinefoo
  & \foo \cite{liang2023iterative} shows that CLIP can be used to discriminate between regions with different lighting; \splitlinefoo proposes prompt learning to capture image prior (sRGB) \newlinefoo
 &  \foo Wang et al. \cite{wang2023tienet} propose a detection-with-enhancement method for object detection (sRGB)  \newlinefoo
 &  \foo Hashmi et al. \cite{hashmi2023featenhancer} low-light task-oriented image enhancement (sRGB) \newlinefoo
 &  \foo Ljungbergh and Johnander \cite{ljungbergh2023raw} task-oriented neural ISP (RAW) \newlinefoo
 &  \foo Yoshimura et al. \cite{yoshimura2023dynamicisp} task-oriented neural ISP (RAW) \newlinefoo
 &  \foo Lee et al. \cite{ExLPose_2023_CVPR} releases ExLPose, a low-light human pose estimation dataset (sRGB) \newlinefoo
 &  \foo Zhang et al. \cite{zhang2023darkvision} releases DarkVision, a low-light video object detection dataset (sRGB) \newlinefoo
 &  \foo \rebuttal{Yi et al. \cite{yi2023diff} combines diffusion-based enhancement \& Retinex theory (diffusion-based, sRGB) }\newlinefoo
 
 2024
 & \foo \rebuttal{Wang et al. \cite{wang2024zero} proposes novel quadruple prior \& a zero-ref. enhancement method (diffusion-based, sRGB)} \\

\end{tabular}

\captionsetup{type=figure, font=footnotesize, labelsep=period, justification=raggedright, textfont=normalfont}
\caption{Timeline including recent developments in task-based optimization of low-light image enhancement methods, datasets dedicated to low-light image understanding, and seminal works important for the field.}

\label{fig:timeline}
\end{table*}

\vspace{10pt}
\vspace{1.2em}
\vspace{2.2em}

\begin{table*}[t]
\caption{\rebuttal{A short summary of advantages of raw and sRGB data in low-light enhancement -- alternative modalities proposed for optimization for machine cognition rather than human perception}.}
\centering

\footnotesize
\rebuttalcontainer
\begin{tabular}{r|c|l}
Aspect & sRGB vs RAW & Comment \\
\hline
Human perception & sRGB $>$ RAW & raw sensor data is not readily viewable and needs postprocessing for visualization \\
\hline
Low-light downstream-task performance & sRGB $\leq$ RAW & traditional ISPs prone to introducing irreversible errors in extreme low-light \\
Low-light perceptual performance & sRGB $\leq$ RAW & traditional ISPs prone to introducing irreversible errors in extreme low-light \\
Richness of information & sRGB $<$ RAW & raw files are linear w.r.t. scene irradiance, contain no ISP errors, \\
& & employ minimal denoising and have higher bit-depth \\ \hline
Dataset Availability & sRGB $>$ RAW & standard $>$ sensor-specific color space \\
Dataset Collection Ease & sRGB $>$ RAW & not all cameras allow saving raw files \\
Dataset Storage Ease & sRGB $>$ RAW & sRGB typically denoised \& compressed \\ \hline
Model Generalization & sRGB $>$ RAW & standard $>$ sensor-specific color space \\
Pre-Trained Models Availability & sRGB $>$ RAW & standard $>$ sensor-specific color space \\
\end{tabular}
\label{tab:table-advantages}
\end{table*}

%% file: Figures/Figure_Method.tex
\begin{figure*}[]
    \centering
    \small
    \includegraphics[trim={0cm 6cm 0cm 0cm},clip,width=1\linewidth]{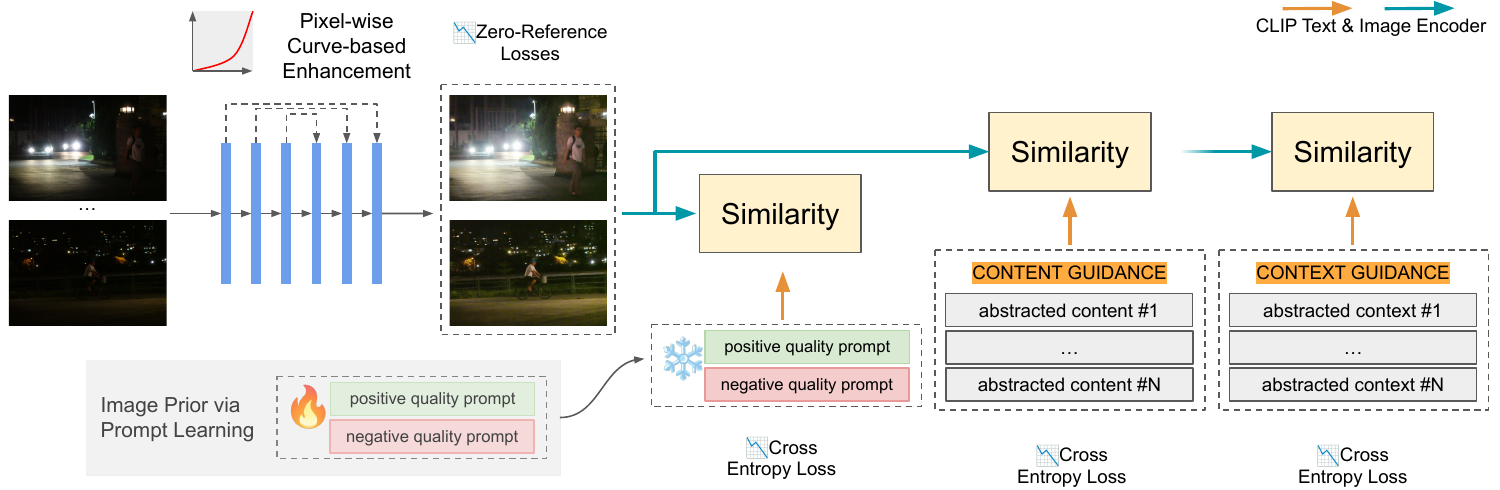}
    \caption{Our proposed method. In the first stage, we propose to learn the positive and negative image priors without any need for paired or unpaired normal-light data. Next, we train the image enhancement model, using zero-reference image losses, image prior prompts and semantic guidance. We propose to maximally use existing image annotation, by using cues about both the content and context of an image patch, that is, about the objects within and outside the patch.  \rebuttal{The two guidance steps work synergistically to improve the performance and, at the same time, only increase computational complexity at the training time, without incurring any additional costs during inference.} Our proposed method leverages the CLIP model and its zero-shot capabilities to scale favorably to include various datasets with any limitations on annotated object categories, instead of fixing the training category set.}
    \label{fig:proposed-method}
\end{figure*}

%% file: Figures/Figure_Guidance_Methods.tex
\begin{figure*}[]
    \centering
    \small
    \includegraphics[trim={0cm 5.75cm 0cm 0cm},clip,width=0.8\linewidth]{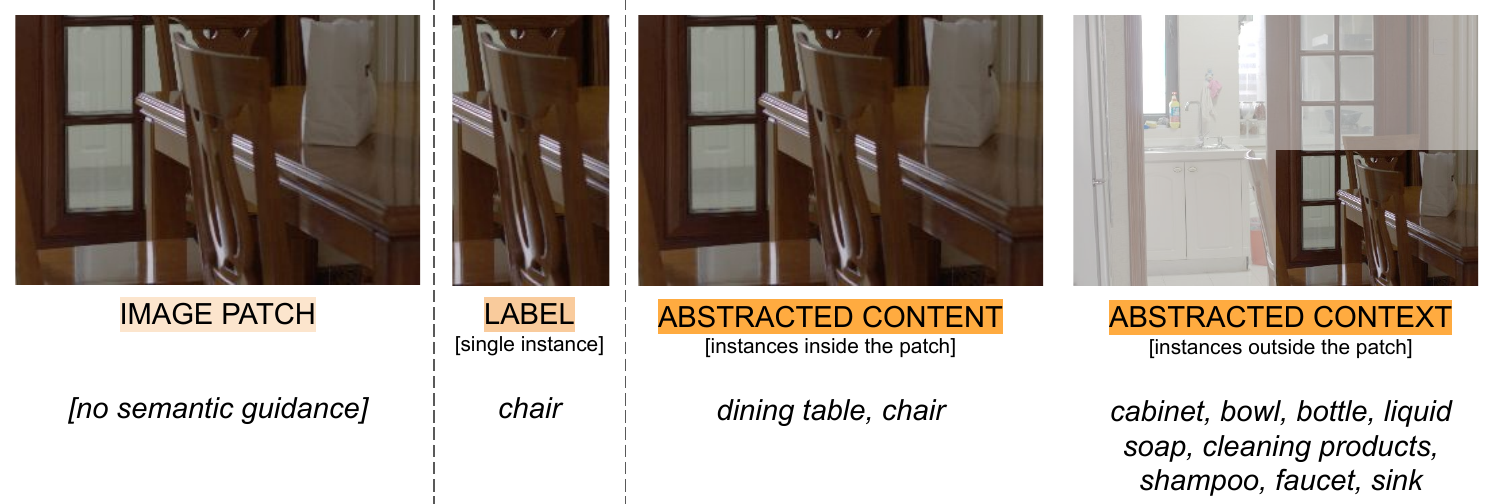}
    \caption{\rebuttal{Since our motivation is to make the maximal use of the existing low-light annotation, which may be difficult or costly to obtain, we apply two separate steps which use different information for semantic guidance}. Instead of realizing semantic guidance by training on object instances and corresponding labels, we propose to train on patches sampled from the image, together with descriptions of the scene within and outside the patch. In the first step, within a training batch, we match image description to image, and in the second step, similarly, we match descriptions of objects extending outside of the image to each image.}
    \label{fig:semantic-guidance-method}
\end{figure*}

%% file: Sections/3_Proposed_Method.tex
\section{Proposed Method}
\label{sec:proposed_method}
We propose to leverage the CLIP \cite{radford2021learning} model in a two-fold manner: first, we propose to learn an image prior from low-light data without any need for paired or unpaired normal-light data, and second, we propose to use the learned prior and additional semantic guidance to train the image enhancement model, in addition to regular zero-reference losses. We present the overview of the proposed method in Fig. \ref{fig:proposed-method}.

Motivated by the observation that the CLIP \cite{radford2021learning} encoder can accurately capture image quality as well as the abstract \textit{feel} of images \cite{wang2023exploring} and inspired by learning image prior via prompt learning introduced by \cite{liang2023iterative}, we propose to learn the image prior using an augmentation strategy based on image resampling, eliminating any need of well-lit data. Next, the learned prompt is used to guide the enhancement during the training stage, in addition to regular zero-reference losses. We experimentally show that the learned prior is effective at normalizing the overall image contrast and decreases over-amplification common in other zero-reference methods. 

Next, we propose to use existing \rebuttal{bounding box} or equivalent annotation to semantically guide the image enhancement model and enrich images in a way that benefits downstream task models. In order to maximally use the valuable annotation, we propose to use both annotation inside and outside image patches and thus integrate cues about the content and context of training images. We \rebuttal{conduct experiments to show} that the proposed method improves task-based performance of our method, and that introducing the context matching task improves the performance beyond only using the content cues.

\subsection{Unsupervised Image Prior via Prompt Learning}
\label{ssec:learnable-image-prior}
Leveraging the observation that the CLIP image prior captures diverse lighting conditions and image quality \cite{wang2023exploring}, we propose to use CLIP \cite{radford2021learning} to learn an image prior without any need for paired or unpaired normal-light data via prompt learning, to guide the image enhancement model in addition to zero-reference and semantic classification losses. To learn the image prior prompt pair, we use the pre-trained CLIP model \cite{radford2021learning} and a data augmentation strategy based on image resampling, as illustrated in Fig. \ref{fig:aug-visu}.

Given a low-light image $I \in \mathbb{R}^{H\times W \times C}$ and a pair of positive and negative prompts, $P_p \in \mathbb{R}^{N \times 512}$ and $P_n \in \mathbb{R}^{N \times 512}$, randomly initialized, where $N$ is the prompt length, we first apply random photometric augmentation, helping to avoid the prompts over-constraining the image prior. Next, we use the augmented low-light image $I'$ to synthesize a corresponding positive image $I_p$ and negative $I_n$ image pair. 

\input{Figures/Figures_Aug_Visu}

As illustrated in Fig. \ref{fig:aug-visu}, positive image $I_p = avg_{m \times m}(I')$ is generated by applying $m \times m$ average pooling, which acts as a fast and simple proxy for denoising, to the brightness-augmented image $I'$. On the other hand, negative image $I_n = sub_{1:m}(I')$  is generated by applying $1:m$ subsampling to the augmented image $I'$, preserving the noise in the image. 

As showed in Fig. \ref{fig:init-method}, we use the binary cross-entropy loss to learn the image prior prompt pair which will be later used to guide the image enhancement model by differentiating between the image quality. 

\input{Figures/Figure_Init_Method}

\begin{equation}
  L_{prompt\ init.} = -( y \log{ \hat{y} } + (1-y) \log{ (1-\hat{y}) } ),
  \label{eq:prompt-learning-ce}
\end{equation}

\noindent where $y$ is the image label, $y=0$ for a positive, $m \times m$ averaged image, and $y=1$ for a negative, $1:m$ subsampled image, and $\hat{y}$ is based on the softmax of cosine similarity between each of the prompts $P \in \{P_p, P_n\}$ and each of the images $I \in \{I_p, I_n\}$:

\begin{equation}
  \hat{y} = \frac
  {e^{cos(\Phi_{img}(I), \Phi_{txt}(P_p)}}
  {\sum_{i \in \{p, n\}} 
 e^{cos(\Phi_{img}(I), \Phi_{txt}(P_i)}},
  \label{eq:prompt-learning-cs}
\end{equation}

\noindent where $\Phi_{img}$ is the pre-trained CLIP image encoder and $\Phi_{txt}$ is the pretrained CLIP text encoder. 

We experimentally show that the proposed prompts help to guide the image enhancement model
by improving the overall image contrast, reducing under- and overexposure leading to decreased information loss, and reducing over-amplification of noise common in unsupervised enhancement models.

\input{Figures/Figure_Dataset_Levels}

\input{Figures/Figure_Dataset_Sources}

\subsection{Unsupervised Low-Light Enhancement}
Inspired by the success of ZDCE \cite{guo2020zero, li2021learning} in zero-reference low-light enhancement, we adopt the curve-based enhancement network architecture DCE-Net \cite{guo2020zero} as our baseline, along with the proposed zero-reference losses based on assumptions about natural images. The enhancement process, following Guo \textit{et al.} \cite{guo2020zero}, is formulated as a pixel-wise curve parameter prediction and is iteratively applied to the input image $I$:

\begin{equation}
  LE_n(x) = LE_{n-1}(x) + A_n (x) LE_{n-1}(x)(1-LE_{n-1}(x)),
  \label{eq:iterative-curve}
\end{equation}
\noindent where $A$ is a set of $N$ pixel-wise parameter maps $\alpha \in [-1, 1]$ applied iteratively in $N$ iterations to the input image $I$.

Focusing on leveraging the CLIP \cite{radford2021learning} model for semantic segmentation, we adopt the original zero-reference losses proposed by Guo \textit{et al.} \cite{guo2020zero} without any modification. We experimentally found that replacing curve estimation with direct end-to-end regression of the output image, with or without additional semantic guidance losses, results in the network producing mean expected exposure as set by exposure control loss. 

For completeness, we briefly discuss the employed zero-reference losses, along with their contribution to the image appearance:

\begin{equation}
\begin{split}
  L_{zero reference} = 
  \lambda_{exp} L_{exp} + 
  \lambda_{spa} L_{spa} + \\
  \lambda_{RGB} L_{RGB} + 
  \lambda_{TV_{A}} L_{TV_{A}},
  \label{eq:loss-total}
\end{split}
\end{equation}
\noindent where $L_{exp}$, $L_{spa}$, $L_{RGB}$, $L{TV_{A}}$ are the exposure, spatial consistency, color and illumination smoothness losses and $\lambda$ are the corresponding loss weights.

\textbf{Exposure loss} $L_{exp}$ controls the exposure of the image by constraining an expected average brightness $E$ of the local image region:

\begin{equation}
  L_{exp} = \frac{1}{M}
  \sum^{M}_{i=1} 
  \sum_{i=1} 
  |\hat{I}_i - E|,
  \label{eq:loss-exposure}
\end{equation}

\noindent where  $\hat{I}_{k}$ is the $k$-th $16 \times 16$ region of the output image, $M$ is the number of non-overlapping $16 \times 16$ patches in the image, and $E$ is the expected average brightness. $E$ is set to $0.6$, empirically.

\textbf{Spatial consistency loss} $L_{spa}$ encourages spatial coherence of the output image by preserving relations between adjacent image regions in terms of relative differences:

\begin{equation}
  L_{spa} = \frac{1}{K}
  \sum^{K}_{i=1} 
  \sum_{j \in \Omega (i)} 
  ( |\hat{I}_{i} - \hat{I}_{j}| -
  |{I}_{i} - {I}_{j}| )^2,
  \label{eq:loss-spatial}
\end{equation}

\noindent where  ${I}$ and $\hat{I}$ are the original and output images, respectively, $K$ is the number of local image regions, and $\Omega (i)$ denotes the 4-connected neighborhood of the $4 \times 4$ region $i$ center.

\textbf{Color loss} $L_{RGB}$ controls the overall image hue, inspired by the Gray-World hypothesis \cite{buchsbaum1980spatial}, by constraining relative differences between color channels:

\begin{equation}
  L_{RGB} = (\hat{I}_{R,\mu} - \hat{I}_{G,\mu})^2 + (\hat{I}_{R,\mu} - \hat{I}_{B,\mu})^2 + (\hat{I}_{G,\mu} - \hat{I}_{B,\mu})^2
  \label{eq:loss-color}
\end{equation}

\noindent where $\hat{I}_{c,\mu}$ is the average brightness of color channel $c \in \{R,G,B\}$ of the output image $\hat{I}$.

\textbf{Illumination smoothness loss} $L_{TV_{A}}$ is the TV-based smoothness constrain on the intermediate curve parameter map $A$ and encourages the neighboring parameters to have a monotonicity relation:

\begin{equation}
  L_{TV_{A}} =
  \sum_{c\in\{R,G,B\}}
  (|\nabla_x A_c| + |\nabla_y A_c|)^2,
  \label{eq:loss-tv}
\end{equation}

\noindent where $A_{c}$ is the curve parameter, corresponding to each of the color channels $c$ of the image, applied to enhance the image $\hat{I}$, and $\nabla$ is the gradient operation.

\input{Tables/Tab_Ablation_Label}

\subsubsection{Leveraging CLIP for Image Enhancement}
\label{sssec:leveraging-clip-training}

We leverage the CLIP model \cite{radford2021learning} at the training stage to guide the enhancement model in a two-fold manner: we use the learned CLIP \cite{radford2021learning} prompt pair, which represents image prior, and we use to perform semantic guidance at a loss level, via zero-shot open vocabulary classification.

\textbf{Learned CLIP Image Priors}. We use the learned CLIP image prior, trained as described in subsec. \ref{ssec:learnable-image-prior}, to further constrain the enhanced image $\hat{I}$. We experimentally show that using a prompt pair learned in this way helps to improve the overall image contrast and significantly reduce over-amplification of image noise by better discriminating between foreground and background regions of images.

To this end, we first embed the enhanced image $\hat{I}$ with the CLIP image encoder $\Phi_{img}$ and the learned prompt pair with the CLIP text encoder $\Phi_{txt}$. Next, we calculate the cosine similarity between the encoded enhanced image $\Phi_{img}(\hat{I})$ and the learned prompt pair $\Phi_{txt}(P)$, where $P={P_p,P_n}$ is a pair of the positive and negative prompts, described in subsec. \ref{ssec:learnable-image-prior}. 

\begin{equation}
  \hat{y}_{prompt} = \frac
  {e^{cos(\Phi_{img}(\hat{I}), \Phi_{txt}(P_p)}}
  {\sum_{i \in \{p, n\}} 
 e^{cos(\Phi_{img}(\hat{I}). \Phi_{txt}(P_i)}}.
  \label{eqprior-cs}
\end{equation}

\noindent The final loss function is formulated as the cross-entropy loss, based on $\hat{y}$.  

\textbf{Content- and Context-based Semantic Guidance}. We propose to use \rebuttal{bounding box} or equivalent annotation to semantically guide the image enhancement model, illustrated in Fig. \ref{fig:semantic-guidance-method}, enriching the images with features relevant to machine cognition, rather than the human perception. To maximize the utility of the valuable low-light annotation, we use the annotation both within and outside the patch to incorporate cues about the content of the patch and its context, respectively. To this end, we use the CLIP\cite{radford2021learning} model and formulate the proposed semantic guidance as a classification problem.

First, we convert the existing \rebuttal{bounding box} or equivalent annotation inside and outside the patch to abstracted content and context descriptions by itemizing each of the object instances, separated by commas. In our pilot study, based on the pre-trained CLIP \cite{radford2021learning} and normal-light images, we experimentally found that the order of itemization has a minimal level on the accuracy when using CLIP \cite{radford2021learning} to match images with their abstracted content descriptions. 

Next, given a batch of enhanced $N$ low-light images $\bm{\hat{I}} = \{\hat{I}_i\}$ and the corresponding set of abstracted descriptions $\bm{Y} = \{Y_i\}$, we encode the image batch using the CLIP \cite{radford2021learning} image encoder $\Phi_{img}$ and the descriptions using the CLIP text encoder $\Phi_{txt}$. For brevity, we use \textit{abstracted descriptions} to mean either content or context descriptions, computed in two separate identical steps. Next, we compute the similarity between each of the images and descriptions as follows:

\begin{equation}
  s_{i,j} = {cos(\Phi_{img}(\hat{I_i}), \Phi_{txt}(A_j)},
  \label{eqsem-sim}
\end{equation}

\noindent and use the $N \times N$ computed similarity tensor to compute two cross-entropy loss terms in the two dimensions, corresponding to two tasks: 1) given an image $\hat{I}_i$, selecting a corresponding description $Y_i$ from the set $\bm{Y} = \{Y_j\}, j=0,1,...,N$, and 2) given a description $\hat{Y}_i$,  selecting a corresponding image $\bm{I} = \{I_j\}, j=0,1,...,N$.

%% file: Figures/Figures_Aug_Visu.tex
\begin{figure}[]
    \centering
    \small
    \includegraphics[trim={0cm 7.5cm 16cm 0cm},clip,width=1\linewidth]{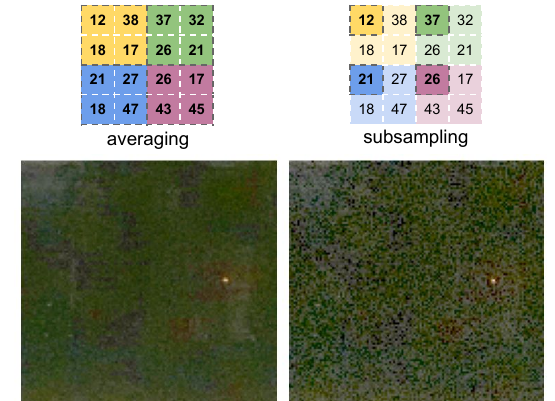}
    \caption{We use image sampling to augment positive and negative image prompt pair. The positive sample (on the left) is augmented using $4 \times 4$ averaging, acting as a fast and simple proxy for denoising, and the negative sample (on the right) is augmented using $1:4$ subsampling, preserving the noise in the image. Later, we use the learned prompt pair to help guide the enhancement model, leading to improved image contrast, reduced under- and overexposure, and reduced over-amplification of noise.}
    \label{fig:aug-visu}
\end{figure}

%% file: Figures/Figure_Init_Method.tex
\begin{figure}[]
    \centering
    \small
    \includegraphics[trim={0cm 5.5cm 15cm 0cm},clip,width=1\linewidth]{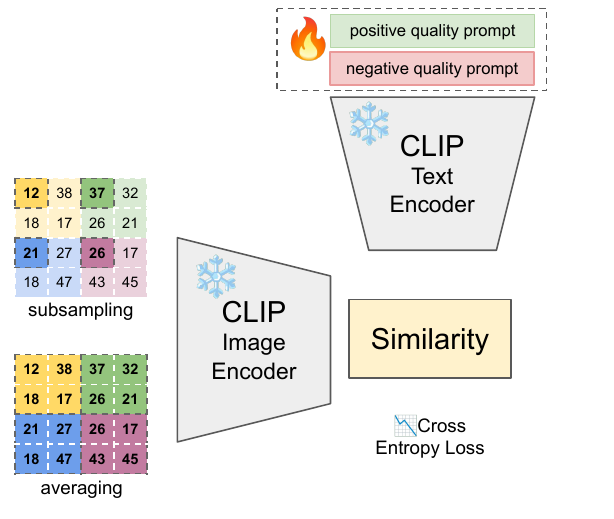}
    \caption{We first use the CLIP model to learn the positive and negative image prior pair using simple data augmentation strategy, based on image resampling, eliminating any need for paired or unpaired normal-light data. The learned prompt pair is later used for guiding the enhancement model. We experimentally show that the proposed prompts help to guide the image enhancement model by improving the overall image contrast, reducing under- and overexposure leading to decreased information loss, and reducing over-amplification of noise common in unsupervised enhancement models. }
    \label{fig:init-method}
\end{figure}

%% file: Figures/Figure_Dataset_Levels.tex
\begin{figure}[]
    \centering
    \small
    \includegraphics[trim={0cm 0cm 0cm 0cm},clip,width=1\linewidth]{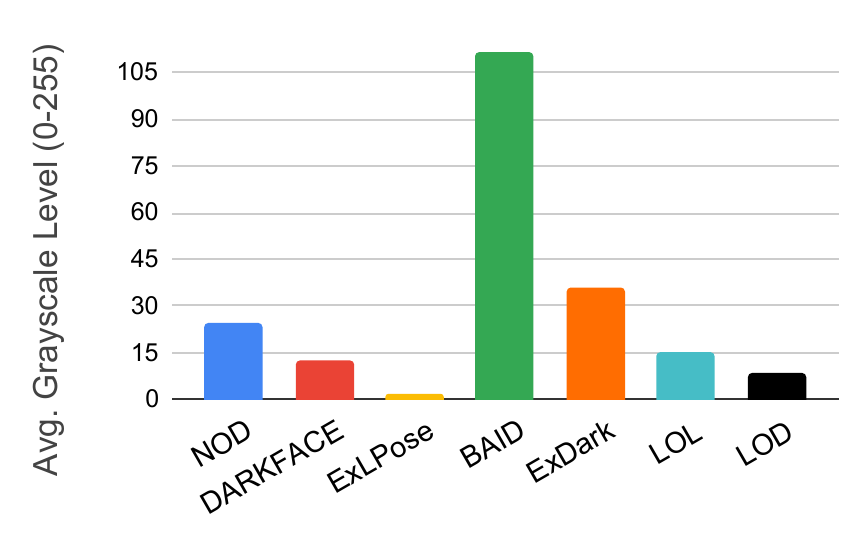}
    \caption{Average image pixel intensities in our \rebuttal{training} dataset, by sample source.}
    \label{fig:dataset-statistics}
\end{figure}

%% file: Figures/Figure_Dataset_Sources.tex
\begin{figure}[]
    \centering
    \small
    \includegraphics[trim={0cm 0cm 0cm 0cm},clip,width=1\linewidth]{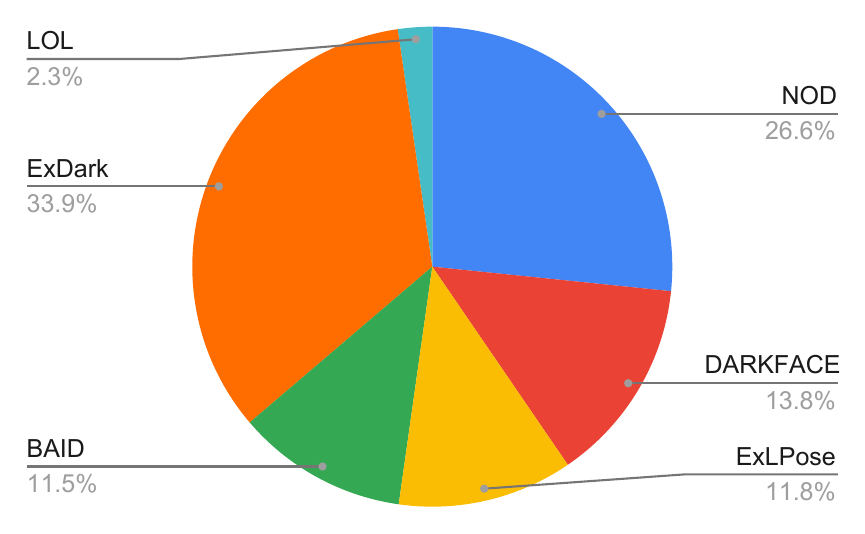}
    \caption{Sources of samples in our \rebuttal{training} dataset.}
    \label{fig:sources-samples}
\end{figure}

%% file: Tables/Tab_Ablation_Label.tex

\begin{table*}[t]
    \caption{Task-based evaluation for ablation study of the proposed methods for semantic guidance in image enhancement, trained on the LOL dataset \cite{wei2018deep}. The models in the experiment are implemented as the baseline model \cite{guo2020zero} using zero-reference image losses + semantic guidance method}
    \label{tab:ablation-label}
    \centering
    \footnotesize
    \begin{tabular}{c|cccccc}
    \hline
    
     Semantic guidance & NOD \cite{morawski2021nod} & NOD SE \cite{morawski2021nod} &  LOD \cite{hong2021crafting} & ExDark \cite{loh2019getting} & ExDark \cite{loh2019getting} & DarkFace \cite{poor_visibility_benchmark} \\
    
     method & mAP $\uparrow$ & mAP $\uparrow$ & mAP $\uparrow$ & mAP $\uparrow$ & class. acc. $\uparrow$ & mAP@.5 $\uparrow$ \\ \hline
     
    
     baseline \cite{guo2020zero} (none) & 41.2\% & 22.4\% & 40.9\% & 34.1\% & 43.7\% & 35.6\% \\

     bounding box label \cite{morawski2024unsupervised} & 41.1\% & 22.5\% & 42.5\% & 34.5\% & 41.2\% & 28.6\% \\
    
     abstracted content & 41.5\% & 23.4\% & 41.8\% & 34.8\% & 44.2\% & 37.1\% \\
     
     abstracted context & 41.4\% & 23.6\% & 41.4\% & 34.7\% & 43.7\% & 36.9\% \\
     
     abstracted content + context & 41.8\% & 24.2\% & 42.9\% & 35.0\% & 44.5\% & 37.4\% \\

    \hline
    \end{tabular}

    \end{table*}

%% file: Sections/4_Experimental_Results.tex
\section{Experimental Results and Discussion}
\label{sec:experimental_results}

\input{Figures/Figure_Ablation_Qual}
\input{Tables/Tab_Ablation}

Following the related work \cite{morawski2021nod,morawski2022genisp,wang2023tienet,hashmi2023featenhancer,robidoux2021end,yoshimura2023dynamicisp}, we focus on task-based evaluation of our proposed and related methods, rather than assuming correlation between the human perception and downstream task performance.

\subsection{Implementation Details}
\label{ssec:experimental_results_implementation_details}
\let\thefootnote\relax\footnotetext{* Experiments were completed by authors at National Taiwan University.} 
In all our experiments, we use Zero-DCE \cite{guo2020zero} along with original zero-reference losses as  the baseline. We use pre-trained CLIP \cite{radford2021learning} to learn image prior prompts, using prompt learning with prompts of length 16, and augmenting data using photometric (brightness, contrast, hue) augmentation before sampling images. We use scale factor $s=4$ to sample the images, selecting every 4th pixel for the negative prompt and averaging \rebuttal{$4 \times 4$} blocks for the positive prompt. During the training of the enhancement model, we extract quadrants of images (bottom-left, top-left, top-right and bottom-right parts of the image) as input images and corresponding annotation of the objects inside and outside the patch. We fine-tune the projection layer of the CLIP \cite{radford2021learning} model for the content and context classification separately and use the fine-tuned weights for the guidance later.

Our method uses annotated low-light datasets, such as low-light object detection datasets. Still, it is easily extendable to paired low- and normal-light datasets by automatic annotation using open-vocabulary object detectors, increasing the amount of valuable low-light data for the training. For paired low- and normal-light datasets without manual annotation, we ran \cite{cheng2024yolo}, an open-vocabulary detector, providing 365 labels from the Objects365 \cite{shao2019objects365} dataset, and used predictions with a confidence score over 30\%.

For training, We use input size $224 \times 224$, batch size $8$, $200$ epochs, the Adam optimizer, learning rate $0.0001$, weight decay $0.0001$ and gradient norm clipping $0.1$. We set the weights of the color loss term to $5$, the spatial consistency term to $1$, the exposure control term to $10$, and the TV term $200$.

Unless noted otherwise, we use a collection of data extracted from the NOD \cite{morawski2021nod}, ExDark \cite{loh2019getting}, DarkFace \cite{poor_visibility_benchmark}, ExLPose \cite{ExLPose_2023_CVPR}, LOL \cite{wei2018deep} and BAID \cite{lv2022backlitnet} datasets  \rebuttal{for training. Basic statistics of our training dataset are shown in Fig. \ref{fig:dataset-statistics} and Fig. \ref{fig:sources-samples}}. For task-based evaluation, we use an open-vocabulary YOLO-World \cite{cheng2024yolo} detector for object detection experiments and YOLO5Face \cite{YOLO5Face} for face detection. 

\subsection{Ablation Study} 

\label{ssec:experimental_results_ablation_study}
We conduct an ablation study showing the effectiveness of each of the proposed loss terms across multiple task-based benchmarks in Tab. \ref{tab:ablation}. Our proposed image prior leads to consistent improvements above the zero-reference baseline method \cite{guo2020zero}, as well as performance without enhancement. The exception is the mAP on the ExDark \cite{loh2019getting} performance, where our method performs similarly to no enhancement. However, our method, built on the baseline method \cite{guo2020zero}, largely eliminates errors of the baseline method \cite{guo2020zero} made on this dataset.

Next, we visualize images enhanced by each of the studied variants in Fig. \ref{fig:ablation-qualitative}. The learned prompt improves the overall image contrast, helps reducing under- and over-exposure and reduces over-amplification of the noisy regions, as seen in the first and last rows. Semantic guidance using content and context cues improves the overall color distribution of the images. When used together, content and context guidance cues synergistically work to improve the accuracy of the colors based on semantic features, as seen by the improvement of the color of the sky in the second row of the figure.

\subsection{Comparison of Guidance Methods} 
We conduct a study to compare semantic guidance methods based on \rebuttal{bounding box} annotation, as in Fig. \ref{fig:semantic-guidance-method}, presented in Tab. \ref{tab:ablation-label}.  While the semantic guidance using bounding box labels leads to mixed performance impact across the datasets, the abstracted content and context guidance lead to consistent improvements over the baseline \cite{guo2020zero} method across all datasets. Moreover, the proposed content and context methods can work synergistically to further improve the performance, maximizing the utility of the valuable low-light annotation.

\input{Figures/Figure_Comparison_Qual}
\input{Tables/Tab_Comparison}
\input{Tables/Tab_Comparison_PSNR}

\input{Tables/Tab_Comparison_LOD}

\input{Figures/Figures_Low_Light_Impact}

\subsection{Quantitative Comparison with Related Methods}
\label{ssec:quant-comparison}
We conduct a study to compare our proposed method with related work, presented in Tab. \ref{tab:comparison} and Tab. \ref{tab:comparison-psnr}. As for the method selection, we selected zero-reference methods for low-light image enhancement, RUAS \cite{liu2021retinex}, SGZ \cite{zheng2022semantic}, Zero-DCE \cite{guo2020zero} and SCI \cite{ma2022toward}. Direct comparison with other methods that use normal-light data was not possible because we fine-tuned each of the methods using the dataset reported in Subsec. \ref{ssec:experimental_results_implementation_details}, with data extracted from the NOD \cite{morawski2021nod}, ExDark \cite{loh2019getting}, DarkFace \cite{poor_visibility_benchmark}, ExLPose \cite{ExLPose_2023_CVPR}, LOL \cite{wei2018deep} and BAID \cite{lv2022backlitnet} datasets.

As seen in Tab. \ref{tab:comparison}, our method leads to favorable results in terms of task-based performance across the variety of tested datasets. Furthermore, as we used Zero-DCE \cite{guo2020zero} as our baseline method, we followed \cite{li2021learning} and used down- and up-sampling to reduce the model complexity at the inference time. As seen in Tab. \ref{tab:comparison}, the computational complexity can be decreased by three orders of magnitude without significantly impacting task-based performance.  Full reference quality assessment using the challenging VE-LOL \cite{liu2021benchmarking} dataset, as seen in Tab. \ref{tab:comparison-psnr}, using PSNR and SSIM metrics, leads to a similar conclusion.

\input{Tables/Tab_crossdataset_comparison}

\rebuttal{Additionally, we investigate how our method compares to related methods that use paired or unpaired low- and normal-light data during the training. We present the results of this experiment, in terms of task-based performance in Tab. \ref{tab:crossdataset_comparison} and in terms of PSNR and SSIM in Tab. \ref{tab:cross-comparison-psnr}.  In contrast with zero-reference methods reported in Tab. \ref{tab:comparison} fine-tuned using  the dataset reported in Subsec. \ref{ssec:experimental_results_implementation_details}, all methods in Tab. \ref{tab:crossdataset_comparison} and Tab. \ref{tab:cross-comparison-psnr} require paired or unpaired low- and normal-light data. Therefore, as direct fine-tuning was not possible, we use original publicly available checkpoints. Such cross-dataset setup was also reported in recent works, such as \cite{wang2024zero}. }

\rebuttal{We observe that our proposed method ranks first or second across all tested sets. At the same, time the computational complexity at the inference time is significantly lower than the majority of other methods.}

\input{Tables/Tab_cross-comparison-psnr}

\rebuttal{Further, comparing Tab. \ref{tab:comparison-psnr} and Tab.   \ref{tab:cross-comparison-psnr}, we observe that in our experiments, paired data methods perform significantly better than unpaired, unsupervised and zero-reference methods in terms of PSNR and SSIM measuring full-reference image restoration performance. }

\rebuttal{Importantly, we observe that higher restoration performance in our experiments does not translate to improved downstream task performance in Tab. \ref{tab:crossdataset_comparison}, which is the main objective of our paper. In fact, in Tab. \ref{tab:crossdataset_comparison}, unpaired and unsupervised methods outperform paired data methods. This observed performance trade-off leads to an important conclusion that, in low-light image enhancement, a correlation between restoration quality and machine cognition performance cannot be assumed.}

\subsection{Qualitative Comparison with Related Methods}

\input{Figures/Figure_comparisonbbox}

\rebuttal{Further, we visualise the results of object detection in Fig. \ref{fig:comparisonbbox} on samples from the NOD \cite{morawski2021nod} dataset. We observe that the performance gap seen in Tab. \ref{tab:comparison} can be attributed to detection noise (category misattribution) seen in extreme low-light samples in Fig. \ref{fig:comparisonbbox}. Furthermore, we observe that over-exposure of bright images can lead to spurious responses in detection, as seen in the last row (RUAS  \cite{liu2021retinex}) of Fig. \ref{fig:comparisonbbox}.}

Next, we investigate qualitative differences between the studied methods, presented in Fig. \ref{fig:comparison-qualitative}. In comparison with RUAS \cite{liu2021retinex}, SGZ \cite{zheng2022semantic}, and Zero-DCE \cite{guo2020zero}, our method leads to better discrimination between background and foreground and reduction of over-enhancement in bright areas. The differences between our method and SCI \cite{ma2022toward} are more subtle, with the most significant difference being the vibrance of semantically important regions, such as the sky in the second row.

\subsection{Impact of Low Light on Object Detection}
We conduct a study to quantify impact of low light on object detection. To this end, we use the LOD dataset \cite{hong2021crafting} of paired low- and normal-light data. As we only have access to JPEG images and it is not possible to revert the image signal processing pipeline on this dataset, we use $\alpha$-blending as a proxy for simulating different levels of low light instead. As seen in the Fig. \ref{fig:aug-lowlight-impact}, as the levels of proportion of low-light data in the $\alpha$-blended input images increases, the performance of the the detector is impacted negatively. When using our method to enhance the images before detection, the impact is lessened, and the performance as measured in mAP is higher than the no enhancement baseline at all levels. We further present lower and higher upper bounds for object detection using the LOD dataset \cite{hong2021crafting} in Tab. \ref{tab:comparison-LOD}, including comparison with related methods. As evident from the Fig. \ref{fig:aug-lowlight-impact} and Tab. \ref{tab:comparison-LOD}, low light conditions reduce the amount of information in images, impacting the detector negatively. Although using our approach does not eliminate this problem completely, the performance is improved at all levels in Fig. \ref{fig:aug-lowlight-impact}. The proposed approach results in minimal complexity increases, when paired with inference-time down- and up-scaling of the parameter map, as presented in Tab. \ref{tab:comparison}, namely 0.08GMACs increase over 225.53GMACs of the no enhancement baseline using YOLO-World \cite{cheng2024yolo}.

\subsection{\rebuttal{Generalization to Other Baseline Models}}
\input{Tables/Tab_alternative_baselines}

\rebuttal{In order to verify the generalizability of our proposed method, we apply our training strategy to related zero-reference methods, presented in Tab. \ref{tab:alternative_baselines}. To this end, we apply proposed content- and context-based guidance and our proposed learned CLIP image prior in addition to the loss function originally used by each of the investigated methods: RUAS \cite{liu2021retinex}, SGZ \cite{zheng2022semantic}, SCI \cite{ma2022toward} and Zero-DCE \cite{guo2020zero}. We observe significant improvement in all methods with exception of SCI \cite{ma2022toward}.}

\rebuttal{We hypothesize that this is because SCI \cite{ma2022toward} introduces a self-calibration module into the learning framework, facilitating the convergence between intermediate results of each stage of the cascaded illumination learning. Possibly, addition of semantic guidance interferes with the training dynamics dictated by the original training objective, leading to sub-optimal results. }

%% file: Figures/Figure_Ablation_Qual.tex
\begin{figure*}[t]
    \captionsetup[subfigure]{justification=centering}
    \centering
    \small
    \begin{subfigure}{.14\textwidth}
      \centering
        \includegraphics[width=1\linewidth]{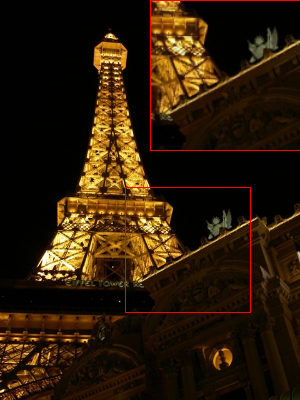}
        \caption*{} 
    \end{subfigure} %
    \begin{subfigure}{.14\textwidth}
      \centering
        \includegraphics[width=1\linewidth]{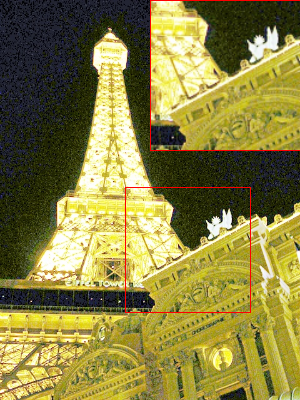}
        \caption*{}
    \end{subfigure} %
    \begin{subfigure}{.14\textwidth}
      \centering
        \includegraphics[width=1\linewidth]{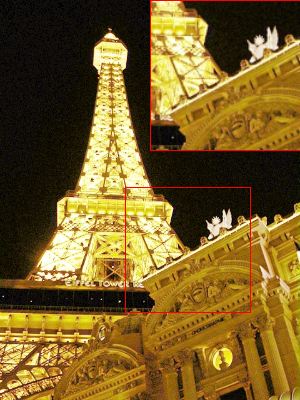}
        \caption*{}
    \end{subfigure} %
    \begin{subfigure}{.14\textwidth}
      \centering
        \includegraphics[width=1\linewidth]{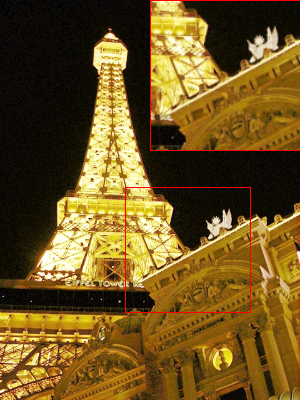}
        \caption*{}
    \end{subfigure} 
    \begin{subfigure}{.14\textwidth}
      \centering
        \includegraphics[width=1\linewidth]{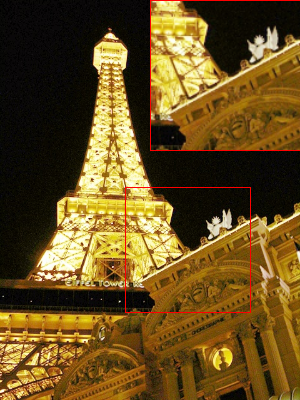}
        \caption*{}
    \end{subfigure} 
    \begin{subfigure}{.14\textwidth}
      \centering
        \includegraphics[width=1\linewidth]{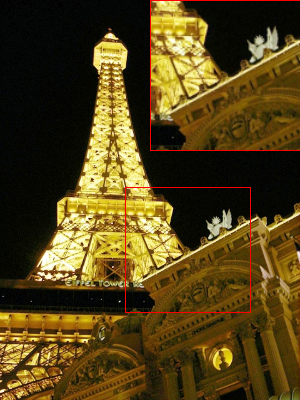}
        \caption*{}
    \end{subfigure} 

\vspace{-1.2em}

    \begin{subfigure}{.14\textwidth}
      \centering
        \includegraphics[width=1\linewidth]{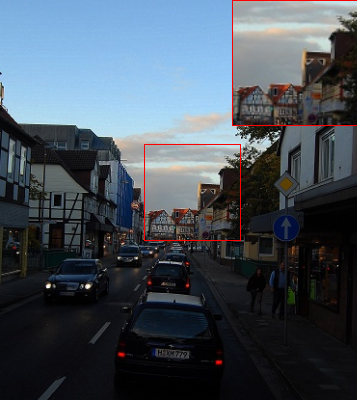}
        \caption*{} 
    \end{subfigure} %
    \begin{subfigure}{.14\textwidth}
      \centering
        \includegraphics[width=1\linewidth]{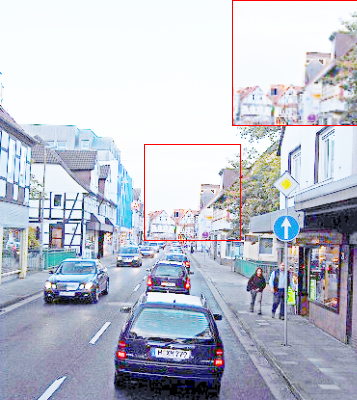}
        \caption*{}
    \end{subfigure} %
    \begin{subfigure}{.14\textwidth}
      \centering
        \includegraphics[width=1\linewidth]{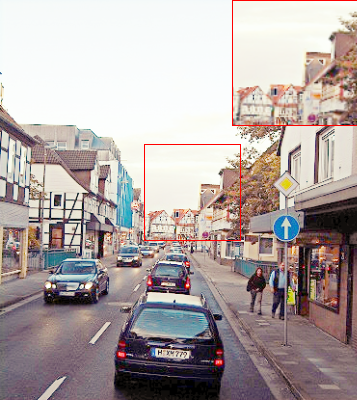}
        \caption*{}
    \end{subfigure} %
    \begin{subfigure}{.14\textwidth}
      \centering
        \includegraphics[width=1\linewidth]{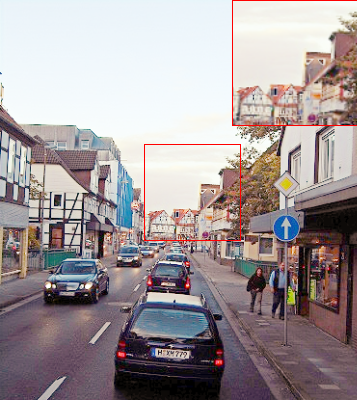}
        \caption*{}
    \end{subfigure} 
    \begin{subfigure}{.14\textwidth}
      \centering
        \includegraphics[width=1\linewidth]{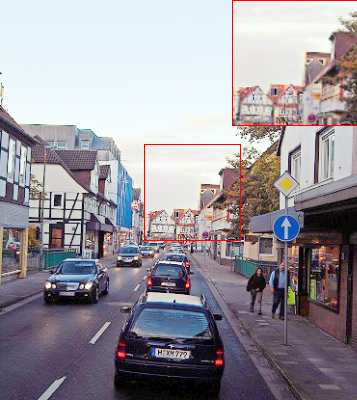}
        \caption*{}
    \end{subfigure} 
    \begin{subfigure}{.14\textwidth}
      \centering
        \includegraphics[width=1\linewidth]{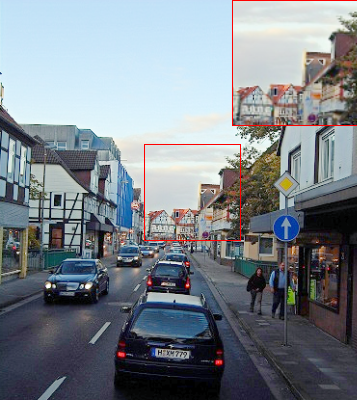}
        \caption*{}
    \end{subfigure} 

\vspace{-1.2em}

    \begin{subfigure}{.14\textwidth}
      \centering
        \includegraphics[width=1\linewidth]{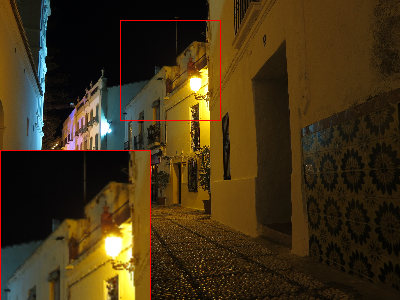}
        \caption*{Input \\ \phantom{ } } 
    \end{subfigure} %
    \begin{subfigure}{.14\textwidth}
      \centering
        \includegraphics[width=1\linewidth]{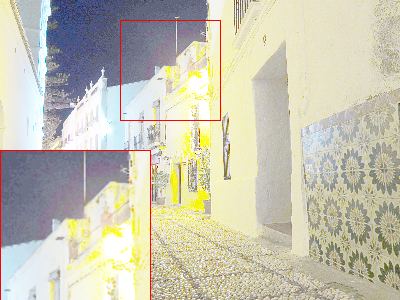}
        \caption*{Baseline \cite{guo2020zero} \\   \phantom{ } }
    \end{subfigure} %
    \begin{subfigure}{.14\textwidth}
      \centering
        \includegraphics[width=1\linewidth]{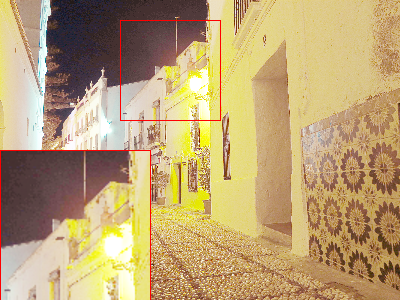}
        \caption*{+ learned \\ \rebuttal{img.} prior}
    \end{subfigure} %
    \begin{subfigure}{.14\textwidth}
      \centering
        \includegraphics[width=1\linewidth]{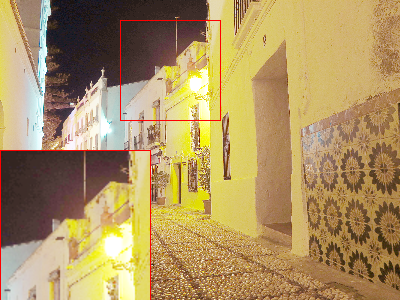}
        \caption*{+ img. prior \\ + content}
    \end{subfigure} 
    \begin{subfigure}{.14\textwidth}
      \centering
        \includegraphics[width=1\linewidth]{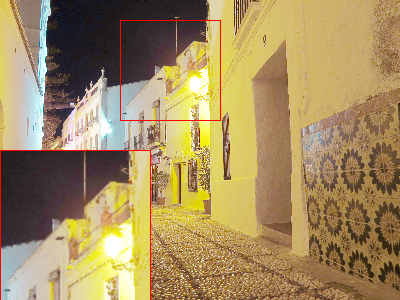}
        \caption*{+ img. prior \\ + context}
    \end{subfigure} 
    \begin{subfigure}{.14\textwidth}
      \centering
        \includegraphics[width=1\linewidth]{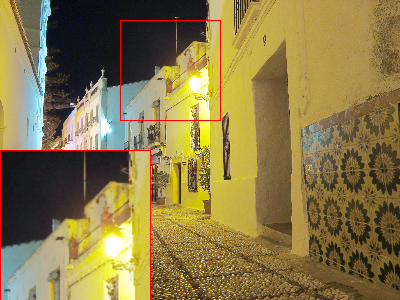}
        \caption*{+ img. prior \\ + content + context}
    \end{subfigure} 

    \caption{Ablation study of our proposed method. The learned prompt improves the overall image contrast, helps reducing under- and overexposure and reduces over-amplification of the noisy regions. Semantic guidance using content and context cues improves the overall color distribution of the images. } 
    \label{fig:ablation-qualitative}
\end{figure*}

%% file: Tables/Tab_Ablation.tex

\begin{table*}[t]
\caption{Quantitative results of the ablation study of our proposed method in terms of task-based performance. Our proposed improvements leads to consistent improvement over the baseline method}
\label{tab:ablation}

\centering
\scriptsize
\begin{tabular}{ccccc|cccccc}
\hline

Zero-ref. & Learned & Label & Context & Content  & NOD \cite{morawski2021nod} & NOD SE \cite{morawski2021nod} &  LOD \cite{hong2021crafting} & ExDark \cite{loh2019getting} & ExDark \cite{loh2019getting} & DarkFace \cite{poor_visibility_benchmark} \\

baseline \cite{guo2020zero} & img. prior & guidance \cite{morawski2024unsupervised} & guidance & guidance & mAP $\uparrow$ & mAP $\uparrow$ & mAP $\uparrow$ & mAP $\uparrow$ & class. acc. $\uparrow$ & mAP@.5 $\uparrow$ \\ \hline

\multicolumn{5}{c|}{(no enhancement)} & 42.1\% & 22.8\% & 37.8\% & 40.3\% & 22.8\% & 33.3\%  \\

\checkmark & & & & & 41.2\% & 22.8\% & 41.1\% & 34.0\% & 44.1\% & 35.9\% \\

 \checkmark & \checkmark & & & &  42.7\% & 26.0\% & 45.5\% & 37.7\% & 47.0\% & 37.1\% \\

 \checkmark & \checkmark & \checkmark & & &  43.2\% & 26.5\% & 47.4\% & 39.7\% & 43.6\% & 38.0\% \\

 \checkmark & \checkmark &  & \checkmark & & 43.0\% & 25.7\% & 46.0\% & 38.4\% & 47.0\% & 38.7\% \\

 \checkmark & \checkmark &  & & \checkmark & 43.2\% & 26.2\% & 46.7\% & 38.9\% & 46.8\% & 40.1\% \\
 
 \checkmark & \checkmark & & \checkmark & \checkmark & 43.9\% & 27.0\% & 47.1\% & 39.9\% & 42.5\% & 42.1\% \\

\hline
\end{tabular}

\end{table*}

%% file: Figures/Figure_Comparison_Qual.tex
\begin{figure*}[t]
    \captionsetup[subfigure]{justification=centering}
    \centering
    \small
    \begin{subfigure}{.14\textwidth}
      \centering
        \includegraphics[width=1\linewidth]{Images/comparison-figures/CROP_000000_ORIG_DICM_01.png}
        \caption*{} 
    \end{subfigure} %
    \begin{subfigure}{.14\textwidth}
      \centering
        \includegraphics[width=1\linewidth]{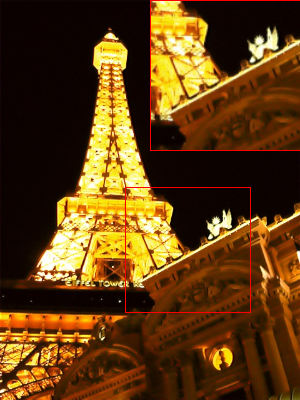}
        \caption*{}
    \end{subfigure} %
    \begin{subfigure}{.14\textwidth}
      \centering
        \includegraphics[width=1\linewidth]{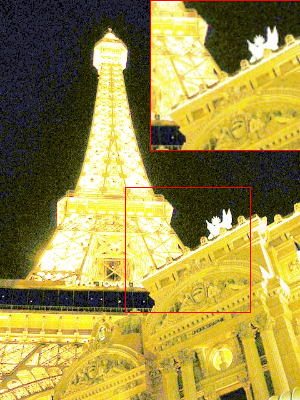}
        \caption*{}
    \end{subfigure} %
    \begin{subfigure}{.14\textwidth}
      \centering
        \includegraphics[width=1\linewidth]{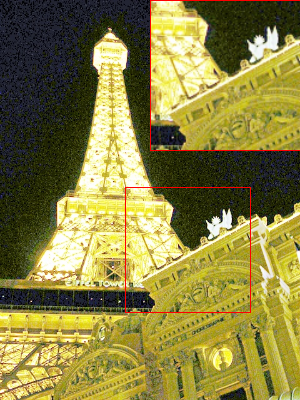}
        \caption*{}
    \end{subfigure} 
    \begin{subfigure}{.14\textwidth}
      \centering
        \includegraphics[width=1\linewidth]{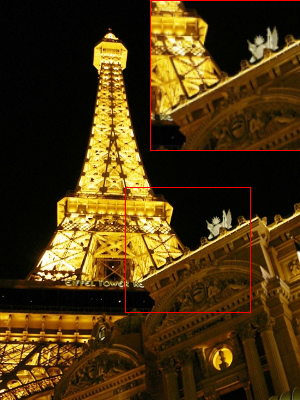}
        \caption*{}
    \end{subfigure} 
    \begin{subfigure}{.14\textwidth}
      \centering
        \includegraphics[width=1\linewidth]{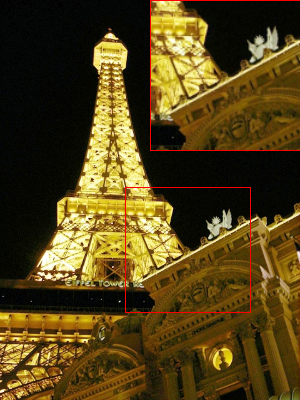}
        \caption*{}
    \end{subfigure} 

\vspace{-1.2em}

    \begin{subfigure}{.14\textwidth}
      \centering
        \includegraphics[width=1\linewidth]{Images/comparison-figures/CROP_000000_ORIG_LIME_4.png}
        \caption*{} 
    \end{subfigure} %
    \begin{subfigure}{.14\textwidth}
      \centering
        \includegraphics[width=1\linewidth]{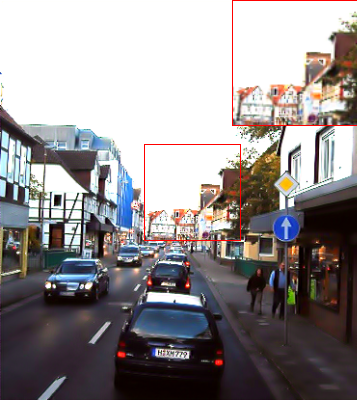}
        \caption*{}
    \end{subfigure} %
    \begin{subfigure}{.14\textwidth}
      \centering
        \includegraphics[width=1\linewidth]{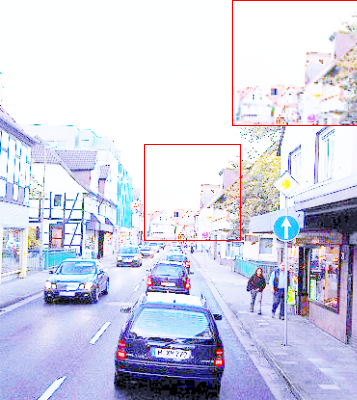}
        \caption*{}
    \end{subfigure} %
    \begin{subfigure}{.14\textwidth}
      \centering
        \includegraphics[width=1\linewidth]{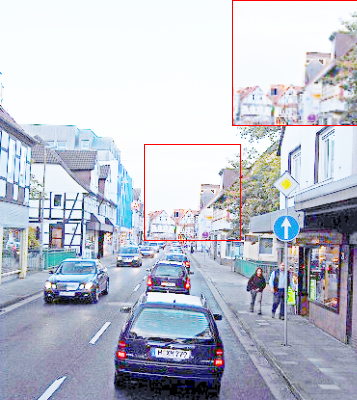}
        \caption*{}
    \end{subfigure} 
    \begin{subfigure}{.14\textwidth}
      \centering
        \includegraphics[width=1\linewidth]{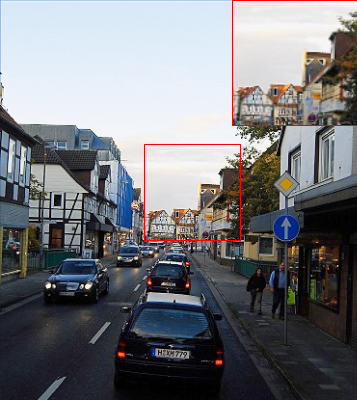}
        \caption*{}
    \end{subfigure} 
    \begin{subfigure}{.14\textwidth}
      \centering
        \includegraphics[width=1\linewidth]{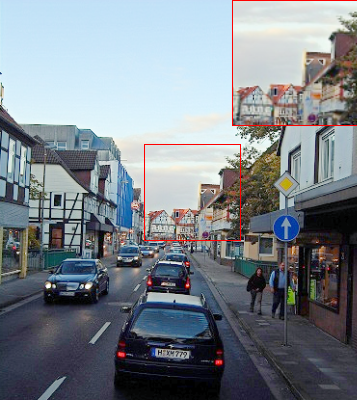}
        \caption*{}
    \end{subfigure} 

\vspace{-1.2em}

    \begin{subfigure}{.14\textwidth}
      \centering
        \includegraphics[width=1\linewidth]{Images/comparison-figures/CROP_000000_ORIG_LIME_5.png}
        \caption*{Input } 
    \end{subfigure} %
    \begin{subfigure}{.14\textwidth}
      \centering
        \includegraphics[width=1\linewidth]{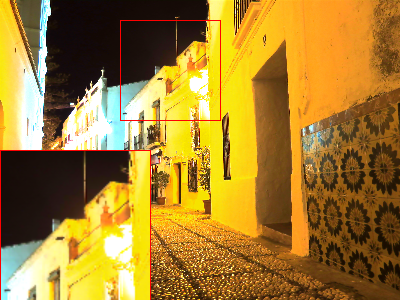}
        \caption*{RUAS \cite{liu2021retinex}}
    \end{subfigure} %
    \begin{subfigure}{.14\textwidth}
      \centering
        \includegraphics[width=1\linewidth]{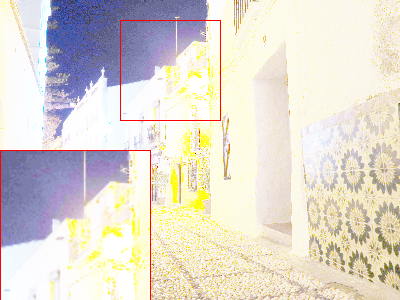}
        \caption*{SGZ \cite{zheng2022semantic}}
    \end{subfigure} %
    \begin{subfigure}{.14\textwidth}
      \centering
        \includegraphics[width=1\linewidth]{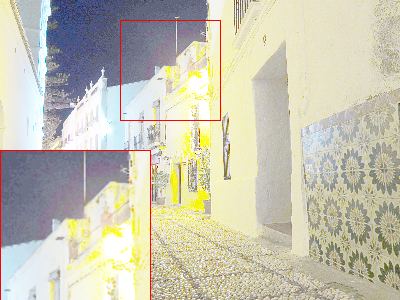}
        \caption*{Zero-DCE \cite{guo2020zero}}
    \end{subfigure} 
    \begin{subfigure}{.14\textwidth}
      \centering
        \includegraphics[width=1\linewidth]{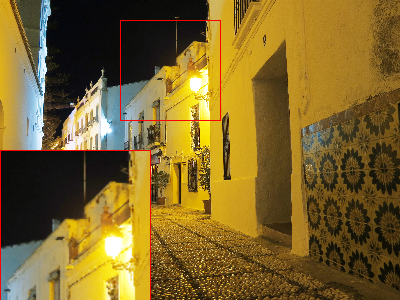}
        \caption*{SCI \cite{ma2022toward}}
    \end{subfigure} 
    \begin{subfigure}{.14\textwidth}
      \centering
        \includegraphics[width=1\linewidth]{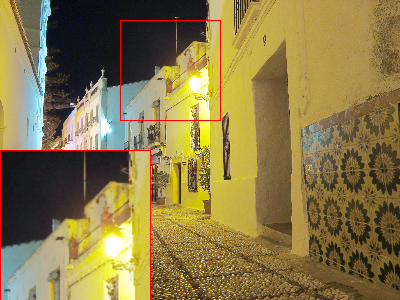}
        \caption*{Ours}
    \end{subfigure} 

    \caption{Qualitative comparison with related methods.} 
    \label{fig:comparison-qualitative}
\end{figure*}

%% file: Tables/Tab_Comparison.tex
\begin{table*}[]
\caption{Quantitative comparisons of the proposed and related works in terms of task-based performance. The computational complexity is calculated for processing a $1024 \times 1024$ px image. For comparison, the employed Yolo-World model \cite{cheng2024yolo} has a computational complexity of 225.53GMacs for this input size. 
\rebuttal{Each of the methods is trained using our training dataset introduced in Subsec. \ref{ssec:experimental_results_implementation_details}, Fig. \ref{fig:sources-samples} and Fig. \ref{fig:dataset-statistics}. }}
\label{tab:comparison}

\centering
\scriptsize

\begin{tabular}{c|cccccc|cc}
\hline

Method & NOD \cite{morawski2021nod} & NOD SE \cite{morawski2021nod} & LOD \cite{hong2021crafting} & ExDark \cite{loh2019getting} & ExDark \cite{loh2019getting} & DarkFace \cite{poor_visibility_benchmark} & \tiny{Complexity} &  \\
 & mAP $\uparrow$ & mAP $\uparrow$ & mAP $\uparrow$ & mAP $\uparrow$ & class. acc. $\uparrow$ & mAP@.5 $\uparrow$ & MACs $\downarrow$ & Params. $\downarrow$ \\ \hline

\rule{0pt}{2ex} (no enhancement) & 42.1\% & 22.8\% & 37.8\% & \firstbest{40.3\%} & 22.8\% & 33.3\% & - & - \\

Histogram Equalization & 39.8\% & 18.4\% & 38.9\% & 33.3\% & 39.9\% & 32.9\% & - & - \\

LIME \cite{guo2016lime} & 42.0\% & 23.7\% & 40.5\% & 38.3\% & 38.9\% & 38.3\% & - & -  \\
\hline

\rule{0pt}{3ex} RUAS \cite{liu2021retinex} \textsuperscript{(CVPR21)} & 38.0\% & 17.4\% & 36.3\% & 38.1\% & 34.7\% & 32.7\% & 4G & 3.4K\\

SGZ \cite{zheng2022semantic} \textsuperscript{(WACV22)} & 39.1\% & 21.9\% & 41.3\% & 31.7\% & 34.5\% & 32.4\% & 11G & 11K \\ 

Zero-DCE \cite{guo2020zero} \textsuperscript{(CVPR20)}  & 41.2\%  & 22.8\%  & 41.1\%  & 34.0\%  & \firstbest{44.1\%}  & 35.9\% & 83G & 79K \\ 

SCI \cite{ma2022toward} \textsuperscript{(CVPR22)}  & \firstbest{43.9\%}& \firstbest{27.0\%} & 45.6\% & \secondbest{40.2\%} & 35.3\% & 39.4\% & 0.37G & 0.36K \\ 

Ours (inference-time $32 \times $ $\downarrow$ \& $\uparrow$-scaling) *  & \secondbest{43.8\%}  & \secondbest{26.7\%}  & \secondbest{47.0\%}  & 39.9\%  & 40.1\%  & \secondbest{42.0\%} & 0.08G & 79K \\

Ours  & \firstbest{43.9\%} & \firstbest{27.0\%}  & \firstbest{47.1\%}  & 39.9\%  & \secondbest{42.5\%} & \firstbest{42.1\%} & 83G & 79K \\

\hline
\end{tabular}
\\ \colorlegend
\\ * the input to the network is down-scaled $32 \times $, the output parameter map is bilinearly up-scaled back and applied to the original image \cite{li2021learning} 

\end{table*}

%% file: Tables/Tab_Comparison_PSNR.tex
\begin{table}[]
\caption{Quantitative comparisons of the proposed and related works in terms of the PSNR and SSIM metrics on the VE-LOL \cite{liu2021benchmarking} dataset. The computational complexity is calculated for processing a $1024 \times 1024$ px image. \rebuttal{Each of the methods is trained using our training dataset introduced in Subsec. \ref{ssec:experimental_results_implementation_details}, Fig. \ref{fig:sources-samples} and Fig. \ref{fig:dataset-statistics}. }}
\label{tab:comparison-psnr}

\centering
\scriptsize

\begin{tabular}{c|ccc}
\hline

Method & PSNR $\uparrow$ & SSIM $\uparrow$ 
& \tiny{Complexity} \\ 
 &  & & MACs $\downarrow$ \\ 
\hline

\rule{0pt}{2ex} (no enhancement) & 10.23 & 0.29 & - \\
\hline

\rule{0pt}{1ex} RUAS \cite{liu2021retinex} & \secondbest{13.74} & \secondbest{0.53} & 4G \\

SGZ \cite{zheng2022semantic} & 8.81 & 0.40 & 11G \\

Zero-DCE \cite{guo2020zero}& 11.18 & 0.45 & 83G \\

SCI \cite{ma2022toward} & 13.25 & 0.48 & 0.37G \\

Ours & \firstbest{15.62} & \firstbest{0.57} & 83G \\

\hline
\end{tabular}
\\ \colorlegend
\end{table}

%% file: Tables/Tab_Comparison_LOD.tex
\begin{table}[]
\caption{Quantitative comparison of the related methods on the LOD \cite{hong2021crafting} dataset, including lower- and upper-bound performance, using corresponding low- and normal-light data, respectively} 
\label{tab:comparison-LOD}
\centering
\scriptsize

\begin{tabular}{c|ccc}
\hline

Method & mAP $\uparrow$ & mAP@.5 $\uparrow$ & mAP@.75 $\uparrow$ \\ \hline

Low-Light (Lower Bound) & 37.8\% & 56.1\% & 41.5\% \\

Normal-Light (Upper Bound) & 59.4\% & 81.4\% & 66.1\% \\
\hline

\rule{0pt}{3ex} 
Histogram Equalization & 38.9\% & 57.4\% & 42.0\% \\

\hline
 RUAS \cite{liu2021retinex} & 36.3\% & 54.0\% & 39.3\% \\

SGZ \cite{zheng2022semantic} &  41.3\% & 61.1\% & 44.7\% \\

Zero-DCE \cite{guo2020zero}& 41.1\% & 60.6\% & 44.6\% \\

SCI \cite{ma2022toward} & \secondbest{45.6\%} & \secondbest{66.7\%} & \secondbest{50.2\%} \\

Ours &  \firstbest{47.1\%} & \firstbest{68.6\%} & \firstbest{51.8\%} \\

\hline

\end{tabular}
\\ \colorlegend
\end{table}

%% file: Figures/Figures_Low_Light_Impact.tex
\begin{figure}[]
    \centering
    \small
    \includegraphics[trim={0cm 0cm 0cm 0cm},clip,width=1\linewidth]{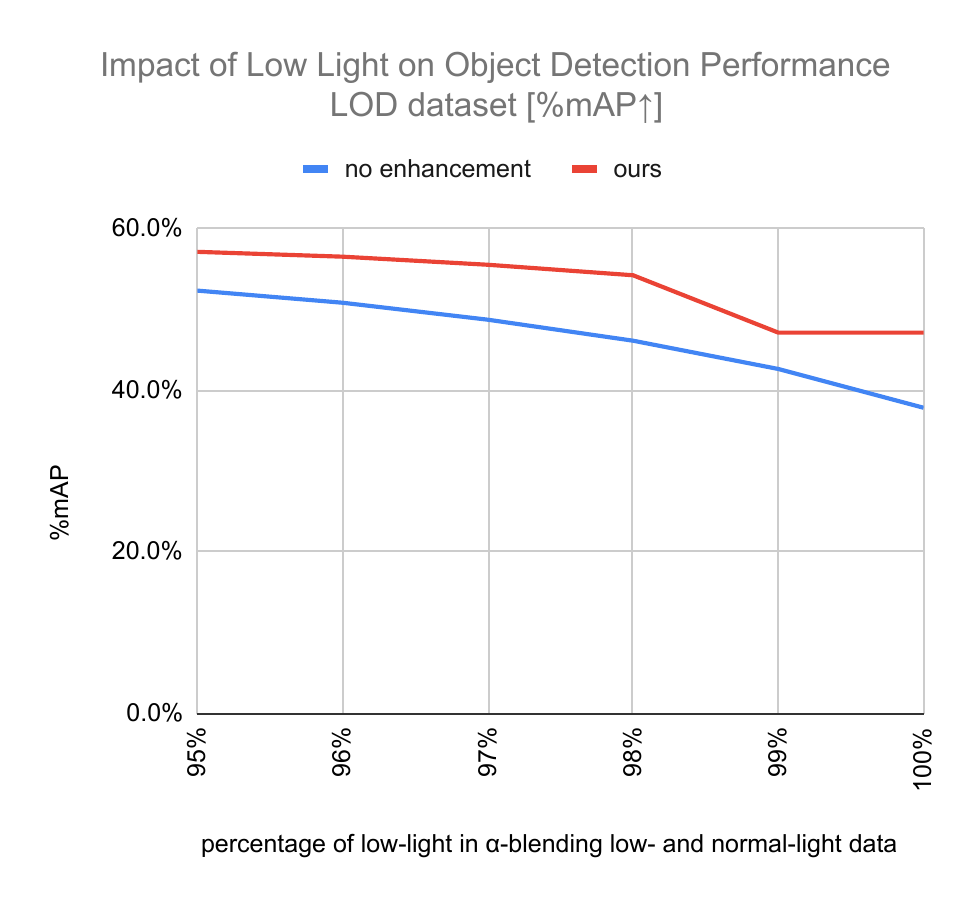}
    \caption{Impact of low light on object detection performance, measured in mean Average Precision, using the LOD \cite{hong2021crafting} dataset. Since we use JPEG data, we use $\alpha$ -blending as a proxy for simulating different levels of low light, varying proportions of corresponding low- and normal-light data.}
    \label{fig:aug-lowlight-impact}
    
\end{figure}

%% file: Tables/Tab_crossdataset_comparison.tex
\begin{table*}[]
\rebuttalcontainer
\caption{\rebuttalcontainer
 Quantitative comparisons of the proposed and related paired and unpaired data methods in terms of task-based performance. The computational complexity is calculated for processing a $1024 \times 1024$ px image. For comparison, the employed Yolo-World model \cite{cheng2024yolo} has a computational complexity of 225.53GMacs for this input size. All methods in this table are compared using original publicly available checkpoints, i.e. trained using original datasets.}
\label{tab:crossdataset_comparison} 

\centering
\scriptsize

\begin{tabular}{c|cccccc|cc}
\hline

Method & NOD \cite{morawski2021nod} & NOD SE \cite{morawski2021nod} & LOD \cite{hong2021crafting} & ExDark \cite{loh2019getting} & ExDark \cite{loh2019getting} & DarkFace \cite{poor_visibility_benchmark} & \tiny{Complexity} &  \\
 & mAP $\uparrow$ & mAP $\uparrow$ & mAP $\uparrow$ & mAP $\uparrow$ & class. acc. $\uparrow$ & mAP@.5 $\uparrow$ & MACs $\downarrow$ & Params. $\downarrow$ \\ \hline

\rule{0pt}{2ex} (no enhancement) & 42.1\% & 22.8\% & 37.8\% & \firstbest{40.3\%} & 22.8\% & 33.3\% & - & - \\

\hline 

\multicolumn{1}{l|}{PAIRED DATA METHODS:} & & & & & & & & \\

SNR+SKF \cite{Wu_2023_CVPR} \textsuperscript{(CVPR2023)} & 41.6\% & 20.1\% & 41.8\% & 39.0\% & 40.8\% & 38.6\% & 886G & 105M\\

DRBN+SKF \cite{Wu_2023_CVPR} \textsuperscript{(CVPR2023)} & 43.1\% & 25.1\% & 46.5\% & 39.5\% & 40.2\% & \firstbest{42.7\%} & 560G & 66M \\

PairLIE \cite{Fu_2023_CVPR_Learning} \textsuperscript{(CVPR2023)} & 41.4\% & 22.1\% & 45.6\% & 38.0\% & \firstbest{47.7\%} & 37.8\% & 359G & 341K \\

DiffLL \cite{jiang2023low} \textsuperscript{(SIGGRAPH ASIA 2023)} & 40.6\% & 22.6\% & 42.8\% & 35.6\% & 41.2\% & 39.4\% & 1410G & 22M \\

\hline 

\multicolumn{1}{l|}{UNPAIRED DATA/UNSUPERVISED METHODS:}  & & & & & & & & \\

CLIP-LIT \cite{liang2023iterative} \textsuperscript{(ICCV2023)} & 42.9\% & 23.0\% & 43.6\% & 38.0\% & 38.5\% & 41.1\% & 292G & 279K\\

PIE {} {} \cite{liang2024pie} \textsuperscript{(IJCV2024)} & 43.7\% & \firstbest{27.3\%}& \secondbest{47.0\%} & 39.6\% & 37.7\% & 40.3\% & 83G & 79K \\

PIE* \cite{liang2024pie} \textsuperscript{(IJCV2024)} & 43.5\% & 26.8\% & 46.5\% & 39.1\% & 33.4\% & 40.6\% & 0.08G & 79K \\

QuadPrior (FP16) \cite{wang2024zero} \textsuperscript{(CVPR24)} & 41.6\% & 22.3\% & 45.1\% & 39.4\% & 33.7\% & 36.5\% & 13740G & 1314M \\

\hline \rule{0pt}{2ex}

Ours*  & \secondbest{43.8\%}  & {26.7\%}  & \secondbest{47.0\%}  & \secondbest{39.9\%}  & 40.1\%  & {42.0\%} & 0.08G & 79K \\

Ours {} {} & \firstbest{43.9\%} & \secondbest{27.0\%}  & \firstbest{47.1\%}  & \secondbest{39.9\%} & \secondbest{42.5\%} & \secondbest{42.1\%} & 83G & 79K \\

\hline
\end{tabular}
\\ \colorlegend
\\ * the input to the network is down-scaled $32 \times $, the output parameter map is bilinearly up-scaled back and applied to the original image \cite{li2021learning} 

\end{table*}

%% file: Tables/Tab_cross-comparison-psnr.tex
\begin{table}[]
\rebuttalcontainer
\caption{\rebuttalcontainer Quantitative comparisons of the proposed and related paired and unpaired data methods in terms of the PSNR and SSIM metrics on the VE-LOL \cite{liu2021benchmarking} dataset. The computational complexity is calculated for processing a $1024 \times 1024$ px image. All methods in this table are compared using original publicly available checkpoints, i.e. trained using original datasets.}
\label{tab:cross-comparison-psnr}

\centering
\scriptsize

\begin{tabular}{c|ccc}
\hline

Method & PSNR $\uparrow$ & SSIM $\uparrow$ 
& \tiny{Complexity} \\ 
 &  & & MACs $\downarrow$ \\ 
\hline

\rule{0pt}{2ex} (no enhancement) & 10.23 & 0.29 \\
\hline

\multicolumn{1}{l|}{PAIRED DATA METHODS:} & \\

SNR+SKF \cite{Wu_2023_CVPR} \textsuperscript{(CVPR2023)} & \secondbest{23.40} & \secondbest{0.76} & 886G \\

DRBN+SKF \cite{Wu_2023_CVPR} \textsuperscript{(CVPR2023)} & 21.02 & \secondbest{0.76} & 560G \\

PairLIE \cite{Fu_2023_CVPR_Learning} \textsuperscript{(CVPR2023)} & 18.38 & 0.69  & 359G \\

DiffLL \cite{jiang2023low} \textsuperscript{(SIGGRAPH ASIA 2023)} & \firstbest{25.07} & \firstbest{0.79} & 1410G \\

\hline 

\multicolumn{1}{l|}{UNPAIRED DATA/UNSUPERVISED:} & \\

CLIP-LIT \cite{liang2023iterative} \textsuperscript{(ICCV2023)} & 14.17 & 0.55 & 292G \\

PIE {} {} \cite{liang2024pie} \textsuperscript{(IJCV2024)} & 13.46 & 0.52 & 83G \\

QuadPrior (FP16) \cite{wang2024zero} \textsuperscript{(CVPR24)} & 17.24 & 0.72 & 13740G \\

\hline 

Ours & 15.62 & 0.57 & 83G \\

\hline
\end{tabular}
\\ \colorlegend
\end{table}

%% file: Figures/Figure_comparisonbbox.tex
\begin{figure*}[t]

    \captionsetup[subfigure]{justification=centering}
    \rebuttalcontainer 
    \centering
    \small

    \begin{subfigure}{.14\textwidth}
      \centering
        \includegraphics[width=1\linewidth]{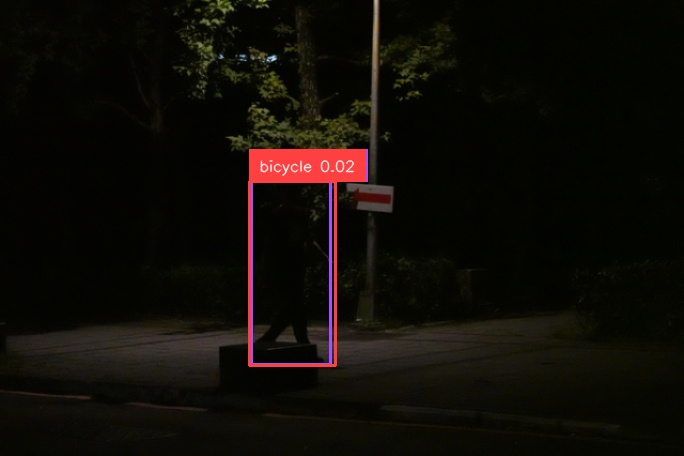}
        \caption*{}
    \end{subfigure} %
    \begin{subfigure}{.14\textwidth}
      \centering
        \includegraphics[width=1\linewidth]{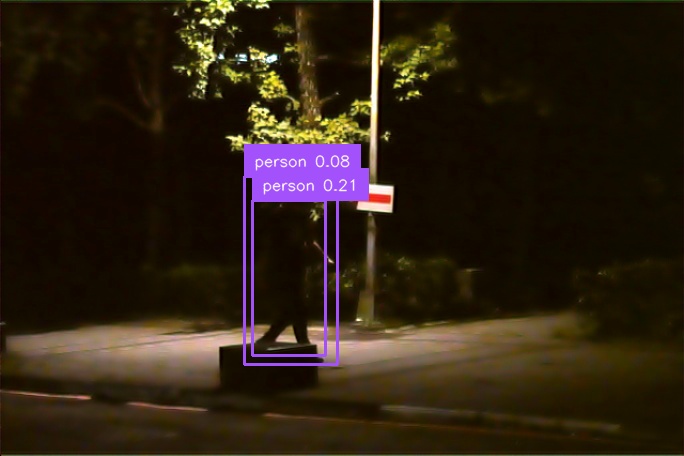}
        \caption*{}
    \end{subfigure} %
    \begin{subfigure}{.14\textwidth}
      \centering
        \includegraphics[width=1\linewidth]{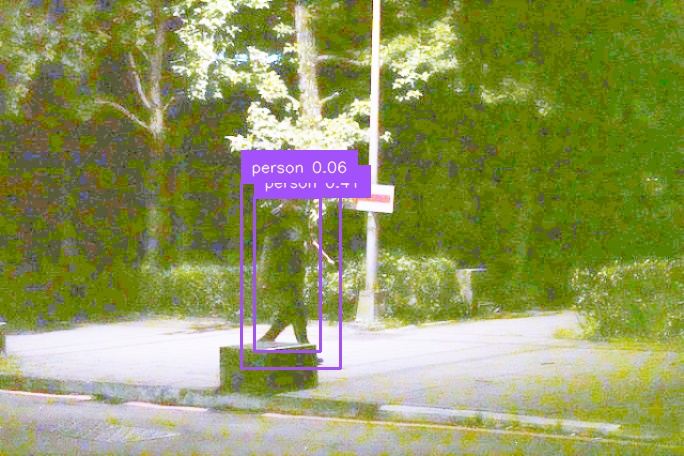}
        \caption*{}
    \end{subfigure} %
    \begin{subfigure}{.14\textwidth}
      \centering
        \includegraphics[width=1\linewidth]{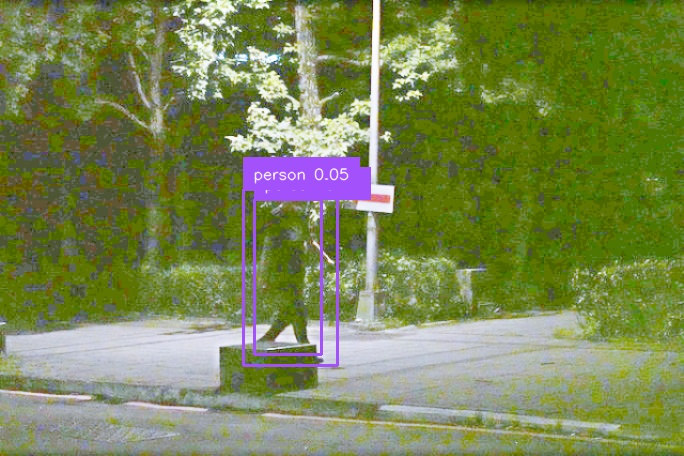}
        \caption*{}
    \end{subfigure} 
    \begin{subfigure}{.14\textwidth}
      \centering
        \includegraphics[width=1\linewidth]{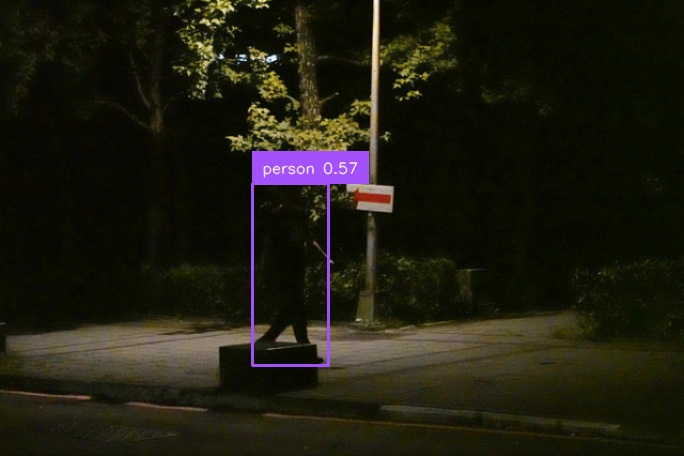}
        \caption*{}
    \end{subfigure} 
    \begin{subfigure}{.14\textwidth}
      \centering
        \includegraphics[width=1\linewidth]{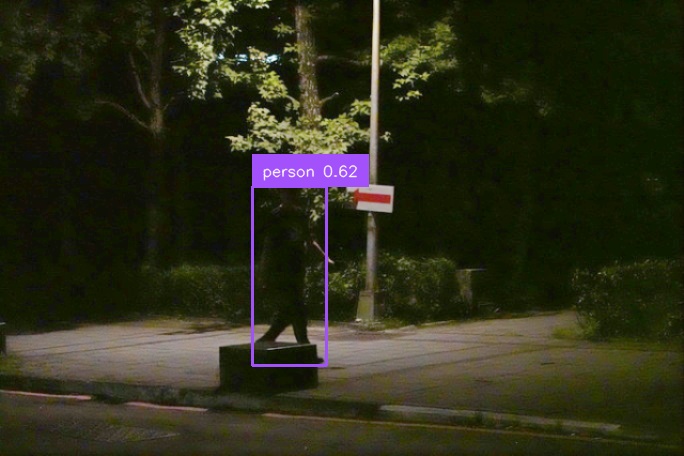}
        \caption*{}
    \end{subfigure} 

\vspace{-1.2em}



    \begin{subfigure}{.14\textwidth}
      \centering
        \includegraphics[width=1\linewidth]{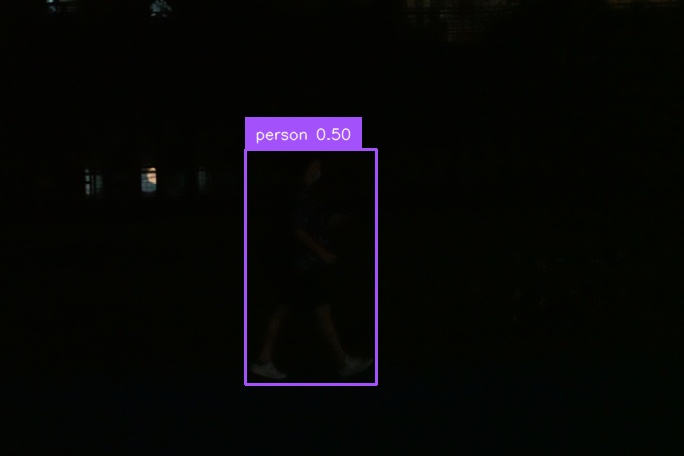}
        \caption*{}
    \end{subfigure} %
    \begin{subfigure}{.14\textwidth}
      \centering
        \includegraphics[width=1\linewidth]{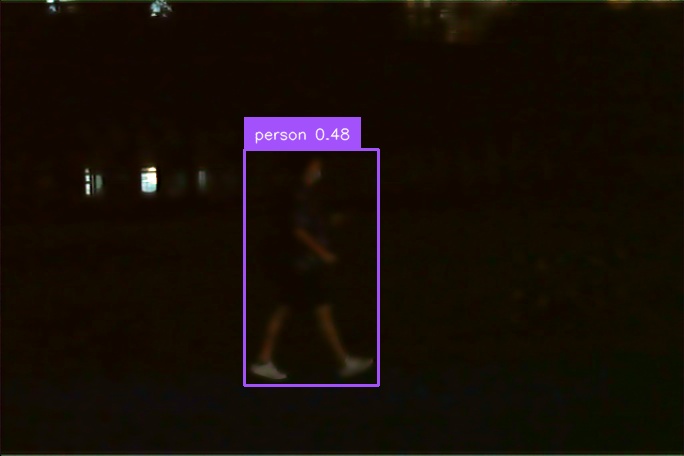}
        \caption*{}
    \end{subfigure} %
    \begin{subfigure}{.14\textwidth}
      \centering
        \includegraphics[width=1\linewidth]{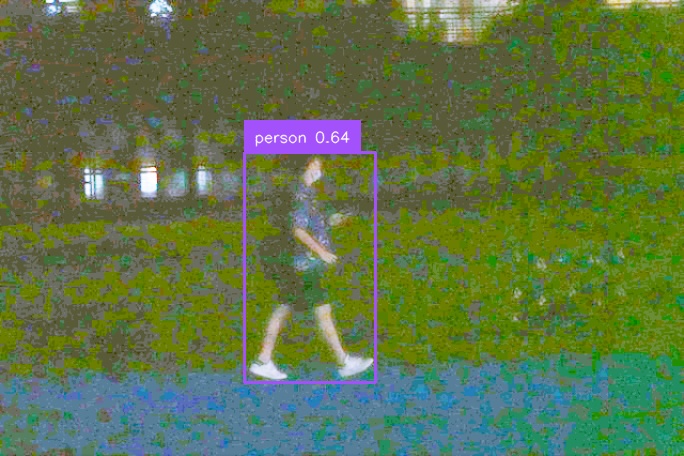}
        \caption*{}
    \end{subfigure} %
    \begin{subfigure}{.14\textwidth}
      \centering
        \includegraphics[width=1\linewidth]{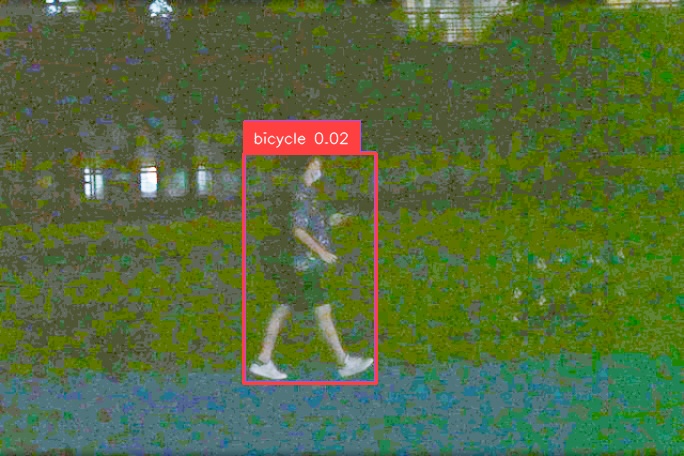}
        \caption*{}
    \end{subfigure} 
    \begin{subfigure}{.14\textwidth}
      \centering
        \includegraphics[width=1\linewidth]{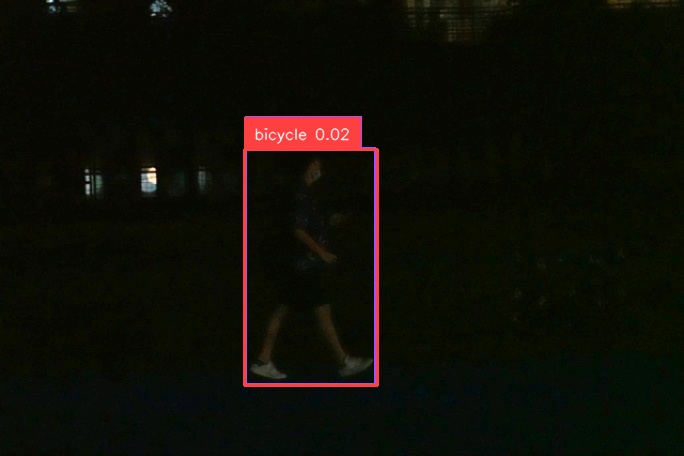}
        \caption*{}
    \end{subfigure} 
    \begin{subfigure}{.14\textwidth}
      \centering
        \includegraphics[width=1\linewidth]{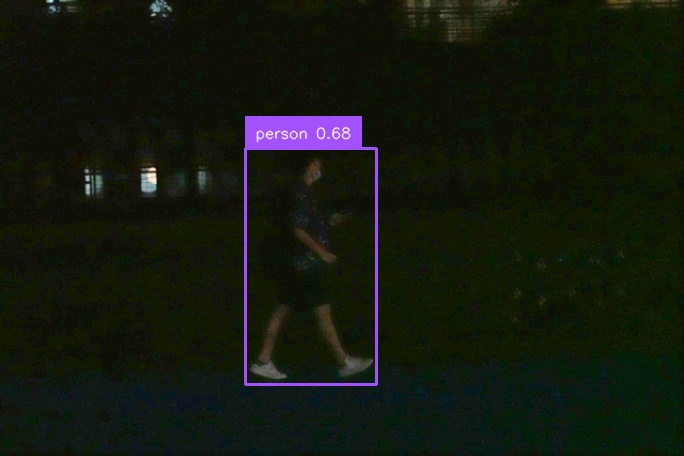}
        \caption*{}
    \end{subfigure} 

\vspace{-1.2em}

    \begin{subfigure}{.14\textwidth}
      \centering
        \includegraphics[width=1\linewidth]{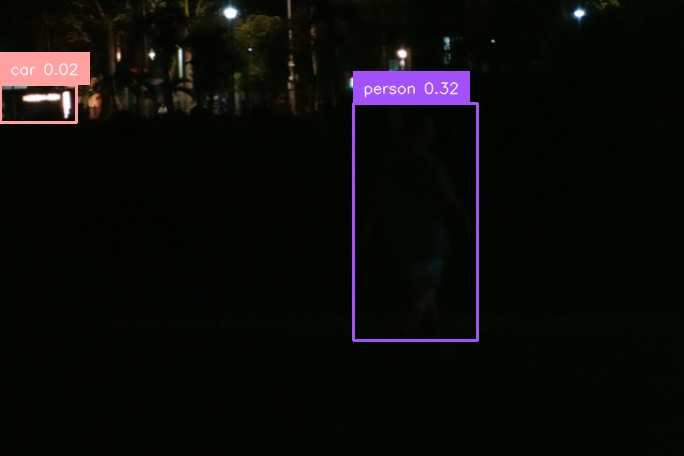}
        \caption*{}
    \end{subfigure} %
    \begin{subfigure}{.14\textwidth}
      \centering
        \includegraphics[width=1\linewidth]{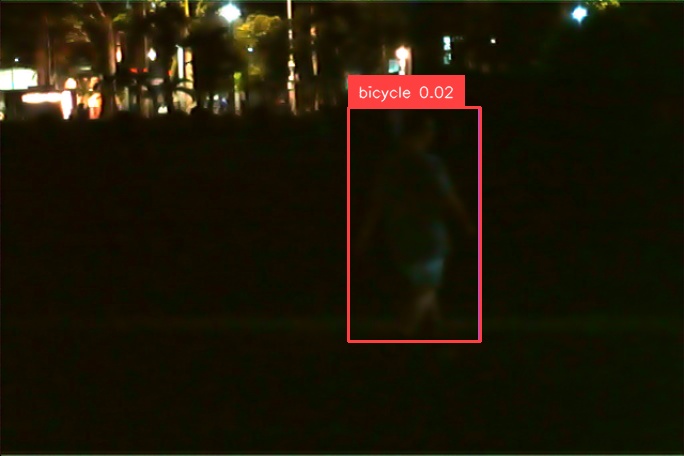}
        \caption*{}
    \end{subfigure} %
    \begin{subfigure}{.14\textwidth}
      \centering
        \includegraphics[width=1\linewidth]{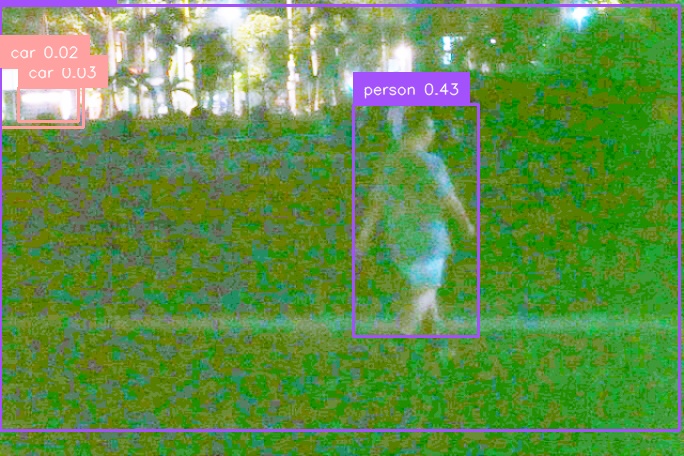}
        \caption*{}
    \end{subfigure} %
    \begin{subfigure}{.14\textwidth}
      \centering
        \includegraphics[width=1\linewidth]{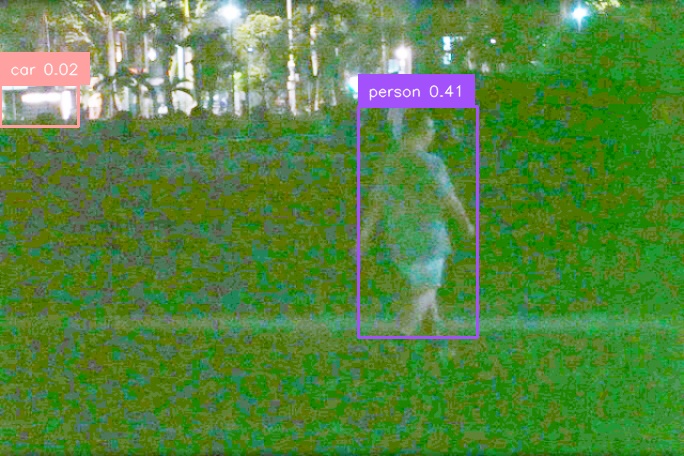}
        \caption*{}
    \end{subfigure} 
    \begin{subfigure}{.14\textwidth}
      \centering
        \includegraphics[width=1\linewidth]{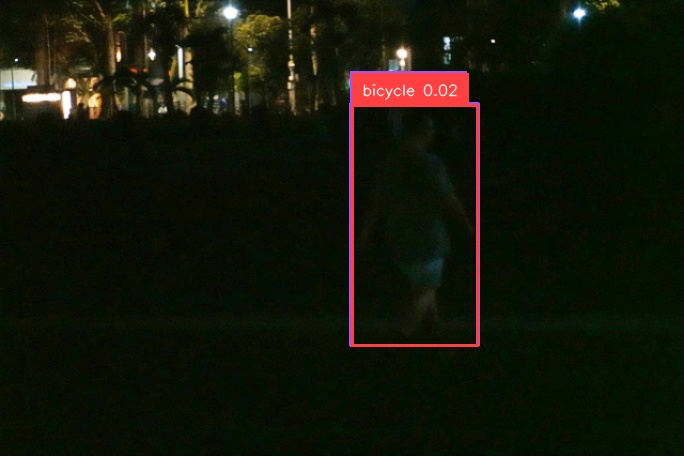}
        \caption*{}
    \end{subfigure} 
    \begin{subfigure}{.14\textwidth}
      \centering
        \includegraphics[width=1\linewidth]{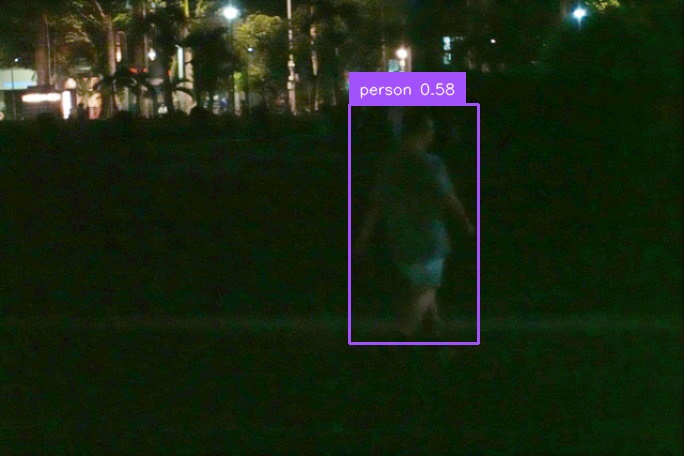}
        \caption*{}
    \end{subfigure} 

\vspace{-1.2em}

    \begin{subfigure}{.14\textwidth}
        \centering
          \includegraphics[width=1\linewidth]{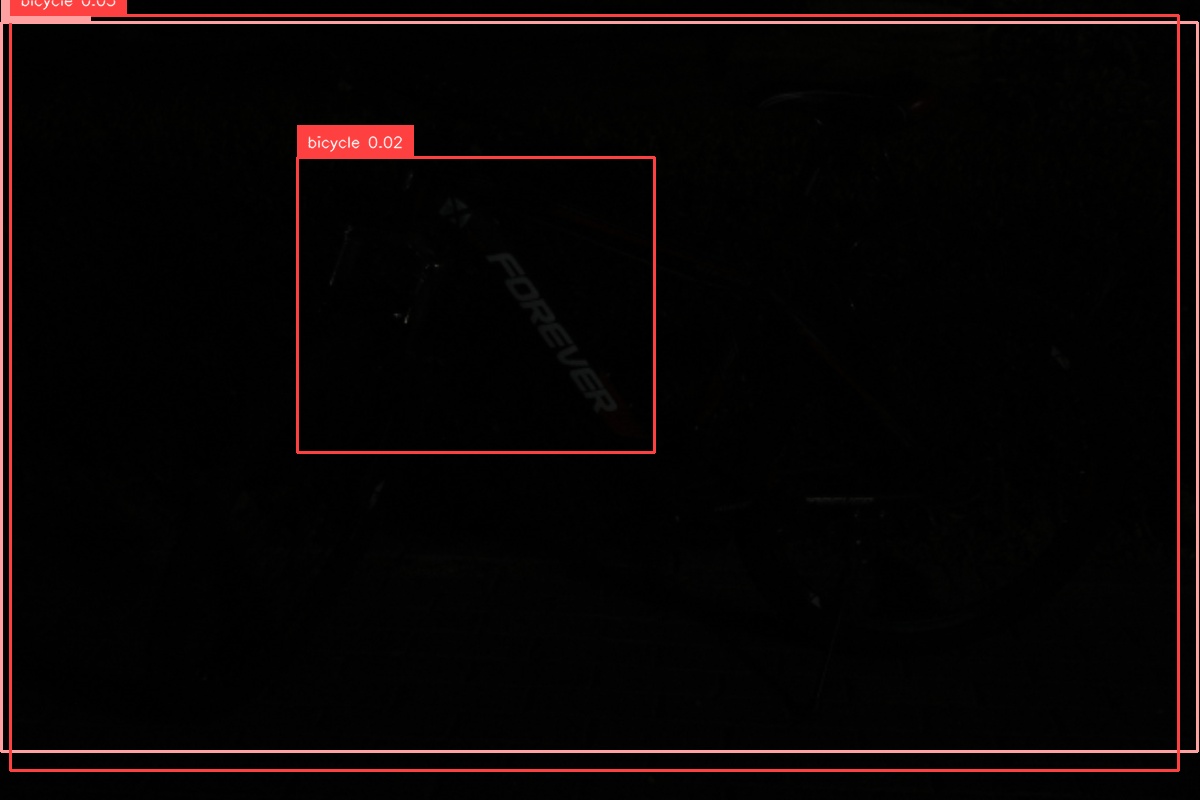}
          \caption*{}
      \end{subfigure} %
      \begin{subfigure}{.14\textwidth}
        \centering
          \includegraphics[width=1\linewidth]{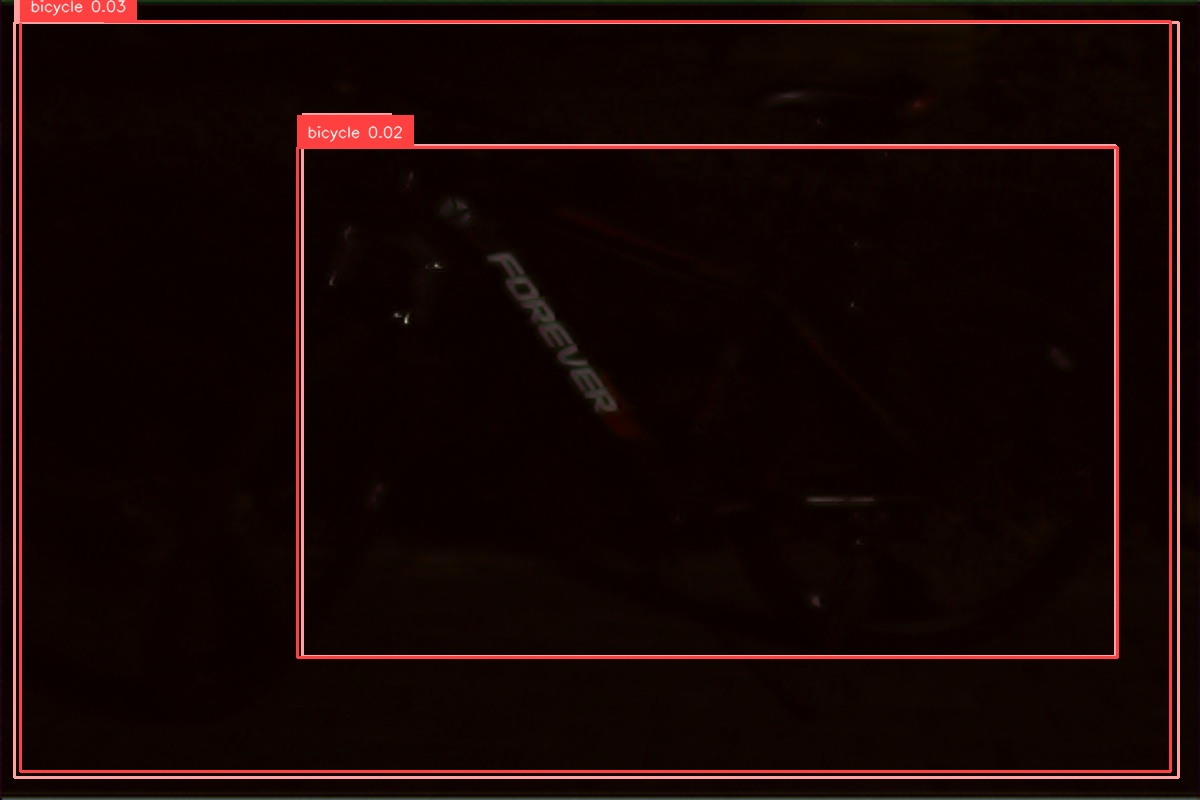}
          \caption*{}
      \end{subfigure} %
      \begin{subfigure}{.14\textwidth}
        \centering
          \includegraphics[width=1\linewidth]{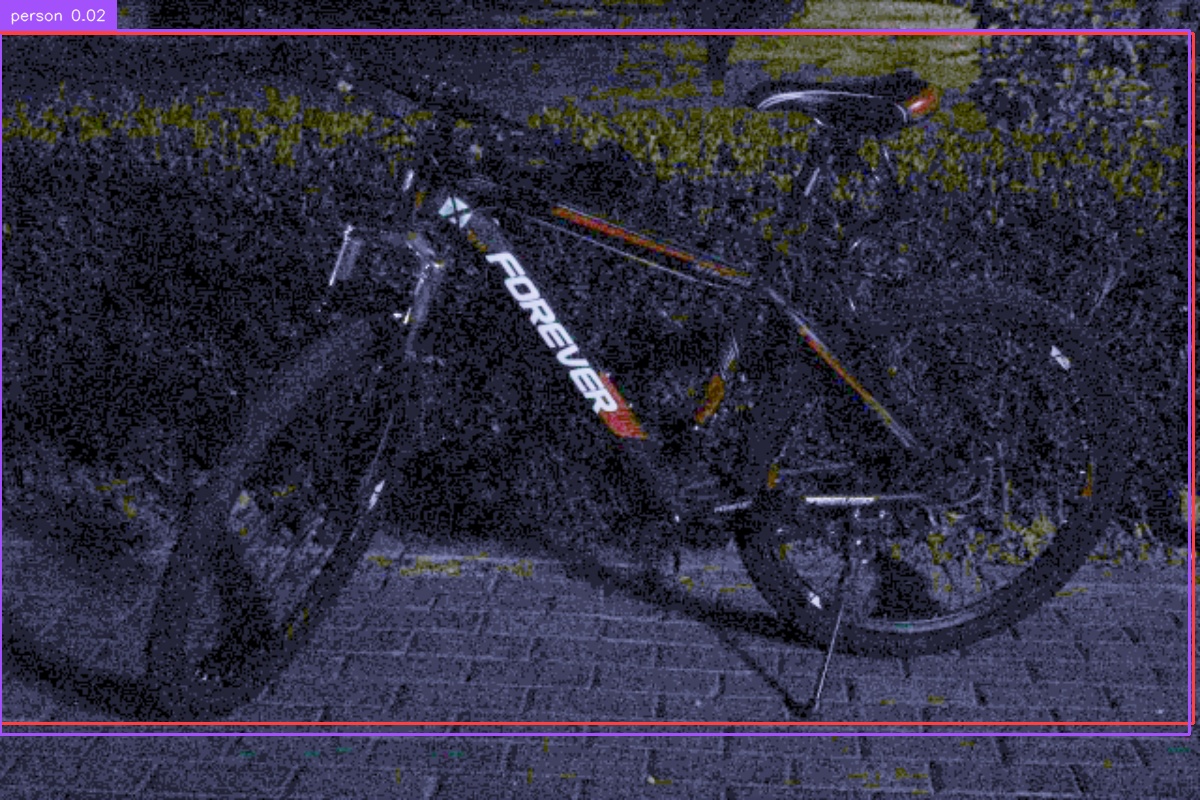}
          \caption*{}
      \end{subfigure} %
      \begin{subfigure}{.14\textwidth}
        \centering
          \includegraphics[width=1\linewidth]{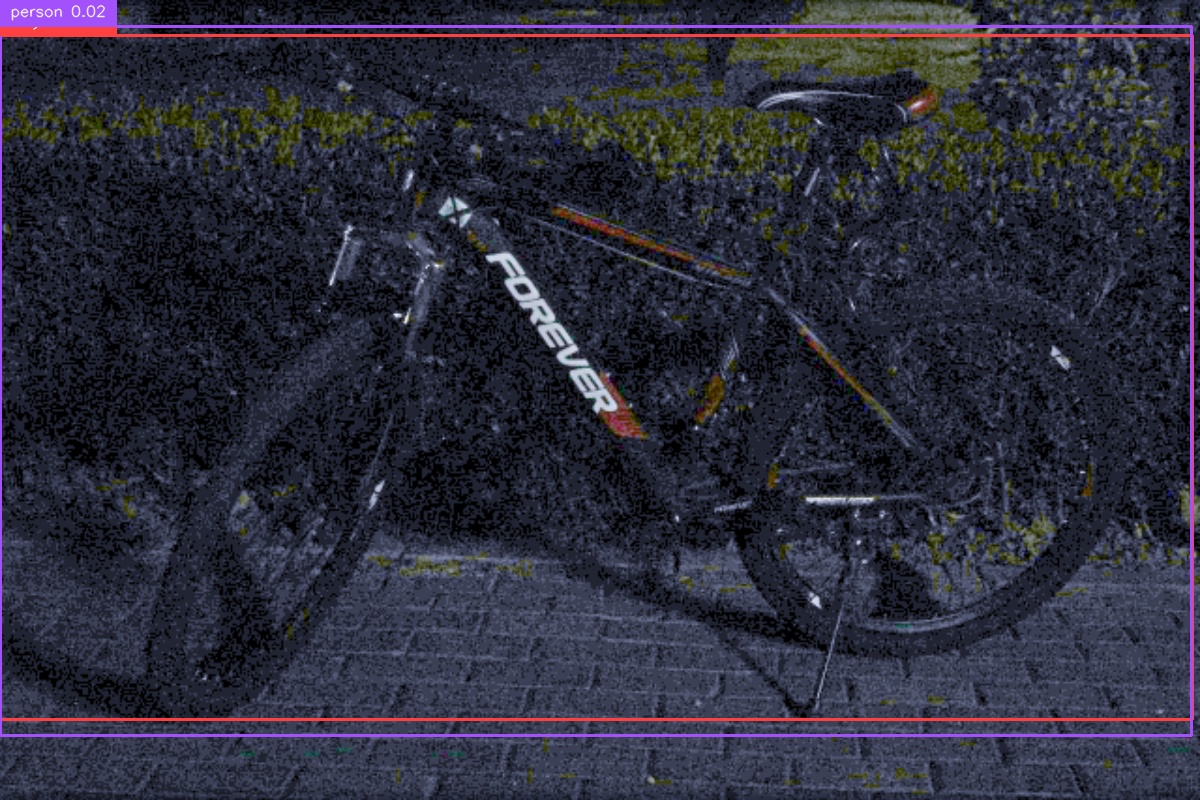}
          \caption*{}
      \end{subfigure} 
      \begin{subfigure}{.14\textwidth}
        \centering
          \includegraphics[width=1\linewidth]{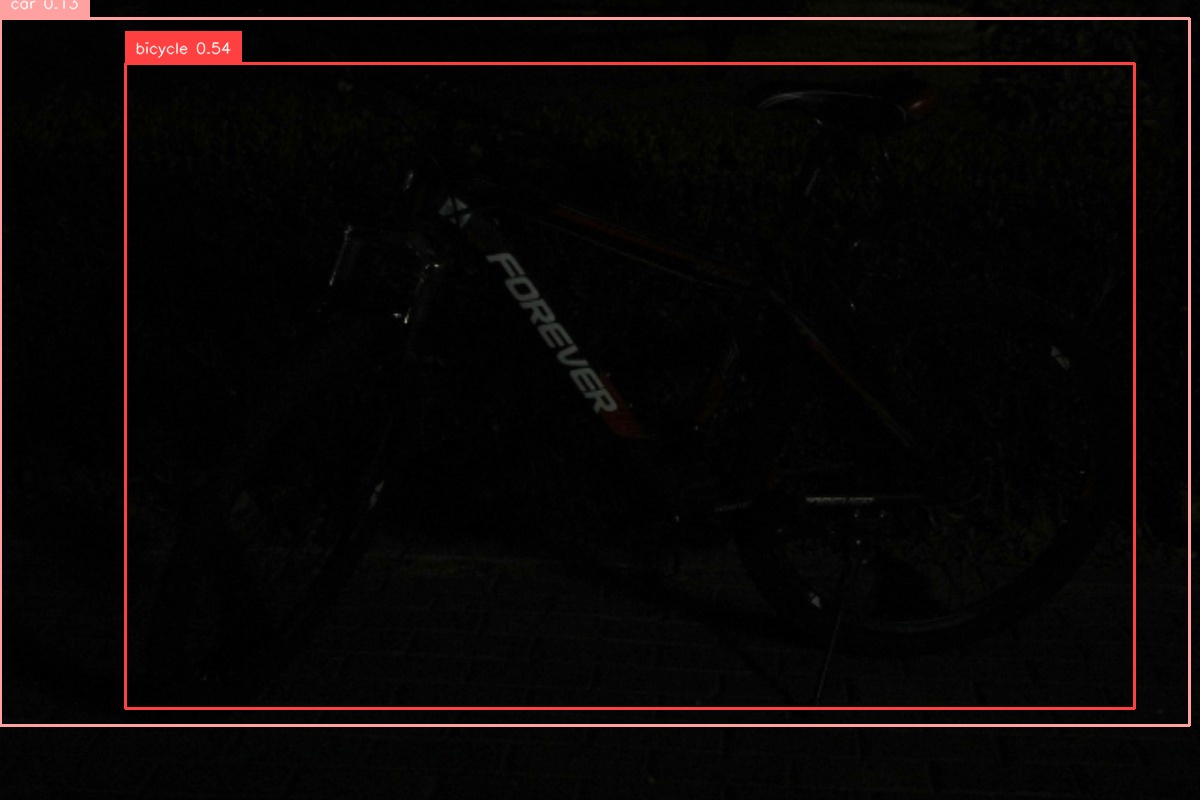}
          \caption*{}
      \end{subfigure} 
      \begin{subfigure}{.14\textwidth}
        \centering
          \includegraphics[width=1\linewidth]{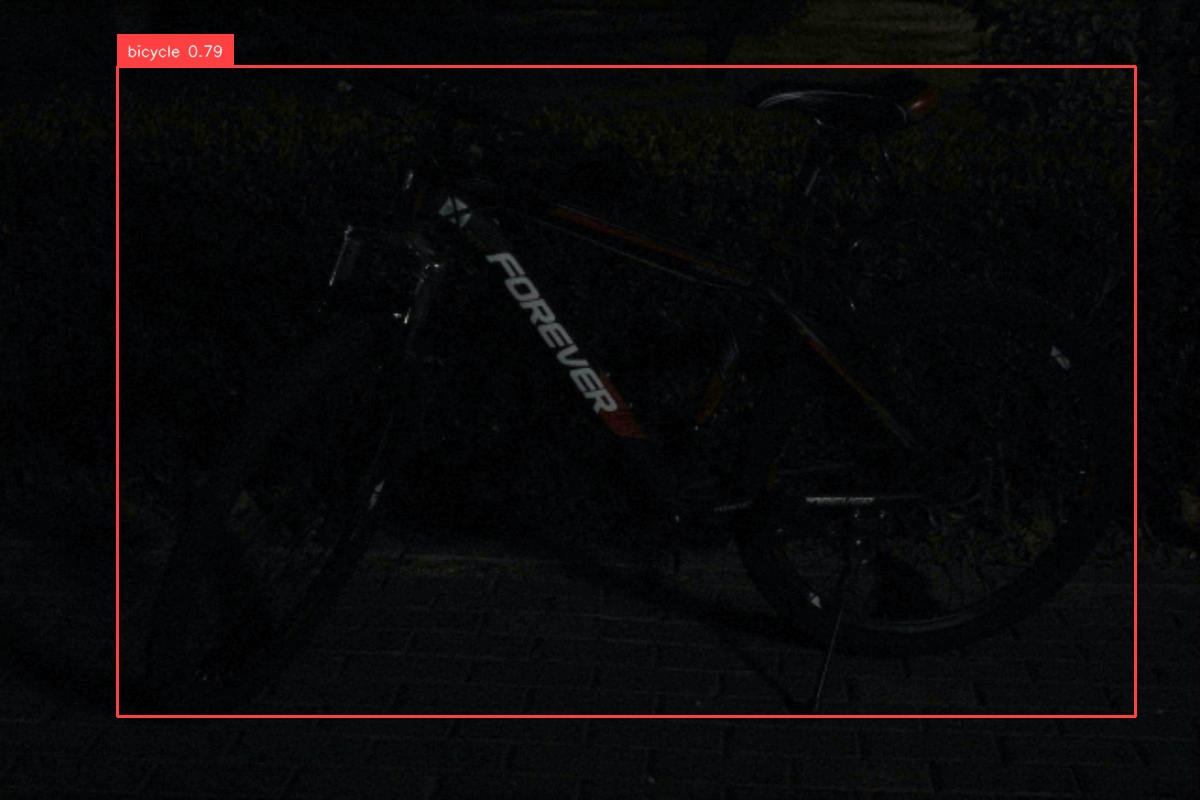}
          \caption*{}
      \end{subfigure} 

\vspace{-1.2em}

    \begin{subfigure}{.14\textwidth}
      \centering
        \includegraphics[width=1\linewidth]{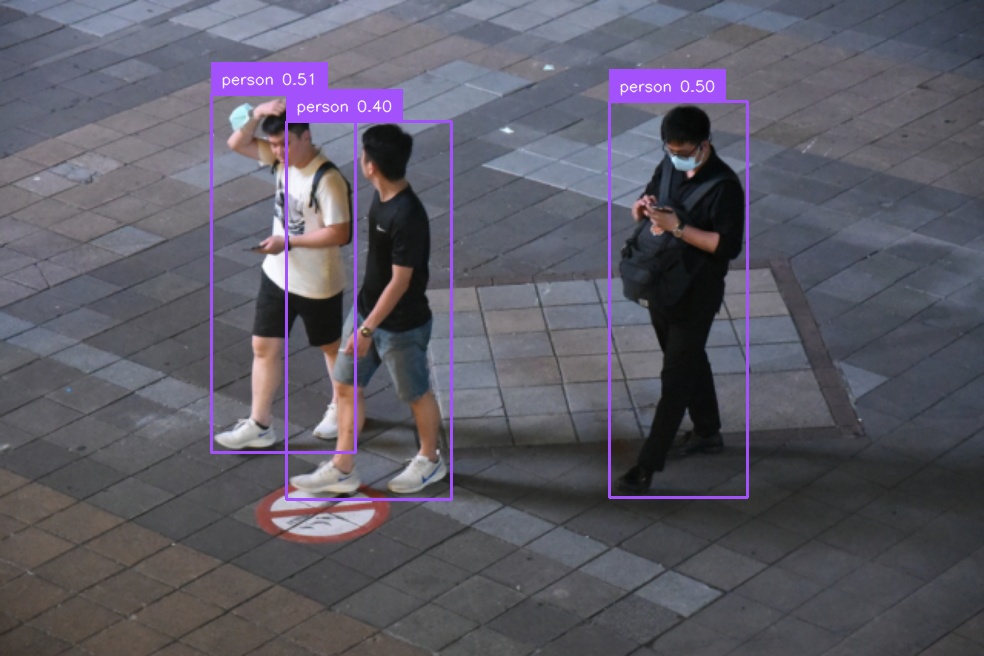}
        \caption*{Input }  
    \end{subfigure} %
    \begin{subfigure}{.14\textwidth}
      \centering
        \includegraphics[width=1\linewidth]{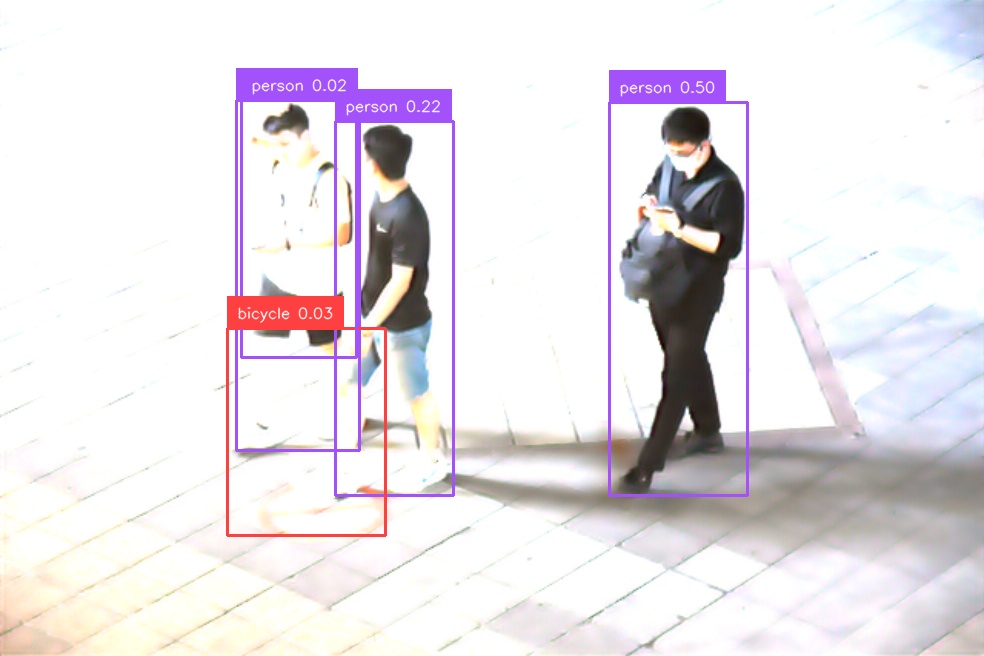}
        \caption*{RUAS \cite{liu2021retinex}}
    \end{subfigure} %
    \begin{subfigure}{.14\textwidth}
      \centering
        \includegraphics[width=1\linewidth]{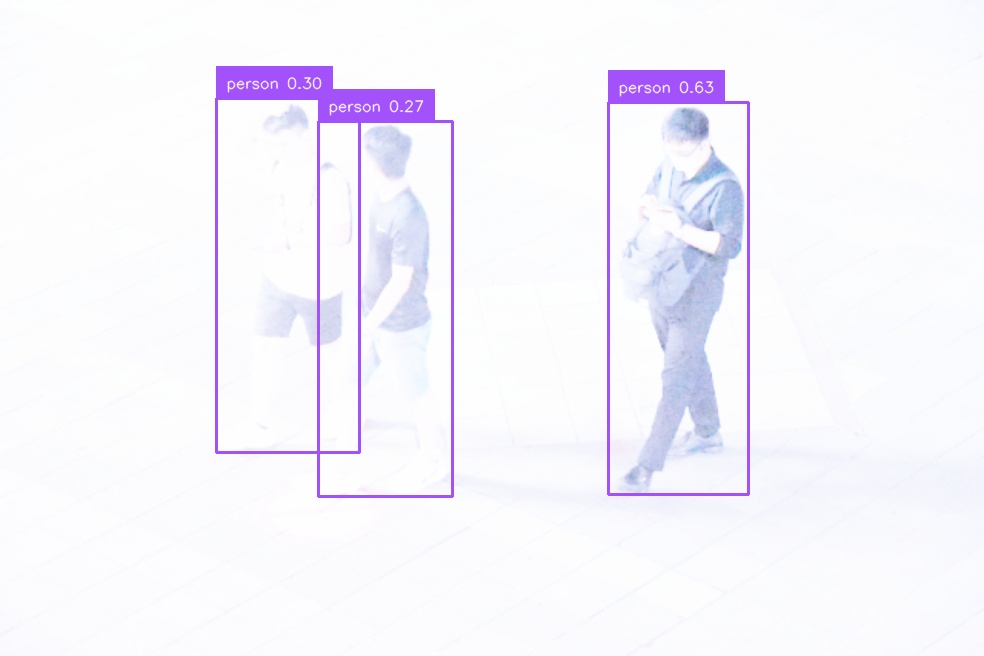}
        \caption*{SGZ \cite{zheng2022semantic}}
    \end{subfigure} %
    \begin{subfigure}{.14\textwidth}
      \centering
        \includegraphics[width=1\linewidth]{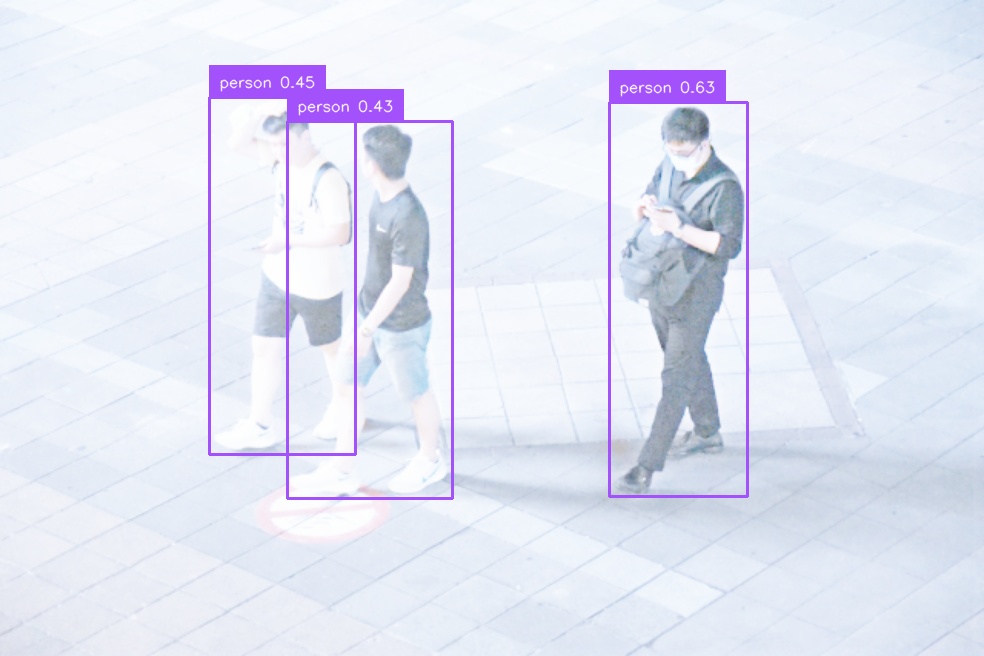}
        \caption*{Zero-DCE \cite{guo2020zero}}
    \end{subfigure} 
    \begin{subfigure}{.14\textwidth}
      \centering
        \includegraphics[width=1\linewidth]{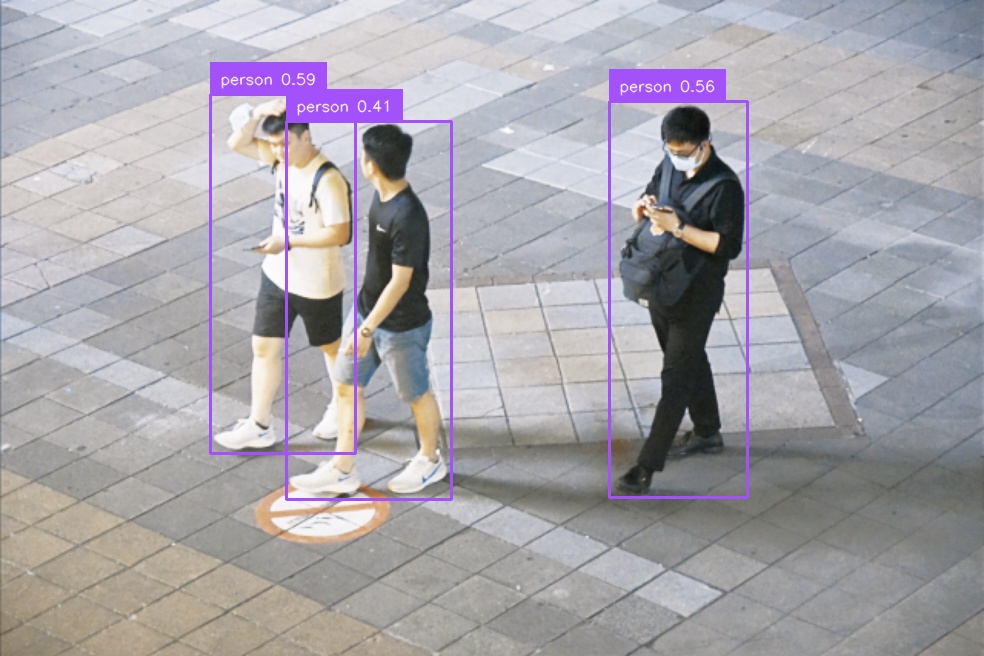}
        \caption*{SCI \cite{ma2022toward}}
    \end{subfigure} 
    \begin{subfigure}{.14\textwidth}
      \centering
        \includegraphics[width=1\linewidth]{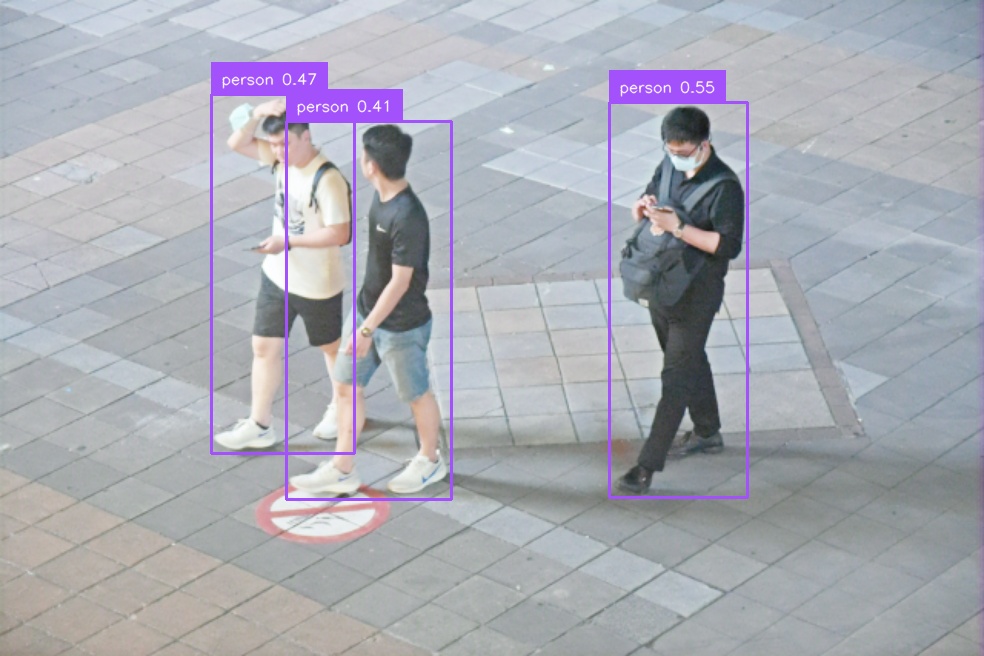}
        \caption*{Ours} 
    \end{subfigure} 

\yololegend
    \caption{\rebuttalcontainer Qualitative results of object detection using Yolo-World model \cite{cheng2024yolo} on the NOD \cite{morawski2021nod} and LOD \cite{hong2021crafting}. Instead of assuming correlation between human and machine cognition, we propose to directly optimize the enhancement for machine cognition, leading to reduction in prediction noise (first -- fourth rows). Moreover, our proposed image prior additionally helps to generalize to bright images, avoiding overexposure which may adversely impact downstream task models (last row). }


    \label{fig:comparisonbbox}
\end{figure*}

%% file: Tables/Tab_alternative_baselines.tex
\begin{table*}[]
\rebuttalcontainer
\caption{ \rebuttalcontainer
Our training strategy applied to related zero-reference baseline methods. To validate the generalizability of our proposed strategy, we add our proposed content- and context-based semantic guidance as well as our proposed learned CLIP image prior-based loss in addition to the loss function originally used by each of the methods
}
\label{tab:alternative_baselines} 

\centering
\scriptsize

\begin{tabular}{rl|cccccc}
\hline

Method & & NOD \cite{morawski2021nod} & NOD SE \cite{morawski2021nod} & LOD \cite{hong2021crafting} & ExDark \cite{loh2019getting} & ExDark \cite{loh2019getting} & DarkFace \cite{poor_visibility_benchmark} \\
  & & mAP $\uparrow$ & mAP $\uparrow$ & mAP $\uparrow$ & mAP $\uparrow$ & class. acc. $\uparrow$ & mAP@.5 $\uparrow$ \\ 
  
  \hline \rule{0pt}{3ex}
  
 RUAS \cite{liu2021retinex} \textsuperscript{(CVPR21)} & & 38.0\% & 17.4\% & 36.3\% & 38.1\% & \textbf{34.7\%} & 32.7\% \\

RUAS \cite{liu2021retinex} \textsuperscript{(CVPR21)} & + Ours & \textbf{42.0\%} & \textbf{23.5\%} &\textbf{ 42.9\%} & \textbf{40.6\%} & 30.0\% & \textbf{36.9\%} \\

\hline \rule{0pt}{3ex}

SGZ \cite{zheng2022semantic} \textsuperscript{(WACV22)} & & 39.1\% & 21.9\% & 41.3\% & 31.7\% & 34.5\% & 32.4\%  \\ 

SGZ \cite{zheng2022semantic} \textsuperscript{(WACV22)} & + Ours & \textbf{43.6\%} & \textbf{27.0\%} & \textbf{47.0\%} &\textbf{ 39.3\%} & \textbf{43.7\%} & \textbf{ 40.7\%} \\ 

\hline \rule{0pt}{3ex}
 
SCI \cite{ma2022toward} \textsuperscript{(CVPR22)} & & \textbf{43.9\%}& \textbf{27.0\%} & \textbf{45.6\%} & 40.2\% & 35.3\% & \textbf{39.4\%} \\ 

SCI \cite{ma2022toward} \textsuperscript{(CVPR22)} & + Ours & 43.7\% & 26.7\% & 45.4\% & \textbf{40.4\% }& \textbf{35.5\%} & 39.2\% \\ 

\hline \rule{0pt}{3ex}

Zero-DCE \cite{guo2020zero} \textsuperscript{(CVPR20)}  &  & 41.2\%  & 22.8\%  & 41.1\%  & 34.0\%  & \textbf{44.1\%}  & 35.9\%  \\ 

Zero-DCE \cite{guo2020zero} \textsuperscript{(CVPR20)} & + Ours  & \textbf{43.9\%} & \textbf{27.0\%}  & \textbf{47.1\%}  & \textbf{39.9\%}  & {42.5\%} & \textbf{42.1\%}  \\

\hline
\end{tabular}
\end{table*}

%% file: Sections/5_Conclusion.tex
\section{Discussion and Future Work}

\input{Figures/Figures_QP}

\rebuttal{Our proposed method has several limitations that offer insights into possible future research directions. As demonstrated, our proposed training strategy can be generalized to many zero-reference low-light enhancement methods without any need for paired or unpaired normal-light data. However, our proposed strategy still requires semantic annotation, and it would be beneficial to eliminate the need for annotations that require human input. }

\rebuttal{Moreover, we observe that the proposed strategy using Zero-DCE \cite{guo2020zero} has limited capabilities to denoise and no observed ability to inpaint. As we propose a sampling strategy to learn a CLIP \cite{radford2021learning} image prior, improvements to the sampling strategy, such as \cite{huang2021neighbor2neighbor} or novel low-light-oriented approaches and model architecture to enable more effective machine cognition-oriented denoising could be of further interest. Furthermore, leveraging pre-trained large-scale generative models could be a promising direction to enable inpainting extreme low-light instances where significant information loss is present. }

\rebuttal{While denoising is presumably a key aspect of improving task-based performance, in Tab. \ref{tab:crossdataset_comparison} and Fig. \ref{fig:quadprior-figure}, we observe that QuadPrior \cite{wang2024zero}, a related diffusion-based low-light method with strong denoising capabilities, leads to significant decrease in performance on the extremely low-light subset (SE) of the NOD \cite{morawski2021nod} dataset. In Fig. \ref{fig:quadprior-figure}, we observe that although the results of QuadPrior \cite{wang2024zero} are suitable for human perception, the over-smoothing in extreme low-light regions of the images leads to spurious responses in the downstream detector. In Sec. \ref{ssec:quant-comparison}, we also discussed that full-reference evaluation results in Tab. \ref{tab:cross-comparison-psnr} did not translate to task-based performance in Tab. \ref{tab:crossdataset_comparison}. Overall, this leads to a conclusion that in low-light enhancement, a correlation between task-based performance (machine cognition) and human perception cannot be assumed. }

\rebuttal{Moreover, maintaining low computational complexity, low inference time and enabling high-resolution inference are some of the challenges that need to be considered before increasing the capacity of the enhancement model. }

\section{Conclusions} 
In this work, motivated by CLIP's \cite{radford2021learning}  zero-shot and open-vocabulary capabilities, we proposed to use CLIP \cite{radford2021learning} to semantically guide and improve task-oriented low-light image enhancement without any need for paired or unpaired normal-light data. First, we proposed to learn an image prior via prompt learning, using a simple but effective data augmentation strategy and experimentally showed that it improved low-light image enhancement by better constraining image contrast and better discrimination of background and foreground objects, resulting in reduced under- and over-enhancement. Next, we proposed to further use the CLIP \cite{radford2021learning} model to semantically guide the enhancement process and maximally take advantage of existing low-light annotation, using image descriptions at a loss level. Specifically, we realized guidance as a classification problem using content and context cues, first matching descriptions of objects present in image patches to image patches, and next matching descriptions of objects present outside the patches to image patches, respectively. Finally, we presented extensive experimental results including ablation of all proposed components and comparison with related zero-reference methods, showing the effectiveness of our proposed method in task-based evaluation, targeting machine cognition, rather than assuming correlation between human perception with task-based performance.

%% file: Figures/Figures_QP.tex
\begin{figure}
    \rebuttalcontainer
    \centering
    \small
    \begin{subfigure}{.20\textwidth}
      \centering
        \includegraphics[width=1\linewidth]{Images/bboxes-figures/240516_181048_from_240507_1354_102.JPG}
        \caption*{}
    \end{subfigure} %
    \begin{subfigure}{.20\textwidth}
      \centering
        \includegraphics[width=1\linewidth]{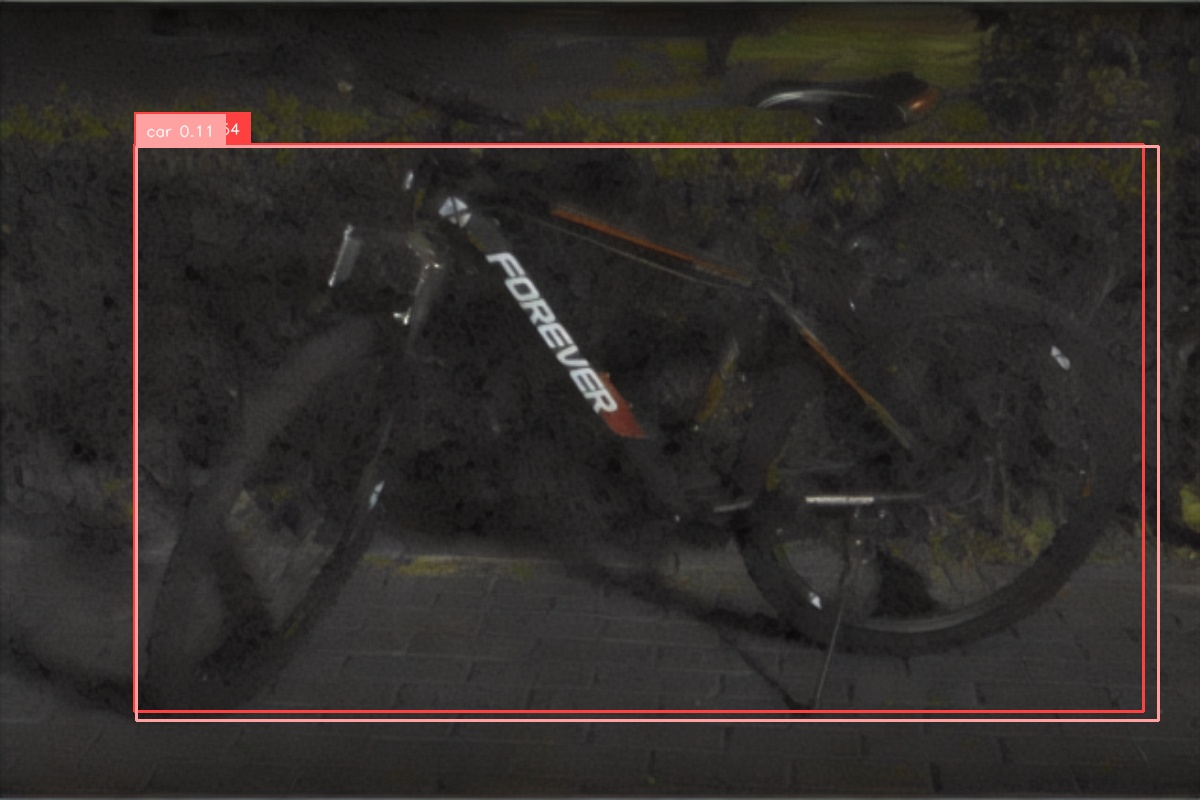}
        \caption*{}
    \end{subfigure} %

\vspace{-1.2em}

    \begin{subfigure}{.20\textwidth}
      \centering
        \includegraphics[width=1\linewidth]{Images/bboxes-figures/240516_181048_from_240507_1354_DSC02620.JPG}
        \caption*{ Ours}
    \end{subfigure} 
    \begin{subfigure}{.20\textwidth}
      \centering
        \includegraphics[width=1\linewidth]{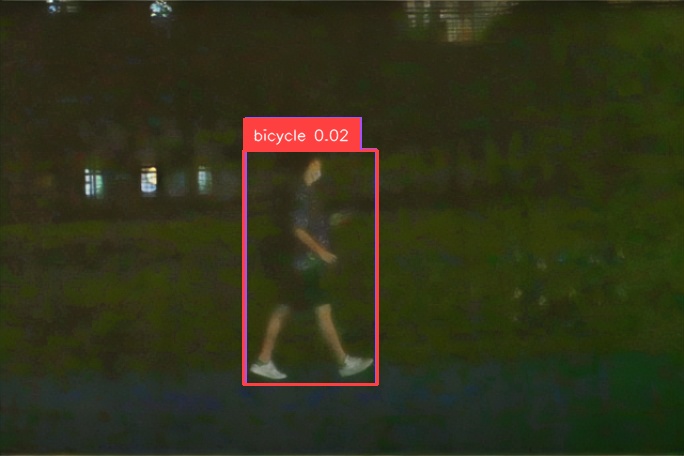}
        \caption*{ QuadPrior \cite{wang2024zero}}
    \end{subfigure} %


    \caption{\rebuttalcontainer Prompted by task-based evaluation results in Tab. \ref{tab:crossdataset_comparison}, we additionally visualize enhancement and detection of QuadPrior \cite{wang2024zero}. We observe that for this method, over-smoothing of extreme low-light instances leads to a reduced detection performance. This leads to a conclusion that in low-light enhancement, a correlation between task-based performance (machine cognition) and human perception cannot be assumed. }
    \label{fig:quadprior-figure}
\end{figure}

%% file: Sections/X_Acknowledgement.tex
\section*{Acknowledgements}
\label{sec:acknowledgements}
This work was supported in part by National Science and Technology Council, Taiwan, under Grant NSTC 112-2634-F-002-006 and by Qualcomm through a Taiwan University Research Collaboration Project.

%% file: Biographies/Biographies_Compilation.tex
\begin{IEEEbiography}[{\includegraphics
[width=1in,height=1.25in,clip,
keepaspectratio]{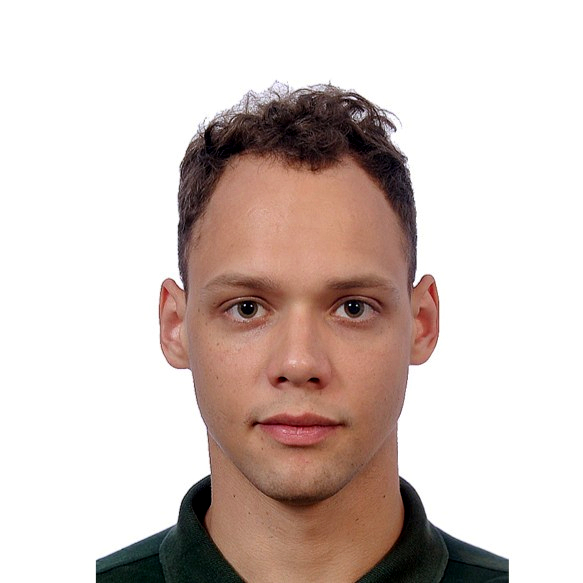}}]
{Igor Morawski} is currently working toward his Ph.D. degree at the Department of Computer Science and Information Engineering, National Taiwan University. He received his B.S. and M.S. degrees in Aerospace Engineering from the Military University of Technology in Warsaw, Poland, and his M.S. degree in Electrical Engineering from National Chung Cheng University, Taiwan. His research interests lie in the fields of computer vision, image processing and machine learning, including neural image signal processing pipelines, visual-linguistic models, object detection and event cameras.
\end{IEEEbiography}

\begin{IEEEbiography}[{\includegraphics
[width=1in,height=1.25in,clip,
keepaspectratio]{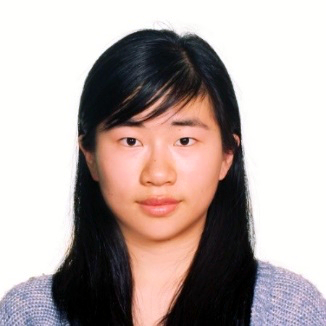}}]
{Kai He}
received the B.S. degree in electrical engineering from Shanghai Jiao Tong University, Shanghai, China, in 2015, and the Ph.D. degree in electrical engineering from Texas A\&M University, College Station, TX, USA, in 2020. She’s currently a Staff Engineer at Camera Machine Learning Team, Qualcomm Technologies, Inc., San Diego, CA, USA. Her research interests include machine learning, probabilistic modeling, and their applications in computational photography and computer vision.
\end{IEEEbiography}

\begin{IEEEbiography}[{\includegraphics
[width=1in,height=1.25in,clip,
keepaspectratio]{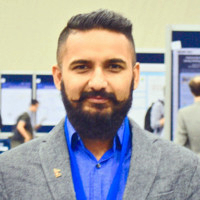}}]
{Shusil Dangi} received his PhD in 2019 from the Rochester Institute of Technology (RIT), Rochester NY, specializing in Machine Learning for Computational Medical Imaging. That same year, he joined Qualcomm's camera machine learning team, where he develops machine learning algorithms for smartphone camera applications. He has co-authored 25 academic publications and holds 2 patents related to camera machine learning. His research interests include image processing, image and language understanding, multi-modal understanding, and generative models, with a focus on applying these technologies to improve the quality of life.
\end{IEEEbiography}

\begin{IEEEbiography}[{\includegraphics
[width=1in,height=1.25in,clip,
keepaspectratio]{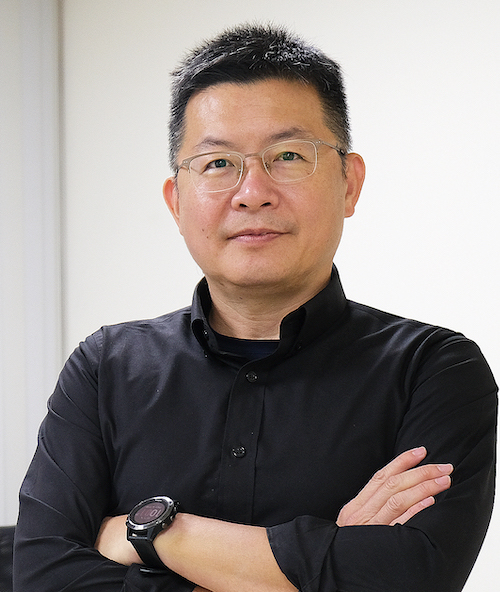}}]
{Winston H. Hsu (Senior Member, IEEE)} 
received the Ph.D. degree in electrical engineering from Columbia University, New York, NY, USA. He is keen to realizing advanced researches towards business deliverables via academia-industry collaborations and co-founding startups. Since 2007, he has been a Professor with the Graduate Institute of Networking and Multimedia and the Department of Computer Science and Information Engineering, National Taiwan University, Taipei, Taiwan. His research interests include large-scale image/video retrieval/mining, visual recognition, and machine intelligence. Dr. Hsu was an Associate Editor for the IEEE Transactions on Multimedia and on the Editorial Board for the IEEE Multimedia Magazine.
\end{IEEEbiography}

%% file: manuscript.bbl
\begin{thebibliography}{10}
\providecommand{\url}[1]{#1}
\csname url@samestyle\endcsname
\providecommand{\newblock}{\relax}
\providecommand{\bibinfo}[2]{#2}
\providecommand{\BIBentrySTDinterwordspacing}{\spaceskip=0pt\relax}
\providecommand{\BIBentryALTinterwordstretchfactor}{4}
\providecommand{\BIBentryALTinterwordspacing}{\spaceskip=\fontdimen2\font plus
\BIBentryALTinterwordstretchfactor\fontdimen3\font minus \fontdimen4\font\relax}
\providecommand{\BIBforeignlanguage}[2]{{%
\expandafter\ifx\csname l@#1\endcsname\relax
\typeout{** WARNING: IEEEtran.bst: No hyphenation pattern has been}%
\typeout{** loaded for the language `#1'. Using the pattern for}%
\typeout{** the default language instead.}%
\else
\language=\csname l@#1\endcsname
\fi
#2}}
\providecommand{\BIBdecl}{\relax}
\BIBdecl

\bibitem{loh2019getting}
Y.~P. Loh and C.~S. Chan, ``Getting to know low-light images with the exclusively dark dataset,'' \emph{Computer Vision and Image Understanding}, vol. 178, pp. 30--42, 2019.

\bibitem{xu2020learning}
K.~Xu, X.~Yang, B.~Yin, and R.~W. Lau, ``Learning to restore low-light images via decomposition-and-enhancement,'' in \emph{Proceedings of the IEEE/CVF conference on computer vision and pattern recognition}, 2020, pp. 2281--2290.

\bibitem{zheng2021adaptive}
C.~Zheng, D.~Shi, and W.~Shi, ``Adaptive unfolding total variation network for low-light image enhancement,'' in \emph{Proceedings of the IEEE/CVF international conference on computer vision}, 2021, pp. 4439--4448.

\bibitem{fan2022half}
C.-M. Fan, T.-J. Liu, and K.-H. Liu, ``Half wavelet attention on m-net+ for low-light image enhancement,'' in \emph{2022 IEEE International Conference on Image Processing (ICIP)}.\hskip 1em plus 0.5em minus 0.4em\relax IEEE, 2022, pp. 3878--3882.

\bibitem{SNRAware}
X.~Xu, R.~Wang, C.-W. Fu, and J.~Jia, ``Snr-aware low-light image enhancement,'' in \emph{2022 IEEE/CVF Conference on Computer Vision and Pattern Recognition (CVPR)}, 2022, pp. 17\,693--17\,703.

\bibitem{Wu_2023_CVPR}
Y.~Wu, C.~Pan, G.~Wang, Y.~Yang, J.~Wei, C.~Li, and H.~T. Shen, ``Learning semantic-aware knowledge guidance for low-light image enhancement,'' in \emph{Proceedings of the IEEE/CVF Conference on Computer Vision and Pattern Recognition (CVPR)}, June 2023, pp. 1662--1671.

\bibitem{Xu_2023_CVPR}
X.~Xu, R.~Wang, and J.~Lu, ``Low-light image enhancement via structure modeling and guidance,'' in \emph{Proceedings of the IEEE/CVF Conference on Computer Vision and Pattern Recognition (CVPR)}, June 2023, pp. 9893--9903.

\bibitem{wei2018deep}
C.~Wei, W.~Wang, W.~Yang, and J.~Liu, ``Deep retinex decomposition for low-light enhancement,'' \emph{arXiv preprint arXiv:1808.04560}, 2018.

\bibitem{zhang2019kindling}
Y.~Zhang, J.~Zhang, and X.~Guo, ``Kindling the darkness: A practical low-light image enhancer,'' in \emph{Proceedings of the 27th ACM international conference on multimedia}, 2019, pp. 1632--1640.

\bibitem{zhang2021beyond}
Y.~Zhang, X.~Guo, J.~Ma, W.~Liu, and J.~Zhang, ``Beyond brightening low-light images,'' \emph{International Journal of Computer Vision}, vol. 129, pp. 1013--1037, 2021.

\bibitem{yang2021sparse}
W.~Yang, W.~Wang, H.~Huang, S.~Wang, and J.~Liu, ``Sparse gradient regularized deep retinex network for robust low-light image enhancement,'' \emph{IEEE Transactions on Image Processing}, vol.~30, pp. 2072--2086, 2021.

\bibitem{zhang2022deep}
Z.~Zhang, H.~Zheng, R.~Hong, M.~Xu, S.~Yan, and M.~Wang, ``Deep color consistent network for low-light image enhancement,'' in \emph{Proceedings of the IEEE/CVF conference on computer vision and pattern recognition}, 2022, pp. 1899--1908.

\bibitem{guo2020zero}
C.~Guo, C.~Li, J.~Guo, C.~C. Loy, J.~Hou, S.~Kwong, and R.~Cong, ``Zero-reference deep curve estimation for low-light image enhancement,'' in \emph{Proceedings of the IEEE/CVF conference on computer vision and pattern recognition}, 2020, pp. 1780--1789.

\bibitem{li2021learning}
C.~Li, C.~Guo, and C.~C. Loy, ``Learning to enhance low-light image via zero-reference deep curve estimation,'' \emph{IEEE Transactions on Pattern Analysis and Machine Intelligence}, vol.~44, no.~8, pp. 4225--4238, 2021.

\bibitem{zheng2022semantic}
S.~Zheng and G.~Gupta, ``Semantic-guided zero-shot learning for low-light image/video enhancement,'' in \emph{Proceedings of the IEEE/CVF Winter conference on applications of computer vision}, 2022, pp. 581--590.

\bibitem{morawski2024unsupervised}
I.~Morawski, K.~He, S.~Dangi, and W.~H. Hsu, ``Unsupervised image prior via prompt learning and clip semantic guidance for low-light image enhancement,'' in \emph{Proceedings of the IEEE/CVF Conference on Computer Vision and Pattern Recognition}, 2024, pp. 5971--5981.

\bibitem{radford2021learning}
A.~Radford, J.~W. Kim, C.~Hallacy, A.~Ramesh, G.~Goh, S.~Agarwal, G.~Sastry, A.~Askell, P.~Mishkin, J.~Clark \emph{et~al.}, ``Learning transferable visual models from natural language supervision,'' in \emph{International conference on machine learning}.\hskip 1em plus 0.5em minus 0.4em\relax PMLR, 2021, pp. 8748--8763.

\bibitem{chen2018learning}
C.~Chen, Q.~Chen, J.~Xu, and V.~Koltun, ``Learning to see in the dark,'' in \emph{Proceedings of the IEEE conference on computer vision and pattern recognition}, 2018, pp. 3291--3300.

\bibitem{afifi2019else}
M.~Afifi and M.~S. Brown, ``What else can fool deep learning? addressing color constancy errors on deep neural network performance,'' in \emph{Proceedings of the IEEE/CVF international conference on computer vision}, 2019, pp. 243--252.

\bibitem{poor_visibility_benchmark}
W.~Yang, Y.~Yuan, W.~Ren, J.~Liu, W.~J. Scheirer, Z.~Wang, Zhang, and et~al., ``Advancing image understanding in poor visibility environments: A collective benchmark study,'' \emph{IEEE Transactions on Image Processing}, vol.~29, pp. 5737--5752, 2020.

\bibitem{ho2020denoising}
J.~Ho, A.~Jain, and P.~Abbeel, ``Denoising diffusion probabilistic models,'' \emph{Advances in neural information processing systems}, vol.~33, pp. 6840--6851, 2020.

\bibitem{mosleh2020hardware}
A.~Mosleh, A.~Sharma, E.~Onzon, F.~Mannan, N.~Robidoux, and F.~Heide, ``Hardware-in-the-loop end-to-end optimization of camera image processing pipelines,'' in \emph{Proceedings of the IEEE/CVF Conference on Computer Vision and Pattern Recognition}, 2020, pp. 7529--7538.

\bibitem{liang2021recurrent}
J.~Liang, J.~Wang, Y.~Quan, T.~Chen, J.~Liu, H.~Ling, and Y.~Xu, ``Recurrent exposure generation for low-light face detection,'' \emph{IEEE Transactions on Multimedia}, vol.~24, pp. 1609--1621, 2021.

\bibitem{morawski2021nod}
I.~Morawski, Y.-A. Chen, Y.-S. Lin, and W.~H. Hsu, ``Nod: Taking a closer look at detection under extreme low-light conditions with night object detection dataset,'' \emph{arXiv preprint arXiv:2110.10364}, 2021.

\bibitem{hong2021crafting}
Y.~Hong, K.~Wei, L.~Chen, and Y.~Fu, ``Crafting object detection in very low light,'' in \emph{BMVC}, vol.~1, 2021, p.~3.

\bibitem{morawski2022genisp}
I.~Morawski, Y.-A. Chen, Y.-S. Lin, S.~Dangi, K.~He, and W.~H. Hsu, ``Genisp: neural isp for low-light machine cognition,'' in \emph{Proceedings of the IEEE/CVF Conference on Computer Vision and Pattern Recognition}, 2022, pp. 630--639.

\bibitem{wang2023exploring}
J.~Wang, K.~C. Chan, and C.~C. Loy, ``Exploring clip for assessing the look and feel of images,'' in \emph{Proceedings of the AAAI Conference on Artificial Intelligence}, vol.~37, 2023, pp. 2555--2563.

\bibitem{liang2023iterative}
Z.~Liang, C.~Li, S.~Zhou, R.~Feng, and C.~C. Loy, ``Iterative prompt learning for unsupervised backlit image enhancement,'' in \emph{Proceedings of the IEEE/CVF International Conference on Computer Vision}, 2023, pp. 8094--8103.

\bibitem{wang2023tienet}
Y.~Wang, J.~Guo, R.~Wang, W.~He, and C.~Li, ``Tienet: task-oriented image enhancement network for degraded object detection,'' \emph{Signal, Image and Video Processing}, pp. 1--8, 2023.

\bibitem{hashmi2023featenhancer}
K.~A. Hashmi, G.~Kallempudi, D.~Stricker, and M.~Z. Afzal, ``Featenhancer: Enhancing hierarchical features for object detection and beyond under low-light vision,'' in \emph{Proceedings of the IEEE/CVF International Conference on Computer Vision}, 2023, pp. 6725--6735.

\bibitem{ljungbergh2023raw}
W.~Ljungbergh, J.~Johnander, C.~Petersson, and M.~Felsberg, ``Raw or cooked? object detection on raw images,'' in \emph{Scandinavian Conference on Image Analysis}.\hskip 1em plus 0.5em minus 0.4em\relax Springer, 2023, pp. 374--385.

\bibitem{yoshimura2023dynamicisp}
M.~Yoshimura, J.~Otsuka, A.~Irie, and T.~Ohashi, ``Dynamicisp: dynamically controlled image signal processor for image recognition,'' in \emph{Proceedings of the IEEE/CVF International Conference on Computer Vision}, 2023, pp. 12\,866--12\,876.

\bibitem{ExLPose_2023_CVPR}
S.~Lee, J.~Rim, B.~Jeong, G.~Kim, B.~Woo, H.~Lee, and S.~K. Sunghyun~Cho, ``Human pose estimation in extremely low-light conditions,'' in \emph{Proceedings of the IEEE/CVF Conference on Computer Vision and Pattern Recognition (CVPR)}, 2023.

\bibitem{zhang2023darkvision}
B.~Zhang, Y.~Guo, R.~Yang, Z.~Zhang, J.~Xie, J.~Suo, and Q.~Dai, ``Darkvision: a benchmark for low-light image/video perception,'' \emph{arXiv preprint arXiv:2301.06269}, 2023.

\bibitem{yi2023diff}
X.~Yi, H.~Xu, H.~Zhang, L.~Tang, and J.~Ma, ``Diff-retinex: Rethinking low-light image enhancement with a generative diffusion model,'' in \emph{Proceedings of the IEEE/CVF International Conference on Computer Vision}, 2023, pp. 12\,302--12\,311.

\bibitem{wang2024zero}
W.~Wang, H.~Yang, J.~Fu, and J.~Liu, ``Zero-reference low-light enhancement via physixcal quadruple priors,'' in \emph{Proceedings of the IEEE/CVF Conference on Computer Vision and Pattern Recognition}, 2024, pp. 26\,057--26\,066.

\bibitem{land1977retinex}
E.~H. Land, ``The retinex theory of color vision,'' \emph{Scientific american}, vol. 237, no.~6, pp. 108--129, 1977.

\bibitem{jobson1997properties}
D.~J. Jobson, Z.-u. Rahman, and G.~A. Woodell, ``Properties and performance of a center/surround retinex,'' \emph{IEEE transactions on image processing}, vol.~6, no.~3, pp. 451--462, 1997.

\bibitem{jobson1997multiscale}
------, ``A multiscale retinex for bridging the gap between color images and the human observation of scenes,'' \emph{IEEE Transactions on Image processing}, vol.~6, no.~7, pp. 965--976, 1997.

\bibitem{wang2013naturalness}
S.~Wang, J.~Zheng, H.-M. Hu, and B.~Li, ``Naturalness preserved enhancement algorithm for non-uniform illumination images,'' \emph{IEEE transactions on image processing}, vol.~22, no.~9, pp. 3538--3548, 2013.

\bibitem{fu2016fusion}
X.~Fu, D.~Zeng, Y.~Huang, Y.~Liao, X.~Ding, and J.~Paisley, ``A fusion-based enhancing method for weakly illuminated images,'' \emph{Signal Processing}, vol. 129, pp. 82--96, 2016.

\bibitem{guo2016lime}
X.~Guo, Y.~Li, and H.~Ling, ``Lime: Low-light image enhancement via illumination map estimation,'' \emph{IEEE Transactions on image processing}, vol.~26, no.~2, pp. 982--993, 2016.

\bibitem{li2017joint}
M.~Li, J.~Liu, W.~Yang, and Z.~Guo, ``Joint denoising and enhancement for low-light images via retinex model,'' in \emph{International Forum on Digital TV and Wireless Multimedia Communications}.\hskip 1em plus 0.5em minus 0.4em\relax Springer, 2017, pp. 91--99.

\bibitem{li2018structure}
M.~Li, J.~Liu, W.~Yang, X.~Sun, and Z.~Guo, ``Structure-revealing low-light image enhancement via robust retinex model,'' \emph{IEEE Transactions on Image Processing}, vol.~27, no.~6, pp. 2828--2841, 2018.

\bibitem{wang2023low}
W.~Wang, D.~Yan, X.~Wu, W.~He, Z.~Chen, X.~Yuan, and L.~Li, ``Low-light image enhancement based on virtual exposure,'' \emph{Signal Processing: Image Communication}, vol. 118, p. 117016, 2023.

\bibitem{jiang2021enlightengan}
Y.~Jiang, X.~Gong, D.~Liu, Y.~Cheng, C.~Fang, X.~Shen, J.~Yang, P.~Zhou, and Z.~Wang, ``Enlightengan: Deep light enhancement without paired supervision,'' \emph{IEEE transactions on image processing}, vol.~30, pp. 2340--2349, 2021.

\bibitem{dong2022abandoning}
X.~Dong, W.~Xu, Z.~Miao, L.~Ma, C.~Zhang, J.~Yang, Z.~Jin, A.~B.~J. Teoh, and J.~Shen, ``Abandoning the bayer-filter to see in the dark,'' in \emph{Proceedings of the IEEE/CVF Conference on Computer Vision and Pattern Recognition}, 2022, pp. 17\,431--17\,440.

\bibitem{liu2021retinex}
R.~Liu, L.~Ma, J.~Zhang, X.~Fan, and Z.~Luo, ``Retinex-inspired unrolling with cooperative prior architecture search for low-light image enhancement,'' in \emph{Proceedings of the IEEE/CVF Conference on Computer Vision and Pattern Recognition}, 2021, pp. 10\,561--10\,570.

\bibitem{wang2022low}
Y.~Wang, R.~Wan, W.~Yang, H.~Li, L.-P. Chau, and A.~Kot, ``Low-light image enhancement with normalizing flow,'' in \emph{Proceedings of the AAAI conference on artificial intelligence}, vol.~36, 2022, pp. 2604--2612.

\bibitem{wu2022uretinex}
W.~Wu, J.~Weng, P.~Zhang, X.~Wang, W.~Yang, and J.~Jiang, ``Uretinex-net: Retinex-based deep unfolding network for low-light image enhancement,'' in \emph{Proceedings of the IEEE/CVF conference on computer vision and pattern recognition}, 2022, pp. 5901--5910.

\bibitem{ma2022toward}
L.~Ma, T.~Ma, R.~Liu, X.~Fan, and Z.~Luo, ``Toward fast, flexible, and robust low-light image enhancement,'' in \emph{Proceedings of the IEEE/CVF Conference on Computer Vision and Pattern Recognition}, 2022, pp. 5637--5646.

\bibitem{fu2023you}
H.~Fu, W.~Zheng, X.~Meng, X.~Wang, C.~Wang, and H.~Ma, ``You do not need additional priors or regularizers in retinex-based low-light image enhancement,'' in \emph{Proceedings of the IEEE/CVF Conference on Computer Vision and Pattern Recognition}, 2023, pp. 18\,125--18\,134.

\bibitem{cai2018learning}
J.~Cai, S.~Gu, and L.~Zhang, ``Learning a deep single image contrast enhancer from multi-exposure images,'' \emph{IEEE Transactions on Image Processing}, vol.~27, no.~4, pp. 2049--2062, 2018.

\bibitem{chen2019seeing}
C.~Chen, Q.~Chen, M.~N. Do, and V.~Koltun, ``Seeing motion in the dark,'' in \emph{Proceedings of the IEEE/CVF International conference on computer vision}, 2019, pp. 3185--3194.

\bibitem{hai2023r2rnet}
J.~Hai, Z.~Xuan, R.~Yang, Y.~Hao, F.~Zou, F.~Lin, and S.~Han, ``R2rnet: Low-light image enhancement via real-low to real-normal network,'' \emph{Journal of Visual Communication and Image Representation}, vol.~90, p. 103712, 2023.

\bibitem{bychkovsky2011learning}
V.~Bychkovsky, S.~Paris, E.~Chan, and F.~Durand, ``Learning photographic global tonal adjustment with a database of input/output image pairs,'' in \emph{CVPR 2011}.\hskip 1em plus 0.5em minus 0.4em\relax IEEE, 2011, pp. 97--104.

\bibitem{jiang2019learning}
H.~Jiang and Y.~Zheng, ``Learning to see moving objects in the dark,'' in \emph{Proceedings of the IEEE/CVF International Conference on Computer Vision}, 2019, pp. 7324--7333.

\bibitem{yang2020fidelity}
W.~Yang, S.~Wang, Y.~Fang, Y.~Wang, and J.~Liu, ``From fidelity to perceptual quality: A semi-supervised approach for low-light image enhancement,'' in \emph{Proceedings of the IEEE/CVF conference on computer vision and pattern recognition}, 2020, pp. 3063--3072.

\bibitem{liang2024pie}
D.~Liang, Z.~Xu, L.~Li, M.~Wei, and S.~Chen, ``Pie: Physics-inspired low-light enhancement,'' \emph{International Journal of Computer Vision}, pp. 1--22, 2024.

\bibitem{Fu_2023_CVPR_Learning}
Z.~Fu, Y.~Yang, X.~Tu, Y.~Huang, X.~Ding, and K.-K. Ma, ``Learning a simple low-light image enhancer from paired low-light instances,'' in \emph{Proceedings of the IEEE/CVF Conference on Computer Vision and Pattern Recognition (CVPR)}, June 2023, pp. 22\,252--22\,261.

\bibitem{buchsbaum1980spatial}
G.~Buchsbaum, ``A spatial processor model for object colour perception,'' \emph{Journal of the Franklin institute}, vol. 310, no.~1, pp. 1--26, 1980.

\bibitem{Feng_2024_CVPR}
Y.~Feng, S.~Hou, H.~Lin, Y.~Zhu, P.~Wu, W.~Dong, J.~Sun, Q.~Yan, and Y.~Zhang, ``Difflight: Integrating content and detail for low-light image enhancement,'' in \emph{Proceedings of the IEEE/CVF Conference on Computer Vision and Pattern Recognition (CVPR) Workshops}, June 2024, pp. 6143--6152.

\bibitem{jiang2023low}
H.~Jiang, A.~Luo, H.~Fan, S.~Han, and S.~Liu, ``Low-light image enhancement with wavelet-based diffusion models,'' \emph{ACM Transactions on Graphics (TOG)}, vol.~42, no.~6, pp. 1--14, 2023.

\bibitem{zhou2023pyramid}
D.~Zhou, Z.~Yang, and Y.~Yang, ``Pyramid diffusion models for low-light image enhancement,'' \emph{arXiv preprint arXiv:2305.10028}, 2023.

\bibitem{li2023low}
Z.~Li, Y.~Wang, and J.~Zhang, ``Low-light image enhancement with knowledge distillation,'' \emph{Neurocomputing}, vol. 518, pp. 332--343, 2023.

\bibitem{wang2021hla}
W.~Wang, W.~Yang, and J.~Liu, ``Hla-face: Joint high-low adaptation for low light face detection,'' in \emph{Proceedings of the IEEE/CVF Conference on Computer Vision and Pattern Recognition}, 2021, pp. 16\,195--16\,204.

\bibitem{xu2021arid}
Y.~Xu, J.~Yang, H.~Cao, K.~Mao, J.~Yin, and S.~See, ``Arid: A new dataset for recognizing action in the dark,'' in \emph{Deep Learning for Human Activity Recognition: Second International Workshop, DL-HAR 2020, Held in Conjunction with IJCAI-PRICAI 2020, Kyoto, Japan, January 8, 2021, Proceedings 2}.\hskip 1em plus 0.5em minus 0.4em\relax Springer, 2021, pp. 70--84.

\bibitem{he2016deep}
K.~He, X.~Zhang, S.~Ren, and J.~Sun, ``Deep residual learning for image recognition,'' in \emph{Proceedings of the IEEE conference on computer vision and pattern recognition}, 2016, pp. 770--778.

\bibitem{liu2017image}
D.~Liu, B.~Wen, X.~Liu, Z.~Wang, and T.~S. Huang, ``When image denoising meets high-level vision tasks: A deep learning approach,'' \emph{arXiv preprint arXiv:1706.04284}, 2017.

\bibitem{aakerberg2022semantic}
A.~Aakerberg, A.~S. Johansen, K.~Nasrollahi, and T.~B. Moeslund, ``Semantic segmentation guided real-world super-resolution,'' in \emph{Proceedings of the IEEE/CVF Winter Conference on Applications of Computer Vision}, 2022, pp. 449--458.

\bibitem{wang2018recovering}
X.~Wang, K.~Yu, C.~Dong, and C.~C. Loy, ``Recovering realistic texture in image super-resolution by deep spatial feature transform,'' in \emph{Proceedings of the IEEE conference on computer vision and pattern recognition}, 2018, pp. 606--615.

\bibitem{li2020blind}
X.~Li, C.~Chen, S.~Zhou, X.~Lin, W.~Zuo, and L.~Zhang, ``Blind face restoration via deep multi-scale component dictionaries,'' in \emph{European conference on computer vision}.\hskip 1em plus 0.5em minus 0.4em\relax Springer, 2020, pp. 399--415.

\bibitem{robidoux2021end}
N.~Robidoux, L.~E.~G. Capel, D.-e. Seo, A.~Sharma, F.~Ariza, and F.~Heide, ``End-to-end high dynamic range camera pipeline optimization,'' in \emph{Proceedings of the IEEE/CVF Conference on Computer Vision and Pattern Recognition}, 2021, pp. 6297--6307.

\bibitem{zang2022open}
Y.~Zang, W.~Li, K.~Zhou, C.~Huang, and C.~C. Loy, ``Open-vocabulary detr with conditional matching,'' in \emph{European Conference on Computer Vision}.\hskip 1em plus 0.5em minus 0.4em\relax Springer, 2022, pp. 106--122.

\bibitem{kuo2022f}
W.~Kuo, Y.~Cui, X.~Gu, A.~Piergiovanni, and A.~Angelova, ``F-vlm: Open-vocabulary object detection upon frozen vision and language models,'' \emph{arXiv preprint arXiv:2209.15639}, 2022.

\bibitem{zhou2022extract}
C.~Zhou, C.~C. Loy, and B.~Dai, ``Extract free dense labels from clip,'' in \emph{European Conference on Computer Vision}.\hskip 1em plus 0.5em minus 0.4em\relax Springer, 2022, pp. 696--712.

\bibitem{cheng2024yolo}
T.~Cheng, L.~Song, Y.~Ge, W.~Liu, X.~Wang, and Y.~Shan, ``Yolo-world: Real-time open-vocabulary object detection,'' \emph{arXiv preprint arXiv:2401.17270}, 2024.

\bibitem{shao2019objects365}
S.~Shao, Z.~Li, T.~Zhang, C.~Peng, G.~Yu, X.~Zhang, J.~Li, and J.~Sun, ``Objects365: A large-scale, high-quality dataset for object detection,'' in \emph{Proceedings of the IEEE/CVF international conference on computer vision}, 2019, pp. 8430--8439.

\bibitem{lv2022backlitnet}
X.~Lv, S.~Zhang, Q.~Liu, H.~Xie, B.~Zhong, and H.~Zhou, ``Backlitnet: A dataset and network for backlit image enhancement,'' \emph{Computer Vision and Image Understanding}, vol. 218, p. 103403, 2022.

\bibitem{YOLO5Face}
D.~Qi, W.~Tan, Q.~Yao, and J.~Liu, ``Yolo5face: Why reinventing a face detector,'' 2021.

\bibitem{liu2021benchmarking}
J.~Liu, D.~Xu, W.~Yang, M.~Fan, and H.~Huang, ``Benchmarking low-light image enhancement and beyond,'' \emph{International Journal of Computer Vision}, vol. 129, pp. 1153--1184, 2021.

\bibitem{huang2021neighbor2neighbor}
T.~Huang, S.~Li, X.~Jia, H.~Lu, and J.~Liu, ``Neighbor2neighbor: Self-supervised denoising from single noisy images,'' in \emph{Proceedings of the IEEE/CVF conference on computer vision and pattern recognition}, 2021, pp. 14\,781--14\,790.

\end{thebibliography}
